\documentclass{article}
\usepackage{iclr2027_conference,times}
\usepackage[utf8]{inputenc}
\usepackage[T1]{fontenc}
\usepackage{hyperref}
\usepackage{url}
\usepackage{booktabs}
\usepackage{graphicx}
\usepackage{pgfplots}
\pgfplotsset{compat=1.18}
\usepackage{xcolor}
\usepackage{amsmath}
\usepackage{amssymb}
\usepackage{mathtools}
\usepackage{amsthm}
\usepackage{algorithm}
\usepackage{algpseudocode}
\usepackage{microtype}
\usepackage{placeins}
\usepackage{listings}

\mathtoolsset{showonlyrefs}
\hypersetup{
  pdftitle={Learning from Context: Leave-One-Out Prediction in Jump Diffusions},
  pdfauthor={Christian Wald}
}
\newtheorem{theorem}{Theorem}[section]
\newtheorem{proposition}[theorem]{Proposition}
\newtheorem{lemma}[theorem]{Lemma}

\theoremstyle{definition}
\newtheorem{example}[theorem]{Example}
\theoremstyle{remark}
\newtheorem{remark}[theorem]{Remark}
\newcommand{\E}{\mathbb E}
\newcommand{\R}{\mathbb R}
\newcommand{\cV}{\mathcal V}
\newcommand{\KL}{\operatorname{KL}}
\newcommand{\softmax}{\operatorname{softmax}}

\title{Learning from Context: Leave-One-Out Prediction in Jump Diffusions}
\author{Christian Wald\\
  INSA Lyon, Centrale Lyon, Universit\'e Lyon 1, Universit\'e Jean Monnet,\\
  CNRS, ICJ UMR5208, 69621 Villeurbanne, France}
\iclrfinalcopy

\begin{document}
\maketitle
\pagestyle{plain}

\begin{abstract}
Continuous diffusion models generate discrete data through stochastic
dynamics in an embedding space. We consider jump diffusions that additionally reset coordinates to embeddings of categorical values. A coordinate's own corrupted state can make
prediction easy, potentially reducing the pressure to learn dependencies
between coordinates. We study leave-one-out (LOO) prediction, in which each coordinate is predicted from the corrupted states of the other coordinates. We show that, under stated assumptions, jump diffusions with complete coordinate resets and state-independent clocks preserve the prescribed marginals when reset destinations use exact LOO probabilities and the drift uses the full endpoint posterior. We further show that this drift can be constructed from LOO probabilities by analytic likelihood completion without additional network evaluations.
{\color{black}We evaluate unrestricted, LOO, and completed predictions on Text8 and LM1B
text generation and ZINC250k molecular graph generation.
LOO-based variants improve key generation metrics over matched unrestricted
controls on all three datasets. Controlled interventions assess reliance on
local inputs and context, and we analyze the effect of analytic likelihood
completion.}\end{abstract}

\section{Introduction}
\label{sec:introduction}

\begingroup
\color{black}
Generating text and molecular graphs requires learning dependencies between
discrete coordinates. Continuous (jump) diffusion models approach this task by
corrupting token embeddings and learning to predict their endpoints
\citep{li2022diffusionlm,dieleman2022continuous,gulrajani2023plaid,jutrasdube2026sticky}.
Yet a coordinate's own noisy embedding can reveal its target without
requiring the model to learn the surrounding dependencies. A masked position
in masked diffusion or flow matching provides no such local token evidence
\citep{sahoo2024mdlm,gat2024dfm}. For continuous models, increasing noise to
remove this shortcut can also destroy useful context \citep{pynadath2026candi}.
We ask whether excluding a coordinate from its own prediction, so called leave-one-out (LOO) prediction, encourages
useful contextual learning, and how to do so while preserving the intended probability path.

We study LOO prediction in embedding-space jump diffusions,
where continuous motion is combined with jumps to token embeddings.
Let $Y=(Y_1,\ldots,Y_L)$ be a token sequence, $X_t=(X_{1,t},\ldots,X_{L,t})$
its corrupted embeddings in continuous space at time $t$, and denote by $X_{-i,t}$ all coordinates except $i$.
For a token $a$, the full and LOO posteriors are
\[
\begin{aligned}
  p_{i,t}(a\mid x)
    &= \mathbb P(Y_i=a\mid X_t=x),\\
  q_{i,t}(a\mid x_{-i})
    &= \mathbb P(Y_i=a\mid X_{-i,t}=x_{-i}).
\end{aligned}
\]
Predicting $q_{i,t}$ requires estimating $Y_i$ without access to $X_{i,t}$.
Our four-token parity experiment illustrates why this can matter: low
full-state prediction loss can precede learning the dependence needed for
joint generation (Table~\ref{tab:local-shortcut}).

Excluding local inputs is established in blind-spot denoising
\citep{batson2019noise2self,laine2019blindspot} and discrete diffusion
\citep{sun2023hollow,gourevitch2026loo,casado2026discrete}. For a broader overview of related work, see Appendix~\ref{app:extended-related-work}.
The gap we address is how to use these
$q_{i,t}(a\mid x_{-i})$ predictions in processes with both a continuous drift and
coordinate resets, i.e. jumps to the token embeddings. In this case the input of the network provides noisy contextual information before discrete decisions. Continuous motion resumes after each reset, and later resets can replace an earlier, possibly wrong, hypothesis. LOO makes the last point particularly meaningful: the learned reset distribution $q_{\theta,i,t}(a\mid x_{-i})$ depends only on the surrounding state $x_{-i}$, so an incorrect local state $x_{i}$ at time $t$ cannot directly influence this reset distribution.

The distinction between $p_{i,t}$ and $q_{i,t}$ is essential.
Posterior averaging of conditional generators uses the full posterior $p_{i,t}(a\mid x)$
\citep{holderrieth2025generator}, whereas LOO training learns $ q_{i,t}(a\mid x_{-i})$ without using $x_i$. We show that substituting the latter into the drift
can change the prescribed marginal path (Example~\ref{ex:loo-drift-failure}).
For jump diffusion, however,
exact LOO jump probabilities $q_{i,t}(a\mid x_{-i})$  preserve the correct probability curve when the drift
retains the full posterior, under the assumptions of
Proposition~\ref{prop:contextual-generator}. Analytic likelihood completion
combines LOO predictions with the local corruption likelihood to recover
the full posterior. Thus one network can supply both prediction roles
without additional network evaluations.

We compare unrestricted, raw LOO, and completed predictors on Text8,
LM1B, and ZINC250k. LOO improves key generation metrics over matched
unrestricted controls, while completion's effect depends on the path
and dataset. Local-state and subject--verb interventions test contextual
information use, see Section~\ref{sec:experiments}.

Our contributions are:
\begin{itemize}
  \item A marginal-preservation result for complete coordinate resets
  using LOO destinations and full-posterior drift, with a
  counterexample for substituting LOO probabilities into the drift.

  \item A repeated-reset Brownian bridge whose marginals converge to
  the data law, using one LOO network and analytic
  likelihood completion for the two prediction roles.

  \item Matched-budget experiments on Text8, LM1B, and ZINC250k showing
  when LOO improves over unrestricted prediction and when raw LOO
  drift outperforms completion despite lacking a marginal-preservation
  guarantee.

  \item We demonstrate empirically that LOO drift without analytic completion can help recover from wrong predictions at earlier time steps, offering a possible explanation for its better performance in some experiments despite lacking a marginal-preservation guarantee. Controlled LM1B interventions show that a token's own noisy input
  can dominate unmasked predictions and final token choices, even when it
  conflicts with context. On LM1B, LOO improves subject tracking and the usefulness of older word order in generated text.
\end{itemize}
\endgroup

\section{Conditional paths and leave-one-out prediction}
\label{sec:background}

Let $\cV$ be a finite vocabulary and let $p_{\mathrm{data}}\in \mathcal P(\cV^L)$ be a data distribution on sequences of length $L$.
Choose a random variable $Y=(Y_1,\ldots,Y_L)\sim p_{\mathrm{data}}$, and consider an embedding $e:\cV\to\R^d$.
For recovery of token sequences, we assume that $e$ is injective. Write
$E(Y)=(e(Y_1),\ldots,e(Y_L))$ and for $x=(x_1,\ldots,x_L)\in \R^{dL}$ let $x^{i\leftarrow z}$ denote $x$
with position $i$ replaced by $z$ i.e. $x^{i\leftarrow z}=(x_1,\ldots,x_{i-1},z,\ldots,x_L)$. We write
$x_{-i}=(x_1,\ldots,x_{i-1},x_{i+1},\ldots,x_L)$ for the sequence
with position $i$ omitted, and use $X_{-i,t}$ analogously.
Time runs from the source at $t=0$ toward data at $t=1$.
For each process below, write $p_t=\mathcal L(X_t)$ for its state law and
$p_{i,t}(a\mid x)=\mathbb P(Y_i=a\mid X_t=x)$ for the endpoint posterior
at position $i$.

\subsection{Discrete flow matching}
\label{sec:dfm-background}
Discrete flow matching (DFM) uses a probability path on token sequences and
learns a jump rate kernel that generates its marginals \citep{gat2024dfm}.
Appendix~\ref{app:jump-terminology} summarizes the terminology.
For the masked path, let $\mathsf m\notin\cV$ be a mask and we consider the following
discrete process $X^y_t\in(\cV\cup\{\mathsf m\})^L$ starting from
$X_0^y=(\mathsf m,\ldots,\mathsf m)$. Let $\kappa:[0,1]\to[0,1]$ be
continuous and nondecreasing, continuously differentiable on $[0,1)$, with
$\kappa(0)=0$, $\kappa(1)=1$, and $\kappa(t)<1$ for $t<1$.
For a fixed endpoint $y$, take independent reveal times $T_1,\ldots,T_L$
with $\mathbb P(T_i\le t)=\kappa(t)$ and define the conditional process
$X_t^y=(X_{1,t}^y,\ldots,X_{L,t}^y)$ by setting
$X_{i,t}^y=\mathsf m$ for $t<T_i$ and $X_{i,t}^y=y_i$ for $t\ge T_i$.
Each position is therefore revealed independently with probability
$\kappa(t)$ at time $t$, so
\begin{equation}
 \mathbb P(X_t^y=x)
 =\prod_{i=1}^L\left[
 (1-\kappa(t))\mathbf 1\{x_i=\mathsf m\}
 +\kappa(t)\mathbf 1\{x_i=y_i\}\right].
 \label{eq:masked-flow-path}
\end{equation}
Since $\kappa(1)=1$, $X_1^y=y$ almost surely. Hence the mixture
$p_t=\mathcal L(X_t^Y)$ satisfies $p_1=p_{\mathrm{data}}$.
This conditional process reveals $y_i$ at rate
$\kappa'(t)/(1-\kappa(t))$ while position $i$ is masked and then leaves
it fixed. \citet{gat2024dfm} obtain the jump rate kernel for the marginal
process $X_t$ with law $p_t$ through the per-position rates
\begin{equation}
 r_{i,t}(x_i\to a\mid x)
 =\mathbf 1\{x_i=\mathsf m\}
   \frac{\kappa'(t)}{1-\kappa(t)}
   p_{i,t}(a\mid x),\qquad a\in\cV.
 \label{eq:masked-flow-rates}
\end{equation}
Thus generation requires $p_{i,t}(a\mid x)$, a categorical endpoint posterior at each position,
conditioned on the entire current sequence. These posteriors can be learned
by cross-entropy on samples of \eqref{eq:masked-flow-path}
\citep{gat2024dfm,sahoo2024mdlm}. At a masked position, the local mask
provides no evidence about its target token. 
\subsection{Continuous embeddings and endpoint posteriors}
\label{sec:embedding-background}
We briefly review continuous diffusion for categorical data using endpoint-posterior prediction \citep{dieleman2022continuous} and posterior averaging of conditional generators \citep{holderrieth2025generator}. We illustrate these established ideas with Brownian bridges \citep{liu2022bridges}, which also provide the continuous component of our later construction.
For each endpoint $y\in\cV^L$, consider Brownian bridges
$X_t^y=(X_{1,t}^y,\ldots,X_{L,t}^y)$ satisfying, for $0\le t<1$,
\begin{equation}
 dX_{i,t}^y=\frac{e(y_i)-X_{i,t}^y}{1-t}\,dt+\sigma_d\,dW_{i,t},
 \qquad X_{i,0}^y\sim\mathcal N(0,\sigma_s^2I_d),\quad\sigma_d>0, \sigma_s\ge0
 \label{eq:continuous-conditional-bridge}
\end{equation}
These paths extend continuously to $t=1$ with
$X_1^y=E(y)$ almost surely.
Let $p_t=\mathcal L(X_t^Y)$ for $Y\sim p_{\mathrm{data}}$.
Posterior averaging gives the marginal SDE
$dX_{i,t}=b_{i,t}(X_t)\,dt+\sigma_d\,dW_{i,t}$, with $X_0\sim \mathcal N(0,\sigma_s^2I_{dL})$, $X_1\sim \mathcal L(E(Y))$ and
\begin{equation}
 b_{i,t}(x)=\frac{\E[e(Y_i)\mid X_t^Y=x]-x_i}{1-t}
 =\frac{\sum_{a\in\cV}p_{i,t}(a\mid x)e(a)-x_i}{1-t}.
 \label{eq:continuous-marginal-drift}
\end{equation}
Under the usual existence and uniqueness assumptions for the forward
equation, the resulting process has law $p_t$ at each time $t$.
We can learn $p_{i,t}(a\mid x)$ with the loss
\begin{equation}
 \mathcal L_{\mathrm{CE}}(\theta)
 =\E_{t,Y,X_t^Y}\left[-\sum_i
       \log p_{\theta,i,t}(Y_i\mid X^Y_t)\right].
 \label{eq:full-endpoint-loss}
\end{equation}

\subsection{An endpoint-directed bridge with repeated resets}

\begin{figure}[t]
 \centering
 \includegraphics[width=\linewidth]{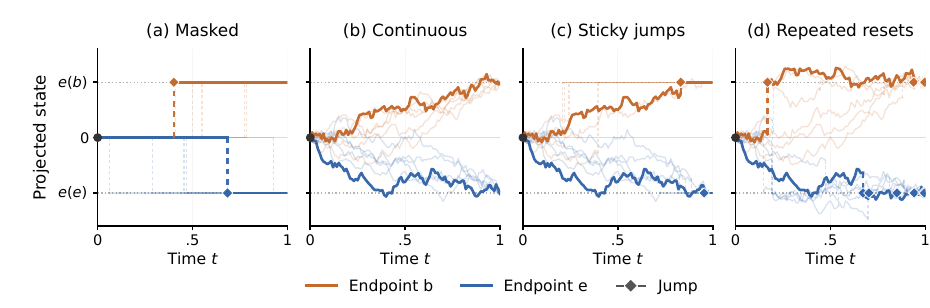}
 \caption{\textbf{Four paths, shown against time.}
 States are projected onto $e(b)$; colors distinguish the antipodal
 unit-circle endpoints $e(b)$ and $e(e)$. Dashed segments mark jumps.
 Masked diffusion reveals once; the schematic sticky jump diffusion
 \citep{jutrasdube2026sticky} stays fixed after commitment.
 The illustrated repeated-reset trajectories use independent clocks
with $\lambda(t)=(1-t)^{-1}$ and continue diffusing after each reset.}
 \label{fig:path-comparison}
\end{figure}

For a fixed endpoint $y$, consider the conditional jump diffusion
$X_t^y=(X_{1,t}^y,\ldots,X_{L,t}^y)$ given by
\begin{equation}
 dX_{i,t}^y
 =\frac{e(y_i)-X_{i,t-}^y}{1-t}\,dt+\sigma_d\,dW_{i,t}
   +(e(y_i)-X_{i,t-}^y)\,dN_{i,t},
 \qquad X_{i,0}^y\sim\mathcal N(0,\sigma_s^2I_d).
 \label{eq:conditional-path}
\end{equation}
Each coordinate follows Brownian bridge dynamics between jumps, resets
exactly to $e(y_i)$ at every jump, and resumes diffusing immediately afterwards.
Here $N_i$ are independent Poisson counting processes with a nonnegative
deterministic rate $\lambda(t)$ satisfying
$\int_0^t\lambda(s)\,ds<\infty$ for every $t<1$.
The notation $X_{i,t-}^y$ denotes the state just before time $t$.
For example, $\lambda_\gamma(t)=\gamma t^{\gamma-1}/(1-t^\gamma)$,
$\gamma\in\{1,2,4\}$, gives first-reset probability
$\mathbb P(N_{i,t}\ge1)=t^\gamma$. Figure~\ref{fig:path-comparison} illustrates one
choice; Appendix~\ref{app:conditional-sde} gives the conditional transition
laws.

\begin{remark}[Bridge clock]
We write \eqref{eq:conditional-path} and the derivations below for
$\kappa(t)=t$. With a continuous bridge clock $u=\kappa(t)$, the drift
factor becomes $\kappa'(t)/(1-\kappa(t))$ and the diffusion coefficient
becomes $\sigma_d\sqrt{\kappa'(t)}$. The reset rate $\lambda(t)$ is
specified independently in physical time. The Text8 and ZINC250k
conditional-path comparisons use $\kappa(t)=t^2$, giving drift factor
$2t/(1-t^2)$ and diffusion coefficient
$\sigma_d\sqrt{2t}$; the jump-schedule comparisons vary only
$\lambda_\gamma(t)$. Training states and analytic likelihoods use the
conditional laws for the same clock.
\end{remark}

Sampling $Y\sim p_{\mathrm{data}}$ defines the mixture $X_t^Y$ and its
marginal curve $p_t=\mathcal L(X_t^Y)$, with
$p_0=\mathcal N(0,\sigma_s^2I_{dL})$.
Posterior averaging \citep[Proposition~1]{holderrieth2025generator}
gives the following operator, whose weak forward equation is satisfied by
$p_t$ (proved directly in Appendix~\ref{app:proof-contextual-generator}):
\begin{align}
 \mathcal A_t f(x)
 ={}&\sum_i\left[
   \frac{\sum_{a\in\cV}\mathbb P(Y_i=a\mid X_t^Y=x)e(a)-x_i}{1-t}
      \cdot\nabla_i f(x)+\frac{\sigma_d^2}{2}\Delta_i f(x)\right]
      \nonumber\\
 &+\lambda(t)\sum_{i,a}\mathbb P(Y_i=a\mid X_t^Y=x)
      \left[f(x^{i\leftarrow e(a)})-f(x)\right].
 \label{eq:averaged-generator}
\end{align}
Here $\nabla_i$ and $\Delta_i$ are the gradient and Laplacian with respect
to $x_i$. The conditional paths concentrate on their endpoints as
$t\uparrow1$, so the marginal curve extends to
$p_1=\mathcal L(E(Y))$, or $p_1=p_{\mathrm{data}}$ when identifying tokens
with their embeddings.

The only unknown quantities in \eqref{eq:averaged-generator} are 
 $p_{i,t}(a\mid x)$, which determine both
the drift and the reset probabilities. We therefore learn $p_{\theta,i,t}$
from conditional-path samples using
\begin{equation}
 \mathcal L_{\mathrm{full}}(\theta)
 =\E_{t,Y,X_t^Y}\left[-\sum_i
       \log p_{\theta,i,t}(Y_i\mid X_t^Y)\right].
 \label{eq:full-jump-loss}
\end{equation}
Its unrestricted minimizer is $p_{i,t}$. Appendices~\ref{app:posterior-learning}
and~\ref{app:sampler} give the posterior-learning argument and sampling
details, respectively.

\begin{remark}[Sampling training states]
\label{rem:training-states}
For $y_i=a$, write $p_{i,t}(\cdot\mid a)=\mathcal L(X_{i,t}^y)$ for the
conditional state law; the argument order distinguishes it from the
endpoint posterior $p_{i,t}(a\mid x)$.
To estimate \eqref{eq:full-jump-loss}, draw $t$ and
$Y\sim p_{\mathrm{data}}$, then sample the coordinates independently
from $p_{i,t}(\cdot\mid Y_i)$. Each coordinate requires only the last
reset time (or no reset) and a Gaussian draw.
Appendix~\ref{app:conditional-sde} gives the mixture law, and
Proposition~\ref{prop:shared-schedule-law} gives its closed form when
$\lambda(t)=\kappa'(t)/(1-\kappa(t))$.
\end{remark}

\begin{remark}[Local endpoint information]
\label{rem:local-denoising-shortcut}
For uniform four-token even-parity data, Figure~\ref{fig:local-information}
\textcolor{black}{in Appendix~\ref{app:parity-learning}}
shows that the current local state already reveals much of its endpoint:
nearest-embedding decoding from $X_{i,t}$ alone is $93.9\%$ accurate at
$t=0.2$ and $99.9\%$ at $t=0.5$.
Thus, endpoint prediction can be easy without learning the dependence
between positions required for joint generation.
\end{remark}
\section{Learning contextual resets}
\label{sec:construction}

Figure~\ref{fig:local-information} \textcolor{black}{in Appendix~\ref{app:parity-learning}}
shows that the current local state can
already reveal most of its endpoint token, making endpoint prediction easy
without learning dependencies between positions. To remove this shortcut
from the jump predictor, we exclude $X_{i,t}$ when predicting the reset
destination at position $i$. This restriction changes the posterior learned
by cross-entropy, so using it in the dynamics requires care. We show that it
can be used while still preserving $p_t$, provided the
continuous drift retains $p_{i,t}(\cdot\mid X_t)$.

\subsection{Endpoint prediction without local information}
Following contextual token prediction \citep{shin2020tta} and
leave-one-out denoising in discrete diffusion \citep{gourevitch2026loo},
we consider a leave-one-out (LOO) predictor at position $i$:
a categorical kernel $q_{\theta,i,t}(a\mid x_{-i})$ invariant
to every change of $x_i$.
Given samples of the conditional path,
we minimize
\begin{equation}
  \mathcal L_{\mathrm{LOO}}(\theta)
  =\E_{t,Y,X_t}\left[-\sum_{i=1}^L
       \log q_{\theta,i,t}(Y_i\mid X_{-i,t})\right].
  \label{eq:loo-loss}
\end{equation}
Equation~\eqref{eq:general-last-reset-law} gives their law for the chosen
reset schedule.

Over all measurable categorical kernels, the minimizers of
\eqref{eq:loo-loss} are versions of
\begin{equation}
  q_{i,t}(a\mid x_{-i})
  =\mathbb P(Y_i=a\mid X_{-i,t}=x_{-i})
  \label{eq:loo-target}
\end{equation}
In general,
$q_{i,t}(\cdot\mid x_{-i})\ne p_{i,t}(\cdot\mid x)$.

\subsection{Preserving the marginal path}

\begin{remark}
\label{rem:full-state-drift}
Replacing $p_{i,t}(\cdot\mid x)$ by $q_{i,t}(\cdot\mid x_{-i})$ changes the bridge drift
by
\[
 b_{i,t}^{\mathrm{LOO}}(x)-b_{i,t}(x)
 =\frac{\E[e(Y_i)\mid X_{-i,t}=x_{-i}]
             -\E[e(Y_i)\mid X_t=x]}{1-t}.
\]
This substitution does not in general preserve the marginal path, even
if the jump probabilities are left unchanged. In
Example~\ref{ex:loo-drift-failure} in Appendix~\ref{app:example-proof},
the jump rate kernel is unchanged, but the second-moment equation for
$X_{1,t}$ fails. Thus the discrepancy is caused by the drift alone.
\end{remark}

Keeping the drift $b_{i,t}$ while using $q_{i,t}$ for  the jump part
preserves $p_t$, although the resulting jump rate kernel differs from that
obtained by generator matching. The trajectory laws can differ while the
one-time laws remain $p_t$.
\begin{proposition}[Drift $b_{i,t}$ and LOO jumps]
\label{prop:contextual-generator}
For the conditional path \eqref{eq:conditional-path} with any rate
$\lambda(t)$ satisfying the conditions above, consider the generator
\begin{align}
  \mathcal A_t^{\mathrm{ctx}}f(x)
  ={}&\sum_i\left[
    \frac{\sum_a p_{i,t}(a\mid x)e(a)-x_i}{1-t}
       \cdot\nabla_i f(x)+\frac{\sigma_d^2}{2}\Delta_i f(x)\right]
       \nonumber\\
   &+\lambda(t)\sum_{i,a}q_{i,t}(a\mid x_{-i})
       \left[f(x^{i\leftarrow e(a)})-f(x)\right]
  \label{eq:contextual-generator}
\end{align}
Under the assumptions stated in
Appendix~\ref{app:proof-contextual-generator}, the resulting process
started from $p_0$ has law $p_t$ for every $t<1$. Moreover,
\[
 p_t\Rightarrow\mathcal L(E(Y))\qquad\text{as }t\uparrow1.
\]
\end{proposition}
The precise statement and proof are given in
Appendix~\ref{app:proof-contextual-generator}.

\subsection{Input-isolated contextual prediction}

We adapt the attention construction of T-TA \citep{shin2020tta} to noisy
embeddings and path time. Each position has a source feature
$s_j=\phi(x_j,j,t)$ depending only on its own input, position, and time.
The sources are fixed across layers. A query $h_i^{(0)}=\psi(i,t)$ has no
state input and is updated by
\begin{equation}
 h_i^{(\ell+1)}
 =F_\ell\!\left(h_i^{(\ell)},
       \operatorname{Attn}_\ell
       (h_i^{(\ell)},\{s_j:j\ne i\},t),t\right),
 \qquad q_{\theta,i,t}=\softmax(W_qh_i^{(D)}+b_q),
 \label{eq:architecture}
\end{equation}
Here $F_\ell$ acts separately at each position; its residual and
feed-forward structure is specified in
Appendix~\ref{app:input-isolated-details}.
Evolving query features never supply source keys or values.

For fixed parameters, $q_{\theta,i,t}$ in \eqref{eq:architecture} is invariant
by construction to changes of $x_i$, so
$\partial q_{\theta,i,t}/\partial x_i=0$ wherever differentiable.

Diagonal masking alone in ordinary self-attention does not ensure this
property: another position can receive $x_i$ and return it in a later layer.
Fixed token-local sources prevent that route; SwinIA uses the same
principle for blind-spot image denoising \citep{papkov2025swinia}.
All $L$ predictions are
computed in one forward pass, with $O(L^2)$ attention cost per layer at fixed
width. Our unmasked control uses the same fixed-source architecture and
allows the additional source $s_i$.
Another possibility is to target invariance to $x_i$ through the loss:
for uniform discrete diffusion, \citet[Section~3.2]{gourevitch2026loo} give
objectives whose expected loss is minimized over unrestricted predictors by the LOO posterior,
even when the predictor receives $X_t$. This invariance need not
hold after finite training \citep[Appendix~H]{gourevitch2026loo}.

\subsection{Recovering the posterior for the drift}

Proposition~\ref{prop:contextual-generator} justifies LOO reset probabilities.
To obtain the posterior $p_{i,t}(a\mid x)$ required by the continuous drift,
we apply analytic likelihood completion to the LOO predictions.
For token-factorizing discrete corruption, the analogous cavity-to-denoiser
conversion is given by \citet[Eq.~(42)]{casado2026discrete}.

The conditional paths 
\eqref{eq:conditional-path} split into products i.e. $p_t(dx\mid y)=\prod_{i}p_{i,t}(dx_i\mid y_i)$. Let $\mu_{i,t}$ be such that $p_{i,t}(\cdot\mid a)\ll \mu_{i,t}$ is absolutely continuous for every $a$ and set
$\ell_{i,t}(x_i\mid a)=dp_{i,t}(\cdot\mid a)/d\mu_{i,t}(x_i)$.
For example, $\mu_{i,t}=\sum_ap_{i,t}(\cdot\mid a)$ is a valid choice.

\begin{proposition}[Likelihood completion]
\label{prop:completion}
For conditional paths as in \eqref{eq:conditional-path}, $p_{i,t}$ satisfies
\begin{equation}
 p_{i,t}(a\mid x)
 =\frac{q_{i,t}(a\mid x_{-i})\ell_{i,t}(x_i\mid a)}
        {\sum_b q_{i,t}(b\mid x_{-i})\ell_{i,t}(x_i\mid b)}
 \label{eq:completion}
\end{equation}
for $p_t$-almost every $x$ where the denominator is positive.
\end{proposition}
The proof is given in Appendix~\ref{app:proof-completion}.
Changing $\lambda$ changes the training law and its local likelihood;
the learned posteriors and likelihood completion must use that same law.

We insert the completed probabilities only into the drift of
\eqref{eq:contextual-generator}. The reset term continues to use
$q_{\theta,i,t}$, so a single network supplies both prediction roles.
Algorithms~\ref{alg:rjd-training} and~\ref{alg:rjd-sampling}
in Appendix~\ref{app:sampler} summarize simulation-free training
and numerical generation.
An alternative with a learned drift predictor is described in
Appendix~\ref{app:separate-drift-head}.

\begin{remark}[Learning dependence between positions]
\label{rem:parity-learning}
\textcolor{black}{Consider the dataset composed by $\{(y_1,y_2,y_3,y_4):y_i\in \{0,1\} \text{ and }\sum y_i\text{ is even}\}$. Each position is individually uniform, but determined by the other three. After
250 updates, $p_{\theta,i,t}(a\mid x)$ and a predictor using only $x_i$
have nearly identical low losses, yet about $50\%$ of their generated
strings violate parity---the rate for independent entries
(Table~\ref{tab:local-shortcut}). Using $q_{\theta,i,t}(a\mid x_{-i})$
with Bayesian-completed drift via \eqref{eq:completion} produces no invalid
strings in the evaluated samples. Its higher loss is not directly
comparable to that of $p_{\theta,i,t}$ because it excludes $x_i$.
By 3,000 updates,
$p_{\theta,i,t}(a\mid x)$ also has zero observed parity violations, so
these results concern learning speed. Appendix~\ref{app:parity-learning}
gives the setup.}
\end{remark}
\vspace{-1em}
\begin{table}[ht]
\centering
\caption{Cross-entropy loss (CE) and percentage of invalid generated strings
at step 250 on the four-token even-parity task. Results average three seeds,
with 8,192 generated strings per seed. }
\label{tab:local-shortcut}
\begin{tabular}{lcc}
\toprule
Prediction target / method & Evaluation CE & Invalid strings \\
\midrule
$p_{i,t}(\cdot\mid X_t)$ & 0.0876 & $49.1\%$ \\
$\mathbb P(Y_i\in\cdot\mid X_{i,t})$ & 0.0876 & $49.5\%$ \\
$q_{i,t}(\cdot\mid X_{-i,t})$ + analytic drift & 0.1677 & $0\%$ observed \\
Masked flow matching & 0.6286 & $1.2\%$ \\
\bottomrule
\end{tabular}
\end{table}

\section{Numerical experiments}
\label{sec:experiments}

\begingroup
\color{black}
We compare unrestricted, raw LOO, and Bayesian-completed predictions on
Text8 \citep{mahoney2006text8}, LM1B \citep{chelba2014billion}, and
ZINC250k \citep{sterling2015zinc15,jo2022gdss}. Local-state replacements
and subject--verb interventions test reliance on local evidence,
contextual prediction, and recovery. Metrics and protocols are in
Appendices~\ref{app:metrics}, \ref{app:numerical-details}, \ref{app:lm1b},
and~\ref{app:zinc-details}.

\paragraph{Prediction variants.}
\label{sec:training}
Repeated-reset jump diffusion (RJD) follows \eqref{eq:conditional-path};
pure diffusion (PD) uses its continuous bridge without jumps, while
absorbing jump diffusion (AJD) stops each coordinate at its first reset
(Appendix~\ref{app:baselines}).

\begin{center}
\small
\setlength{\tabcolsep}{5pt}
\begin{tabular}{@{}llcc@{}}
\toprule
RJD/AJD methods & PD method & Drift (all methods) & Resets (RJD/AJD) \\
\midrule
RJD, AJD & PD & $p_{\theta,i,t}(\cdot\mid x)$ & $p_{\theta,i,t}(\cdot\mid x)$ \\
RJD-LOO, AJD-LOO & PD-LOO & $q_{\theta,i,t}(\cdot\mid x_{-i})$ & $q_{\theta,i,t}(\cdot\mid x_{-i})$ \\
RJD-BC-D, AJD-BC-D & PD-BC & BC of $q_{\theta,i,t}(\cdot\mid x_{-i})$ & $q_{\theta,i,t}(\cdot\mid x_{-i})$ \\
\bottomrule
\end{tabular}
\end{center}
BC denotes Bayesian completion via \eqref{eq:completion} and reintroduces $x_i$.\endgroup

\subsection{Text8 experiments}
\label{sec:text8-experiments}

\textbf{\textcolor{black}{Repeated resets.}}
RJD-LOO improves all four metrics over unrestricted prediction
(Table~\ref{tab:text8-paths}). Compared with completion, raw LOO drift
gives higher validity and lower $\operatorname{JSD}_4$, while completion
gives higher $D_5$ and $E_5$. Each ordering holds in all three seeds.\\
\textbf{\textcolor{black}{Pure diffusion.}}
For pure diffusion, LOO improves all four metrics over both unrestricted
prediction and completion. PD-LOO has the highest validity; RJD-LOO has
higher $D_5$ and $E_5$ and lower $\operatorname{JSD}_4$ than PD-LOO.

\paragraph{Comparison methods.}
\label{sec:text8-baselines}

\begingroup
\color{black}
Our comparison methods are masked discrete flow matching (DFM)
\citep{gat2024dfm} and embedded pure jumps (EPJ), with the same
backbone after the input encoder (Appendix~\ref{app:baselines}).

DFM and EPJ give similar results;
DFM has the lowest mean $\operatorname{JSD}_4$ despite much lower validity.
Thus four-gram agreement alone does not capture lexical quality.
Additional AJD results are in Appendix~\ref{app:text8-absorbing}.
\endgroup

\begin{table}[htbp]
\centering
\small
\setlength{\tabcolsep}{4pt}
\caption{\textcolor{black}{Text8 at 128 NFE after 60,000 training updates.
4,096 samples per seed, mean $\pm$ sample standard deviation
over three training seeds. Bold marks the best mean.
The protocol is in Appendix~\ref{app:numerical-details};
metrics are in Appendix~\ref{app:metrics}.}}
\label{tab:text8-paths}
\begin{tabular}{lcccc}
\toprule
Method & $V_5$ (\%) $\uparrow$ & $D_5$ $\uparrow$ & $E_5$ $\uparrow$ & $\operatorname{JSD}_4$ $\downarrow$ \\
\midrule
\multicolumn{5}{l}{Repeated resets: our variants and unrestricted control} \\
RJD & $28.96\pm 0.62$ & $3051\pm 43$ & $911\pm 21$ & $0.0657\pm 0.0007$ \\
RJD-LOO & $75.60\pm 0.89$ & $6596\pm 66$ & $2411\pm 48$ & $0.0580\pm 0.0011$ \\
RJD-BC-D & $60.16\pm 1.18$ & $\mathbf{6783}\pm 13$ & $\mathbf{2843}\pm 54$ & $0.0672\pm 0.0015$ \\
\midrule
\multicolumn{5}{l}{Pure diffusion: our variants and unrestricted control} \\
PD & $29.47\pm 1.49$ & $3137\pm 191$ & $964\pm 83$ & $0.0664\pm 0.0005$ \\
PD-LOO & $\mathbf{85.49}\pm 0.67$ & $5696\pm 58$ & $1944\pm 36$ & $0.0653\pm 0.0012$ \\
PD-BC & $36.34\pm 0.51$ & $4499\pm 7$ & $1637\pm 23$ & $0.0691\pm 0.0004$ \\
\midrule
\multicolumn{5}{l}{Comparison methods} \\
DFM & $44.62\pm 0.56$ & $4798\pm 159$ & $1501\pm 150$ & $\mathbf{0.0536}\pm 0.0002$ \\
EPJ & $44.96\pm 0.47$ & $4971\pm 75$ & $1703\pm 73$ & $0.0540\pm 0.0002$ \\
\bottomrule
\end{tabular}
\end{table}

\begingroup
\color{black}
Sampling and budget ablations are in
Appendices~\ref{app:text8-jump-sampling} and~\ref{app:text8-budgets};
Appendix~\ref{app:cadd-shared} reports a CADD-derived comparison.
\endgroup

\begingroup
\color{black}
\subsection{LM1B experiments}
\label{sec:lm1b-main}

We use length-128 LM1B windows with the BERT uncased tokenizer.
Figure~\ref{fig:lm1b-topp} plots GPT-2-large generative perplexity
against within-window token entropy
\citep{gat2024dfm,pynadath2026frontiers} across a top-$p$ sweep
\citep{holtzman2020degeneration}, which retains the smallest set of highest-probability tokens with cumulative probability at least $p$. Top-$p$ applies to jump
destinations for RJD/EPJ and drift probabilities for PD.
Appendix~\ref{app:lm1b-uniform} describes the sampling protocol and the approximation used for Bayesian completion.

Within the entropy range of natural text, curves closer to the lower
right indicate a better perplexity--diversity trade-off: lower PPL
without sacrificing within-window token diversity. Low PPL alone
can reflect limited diversity. For reference,
\citet{chemseddine2026spherical} report natural LM1B entropy of
$H\approx4.34$ nats.

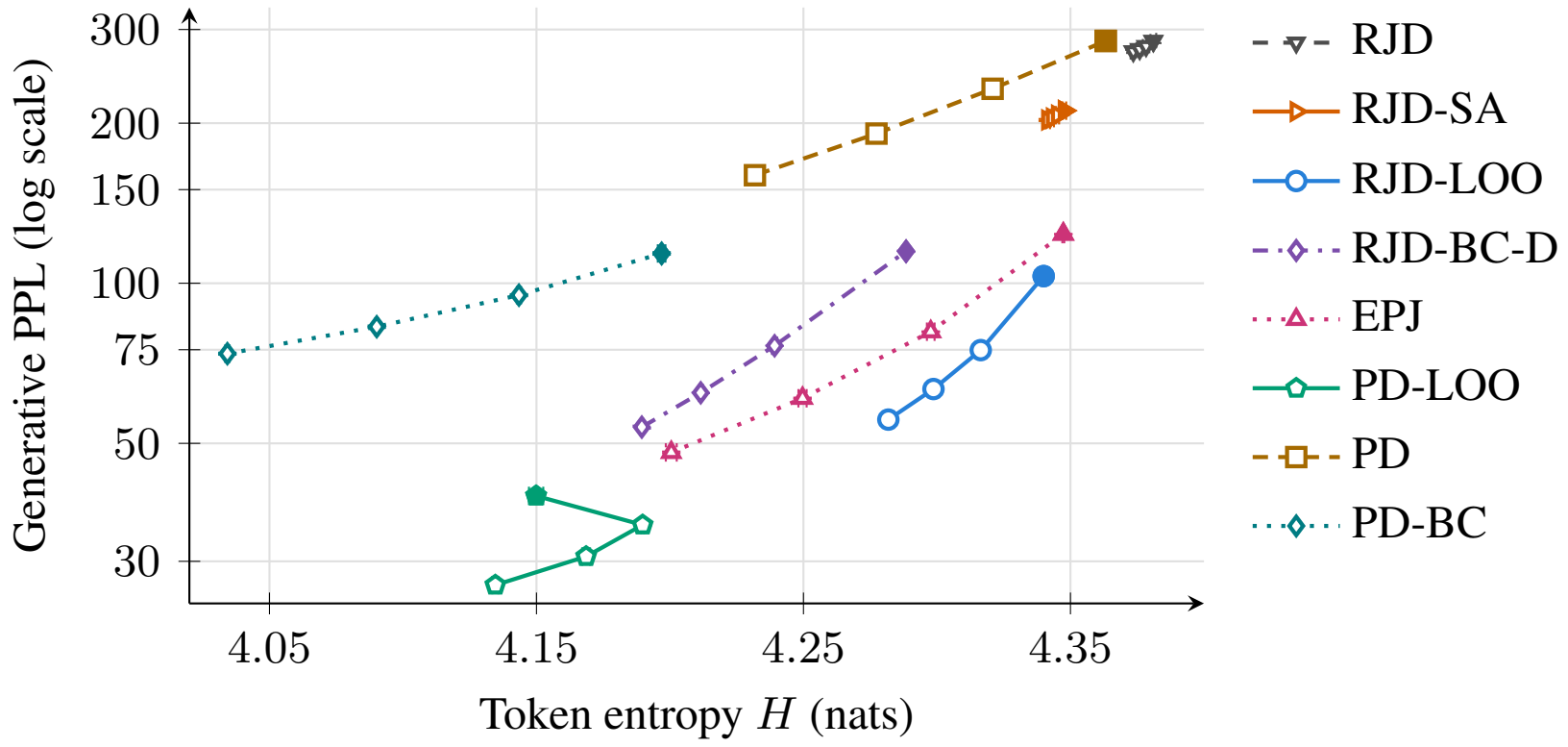
\begin{figure}[htbp]
\color{black}
\centering
\definecolor{lmonebrjd}{HTML}{4D4D4D}
\definecolor{lmonebrjdsa}{HTML}{D55E00}
\definecolor{lmonebloo}{HTML}{2680D9}
\definecolor{lmonebepj}{HTML}{CC3377}
\definecolor{lmonebrjdbc}{HTML}{7C4DAB}
\definecolor{lmonebpdloo}{HTML}{009E73}
\definecolor{lmonebpd}{HTML}{A56A00}
\definecolor{lmonebpdbc}{HTML}{007C83}
\begin{tikzpicture}[color=black]
\begin{axis}[
  width=0.50\linewidth, height=4.1cm, scale only axis,
  /tikz/font=\footnotesize,
  xlabel={Token entropy $H$ (nats)},
  ylabel={Generative PPL (log scale)},
  xmin=4.02, xmax=4.40, xtick={4.05,4.15,4.25,4.35},
  label style={text=black}, tick label style={text=black},
  axis lines=left, axis line style={black}, tick style={black},
  ymode=log, log ticks with fixed point,
  ymin=25, ymax=330, ytick={30,50,75,100,150,200,300},
  minor tick num=0, minor y tick num=0,
  grid=major, grid style={black!12},
  xticklabel style={/pgf/number format/fixed,
    /pgf/number format/precision=2},
  legend style={draw=none, fill=none, text=black,
    font=\footnotesize, cells={anchor=west}, inner sep=1pt,
    row sep=2pt, at={(1.04,1)}, anchor=north west},
  every axis plot/.append style={line width=0.8pt, mark size=1.8pt,
    mark options={solid}},
]
\addplot[lmonebrjd, dashed, mark=triangle*, mark size=1.6pt,
  mark options={solid, rotate=180, fill=white, draw=lmonebrjd},
  error bars/.cd, x dir=both, x explicit, y dir=both, y explicit]
coordinates {
  (4.373540930,272.945045328) +- (0.000988969,1.906486718)
  (4.375932109,275.549691474) +- (0.001198004,2.008004323)
  (4.378330982,279.188563869) +- (0.001211062,1.749537248)
  (4.381048470,285.350927383) +- (0.001074687,2.438896256)
};
\addlegendentry{RJD}
\addplot[lmonebrjd, only marks, mark=triangle*, mark size=1.6pt, forget plot,
  mark options={solid, rotate=180, fill=lmonebrjd, draw=lmonebrjd}]
coordinates {(4.381048470,285.350927383)};
\addplot[lmonebrjdsa, solid, mark=triangle*, mark size=1.8pt,
  mark options={solid, rotate=270, fill=white, draw=lmonebrjdsa},
  error bars/.cd, x dir=both, x explicit, y dir=both, y explicit]
coordinates {
  (4.341489551,202.770493518) +- (0.000857262,0.889016270)
  (4.343247693,204.442971066) +- (0.000567551,0.850943613)
  (4.345172457,207.021621965) +- (0.000565157,0.993851340)
  (4.347738327,211.128981549) +- (0.000668073,1.022301908)
};
\addlegendentry{RJD-SA}
\addplot[lmonebrjdsa, only marks, mark=triangle*, mark size=1.8pt, forget plot,
  mark options={solid, rotate=270, fill=lmonebrjdsa, draw=lmonebrjdsa}]
coordinates {(4.347738327,211.128981549)};
\addplot[lmonebloo, solid, mark=*,
  mark options={solid, fill=white, draw=lmonebloo},
  error bars/.cd, x dir=both, x explicit, y dir=both, y explicit]
coordinates {
  (4.281792471,55.412981512) +- (0.001353081,0.287884175)
  (4.298694706,63.254965791) +- (0.000183954,0.293428063)
  (4.316402917,74.872812757) +- (0.000089545,0.292351132)
  (4.339964037,103.142263629) +- (0.000115690,0.514247249)
};
\addlegendentry{RJD-LOO}
\addplot[lmonebloo, only marks, mark=*, forget plot,
  mark options={solid, fill=lmonebloo, draw=lmonebloo}]
coordinates {(4.339964037,103.142263629)};
\addplot[lmonebrjdbc, dashdotted, mark=diamond*,
  mark options={solid, fill=white, draw=lmonebrjdbc},
  error bars/.cd, x dir=both, x explicit, y dir=both, y explicit]
coordinates {
  (4.189453431,53.645499383) +- (0.000579905,0.341997114)
  (4.211524451,62.245282959) +- (0.000735093,0.095344986)
  (4.239190332,76.306755234) +- (0.000987293,0.065989540)
  (4.288432082,114.849816637) +- (0.000414240,0.301896540)
};
\addlegendentry{RJD-BC-D}
\addplot[lmonebrjdbc, only marks, mark=diamond*, forget plot,
  mark options={solid, fill=lmonebrjdbc, draw=lmonebrjdbc}]
coordinates {(4.288432082,114.849816637)};
\addplot[lmonebepj, dotted, mark=triangle*,
  mark options={solid, fill=white, draw=lmonebepj},
  error bars/.cd, x dir=both, x explicit, y dir=both, y explicit]
coordinates {
  (4.200492712,48.074665759) +- (0.002043117,0.180654428)
  (4.249653643,60.779593232) +- (0.001536462,0.169274805)
  (4.297677749,81.080326596) +- (0.001434440,0.117896976)
  (4.347326630,123.708378054) +- (0.000354833,0.042079438)
};
\addlegendentry{EPJ}
\addplot[lmonebepj, only marks, mark=triangle*, forget plot,
  mark options={solid, fill=lmonebepj, draw=lmonebepj}]
coordinates {(4.347326630,123.708378054)};
\addplot[lmonebpdloo, solid, mark=pentagon*,
  mark options={solid, fill=white, draw=lmonebpdloo},
  error bars/.cd, x dir=both, x explicit, y dir=both, y explicit]
coordinates {
  (4.134598859,27.068484260) +- (0.001151992,0.059604837)
  (4.168597919,30.634685048) +- (0.001467400,0.102964288)
  (4.189763397,35.123121790) +- (0.000537638,0.096346632)
  (4.149869274,39.823615903) +- (0.002679637,0.147447424)
};
\addlegendentry{PD-LOO}
\addplot[lmonebpdloo, only marks, mark=pentagon*, forget plot,
  mark options={solid, fill=lmonebpdloo, draw=lmonebpdloo}]
coordinates {(4.149869274,39.823615903)};
\addplot[lmonebpd, densely dashed, mark=square*,
  mark options={solid, fill=white, draw=lmonebpd},
  error bars/.cd, x dir=both, x explicit, y dir=both, y explicit]
coordinates {
  (4.231823441,159.635965964) +- (0.000402824,0.136763219)
  (4.277340883,191.182269212) +- (0.000495806,0.505736765)
  (4.320848081,232.099187192) +- (0.001548929,0.972276337)
  (4.363203351,285.575962862) +- (0.000975389,0.446543903)
};
\addlegendentry{PD}
\addplot[lmonebpd, only marks, mark=square*, forget plot,
  mark options={solid, fill=lmonebpd, draw=lmonebpd}]
coordinates {(4.363203351,285.575962862)};
\addplot[lmonebpdbc, dotted, mark=diamond*,
  mark options={solid, fill=white, draw=lmonebpdbc},
  error bars/.cd, x dir=both, x explicit, y dir=both, y explicit]
coordinates {
  (4.034233602,73.726640203) +- (0.000833710,0.327327264)
  (4.090171603,82.845383999) +- (0.000614796,0.433811378)
  (4.143438905,94.973816812) +- (0.000769583,0.424449261)
  (4.196877600,113.768854024) +- (0.001425912,0.640775133)
};
\addlegendentry{PD-BC}
\addplot[lmonebpdbc, only marks, mark=diamond*, forget plot,
  mark options={solid, fill=lmonebpdbc, draw=lmonebpdbc}]
coordinates {(4.196877600,113.768854024)};
\end{axis}
\end{tikzpicture}
\caption{LM1B at 750k updates and 128 NFE on a uniform time grid.
 Curves connect all $p\in\{0.7,0.8,0.9,1\}$
 in $p$ order; filled markers denote $p=1$. Points show means over
 three generation seeds (5,120 windows each); error bars indicate
 $\pm1$ sample standard deviation and are mostly smaller than the markers.
 All models use one training seed. RJD-SA uses standard self-attention.}
\label{fig:lm1b-topp}
\end{figure}

MAUVE \citep{pillutla2021mauve} compares generated text with test
LM1B text, with higher scores indicating closer agreement.
Table~\ref{tab:lm1b-best-mauve} reports the best mean score over the
same top-$p$ sweep; Appendix~\ref{app:lm1b-metrics} defines the metrics.
RJD-LOO and EPJ attain similar best MAUVE (0.9039 and 0.8965).
LOO achieves higher best MAUVE
than unrestricted prediction, both with and without Bayesian completion.
Bayesian completion gives lower best MAUVE than raw LOO in both RJD
and PD. Appendix~\ref{app:lm1b-uniform} discusses the full top-$p$ sweep 
and NFE comparisons are in Appendix~\ref{app:lm1b-generation}.

\begin{table}[!htb]
\color{black}
\centering
\small
\caption{Best observed MAUVE over $p\in\{0.7,0.8,0.9,1\}$ for each
method in Figure~\ref{fig:lm1b-topp}. Entries are mean $\pm$ sample SD
over three generation seeds.}
\label{tab:lm1b-best-mauve}
\begin{tabular}{lrr}
\toprule
Method & Best $p$ & MAUVE $\uparrow$ \\
\midrule
RJD & 0.7 & $0.2019\pm0.0032$ \\
RJD-SA & 0.7 & $0.4704\pm0.0192$ \\
RJD-LOO & 0.8 & $0.9039\pm0.0058$ \\
RJD-BC-D & 0.8 & $0.7624\pm0.0126$ \\
EPJ & 0.9 & $0.8965\pm0.0103$ \\
\midrule
PD-LOO & 0.8 & $0.5970\pm0.0089$ \\
PD & 0.7 & $0.2111\pm0.0138$ \\
PD-BC & 1 & $0.3576\pm0.0071$ \\
\bottomrule
\end{tabular}
\end{table}

\textbf{Self-attention control.}
RJD-SA replaces fixed-source attention with standard self-attention,
computing keys and values from evolving hidden states at each layer.
It uses unrestricted predictions with the same RJD path, depth, width,
and training budget (Appendix~\ref{app:lm1b-self-attention}).
Its best MAUVE is 0.4704, compared with 0.2019 for RJD
(Table~\ref{tab:lm1b-best-mauve}). This suggests that fixed-source attention
limits performance. RJD-LOO nevertheless reaches 0.9039, motivating
more expressive architectures that preserve LOO invariance.
\endgroup

\FloatBarrier
\subsection{ZINC250k experiments}
\label{sec:zinc-main}

\begingroup
\color{black}
On ZINC250k \citep{sterling2015zinc15}, we compare unrestricted EPJ
with unrestricted and LOO predictors for PD and AJD.
All five models receive 40,000 updates for each of two training
seeds; Table~\ref{tab:zinc-main} evaluates them at 128 NFE.
Appendix~\ref{app:zinc-details} gives the representation, architecture,
and evaluation protocol.
\endgroup
The two-seed results show that the effect of raw LOO drift depends
on the conditional path. For AJD, raw LOO drift increases the fraction
of valid and connected outputs over unrestricted prediction and achieves
the lowest mean raw and corrected FCD. Completing its drift reduces connected
validity but improves NSPDK agreement. For PD, completion improves raw
validity, connected validity, raw FCD, and NSPDK over raw LOO drift;
corrected FCD improves only slightly on average. PD-BC has the highest
mean raw validity, whereas AJD-LOO has the highest mean fraction of valid
and connected molecules. 
\begin{table}[htbp]
\centering
\small
\setlength{\tabcolsep}{3.8pt}
\caption{\textcolor{black}{ZINC250k test results for 128 NFE:
mean $\pm$ sample standard deviation over training seeds 0 and 1,
with 10,000 generated graphs per seed.
Validity percentages use all generated graphs before correction. Raw
distributional metrics include valid disconnected outputs; corrected FCD
uses bond repair and largest-component selection. NSPDK is squared MMD
(Appendix~\ref{app:zinc-metrics}).}}
\label{tab:zinc-main}
\begin{tabular}{lrrrrr}
\toprule
Method & \shortstack{Raw valid\\(\%) $\uparrow$} & \shortstack{Raw valid \&\\connected (\%) $\uparrow$} & \shortstack{Raw FCD\\$\downarrow$} & \shortstack{Corrected FCD\\$\downarrow$} & \shortstack{Raw NSPDK\\$\downarrow$} \\
\midrule
EPJ & $69.02\pm 0.22$ & $41.81\pm 0.11$ & $21.190\pm 0.039$ & $21.305\pm 0.002$ & $0.04882\pm 0.00088$ \\
PD & $89.98\pm 0.33$ & $49.97\pm 0.37$ & $21.919\pm 0.031$ & $21.740\pm 0.003$ & $0.05154\pm 0.00015$ \\
PD-LOO & $55.38\pm 5.42$ & $20.07\pm 3.91$ & $24.966\pm 1.164$ & $21.670\pm 1.250$ & $0.15202\pm 0.02161$ \\
PD-BC & $\mathbf{91.84}\pm 0.40$ & $54.19\pm 2.21$ & $21.642\pm 0.066$ & $21.535\pm 0.090$ & $0.05149\pm 0.00088$ \\
AJD & $87.41\pm 0.95$ & $47.82\pm 2.49$ & $22.065\pm 0.088$ & $21.943\pm 0.082$ & $0.05054\pm 0.00043$ \\
AJD-LOO & $83.96\pm 0.52$ & $\mathbf{71.07}\pm 0.61$ & $\mathbf{20.460}\pm 0.019$ & $\mathbf{20.968}\pm 0.015$ & $0.05141\pm 0.00011$ \\
AJD-BC-D & $78.53\pm 0.16$ & $59.11\pm 0.38$ & $20.677\pm 0.090$ & $20.990\pm 0.053$ & $\mathbf{0.04629}\pm 0.00003$ \\
\bottomrule
\end{tabular}
\end{table}

\FloatBarrier

\begingroup
\color{black}
\subsection{Testing Contextual Learning}
\label{sec:interventions}

\textbf{Local shortcuts.}
We first investigate whether the model learns a shortcut: predicting the
token at position $i$ from its own noisy input $X_{i,t}$ without relying
on relations between positions. The intervention tests whether conflicting
local evidence at $X_{i,t}$ overrides context and persists during continued generation.
For an LM1B sequence $y=(y_1,\ldots,y_L)$, draw $x$ at time $t$ from the
conditional process \eqref{eq:conditional-path}. Replace only $x_i$
by the state $x_i'$ for a donor $d\ne y_i$, using the
same Gaussian residual and reset times, then continue all positions freely.
At $t=0.50$, unrestricted RJD and RJD-SA almost always retain
$d$, whereas LOO more often recovers $y_i$ but also changes
correct inputs (Table~\ref{tab:intervention-local}). These interventions
show that unrestricted predictions can remain dominated by local evidence
despite conflicting context. The start-time sweep is in
Appendix~\ref{app:lm1b-recovery}.
\begin{table}[htbp]
\color{black}
\centering
\small
\caption{LM1B local-state intervention at $t=0.50$ and 750k updates
(one training seed/model), with 128 further calls on a uniform time grid.
Reset top-$p$ is 0.8 for RJD-LOO and 0.7 for RJD and RJD-SA.
Percentages use 1,024 text pairs with two draws each.
Protocol and start-time sweep: Appendix~\ref{app:lm1b-recovery}.}
\label{tab:intervention-local}
\begin{tabular}{lrrr}
\toprule
Final token (\%) & RJD-LOO & RJD & RJD-SA \\
\midrule
Original $y_i$, after replacement & 30.81 & 0.15 & 0.00 \\
Donor $d$, after replacement & 6.01 & 99.71 & 99.61 \\
Original $y_i$, without replacement & 50.88 & 100.00 & 100.00 \\
\bottomrule
\end{tabular}
\end{table}

\textbf{Bayesian completion and recovery.}
To test whether completion can hinder recovery from misleading local
evidence, we replace a Text8 character state by an independently corrupted
state for a donor $d\ne y_i$ and compare raw and completed drift using the
same LOO checkpoint. Equation~\eqref{eq:completion} explains the immediate effect: replacing only
the local state leaves the LOO prediction unchanged, while completion
multiplies its odds of donor versus original by the local likelihood ratio.
Results are in Table~\ref{tab:text8-intervention-main}: PD-BC retains the donor almost
always, whereas RJD-BC-D can replace it through resets; completion lowers
the final word-check pass rate in both families. Earlier-time results and
controls are in Appendix~\ref{app:text8-interventions}.

\begin{table}[!htb]
\color{black}
\centering
\small
\setlength{\tabcolsep}{3.5pt}
\caption{Text8 donor replacement at $t=0.6973$. Percentages are mean $\pm$ sample SD over three
training seeds (957 character replacements per seed). The local word check accepts
corpus-attested alternatives and retains unresolved cases in the denominator.
Immediate argmax refers to the probabilities used in the drift,
before continued generation.}
\label{tab:text8-intervention-main}
\begin{tabular}{lrrrr}
\toprule
Outcome & PD-LOO & PD-BC & RJD-LOO & RJD-BC-D \\
\midrule
Immediate argmax is $d$ & $0.21\pm0.10$ & $99.79\pm0.10$ & $0.07\pm0.06$ & $100.00\pm0.00$ \\
Final character is $d$ & $8.50\pm0.59$ & $99.83\pm0.06$ & $15.57\pm1.31$ & $20.72\pm2.28$ \\
Final local word check passes & $88.89\pm0.54$ & $9.40\pm0.10$ & $75.51\pm1.46$ & $63.22\pm0.79$ \\
\bottomrule
\end{tabular}
\end{table}

\textbf{LM1B subject--verb agreement.}
To test whether contextual predictions capture grammatical dependencies,
we change the subject's number while holding the local verb state and
all other noisy inputs fixed. For example, changing ``the key near the
cabinets'' to ``the keys near the cabinets'' should switch the preferred
verb from ``is'' to ``are''. We measure whether the preferred form is
correct in both versions, using the two number variants rather than the
full-vocabulary argmax. Figure~\ref{fig:lm1b-agreement-distance} shows
that LOO tracks the subject more often under conflicting local evidence,
but performance deteriorates on longer and harder templates. Distance
also varies with wording and intervening nouns; Appendix~\ref{app:lm1b-context}
separates these configurations and reports all four tested times.

\begin{figure}[htbp]
\centering
{\color{black}\includegraphics[width=\linewidth]{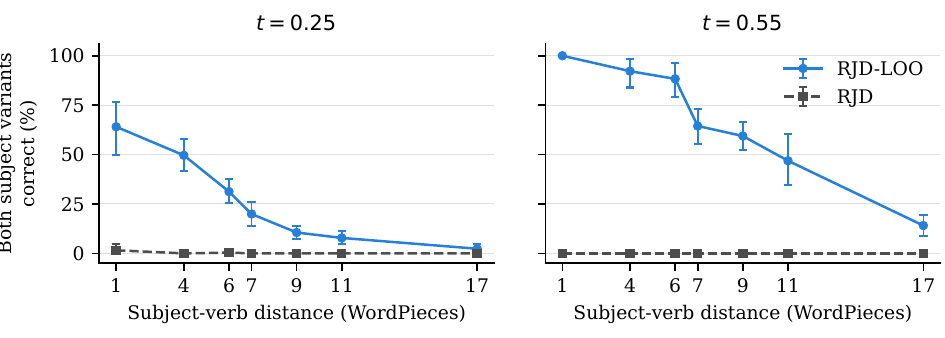}}
\color{black}
\caption{Paired subject tracking with LM1B models at 750k updates. Success requires the
correct verb-form preference before and after changing only the subject;
the local verb state is initially correct and held fixed. Error bars are
95\% lexical-cluster bootstrap intervals. Each distance uses 128 templates
with two noise draws, except distance one (32 templates). Distances count
WordPieces; wording and intervening-noun configurations also vary.}
\label{fig:lm1b-agreement-distance}
\end{figure}

\textbf{Context in generated text.}
We test whether older word order helps predict later tokens.
For illustration, GPT-2-large scores ``oven'' in ``The chef mixed dough
and put it in the oven'' before and after shuffling ``chef mixed dough
and'', keeping ``put it in the'' fixed. A drop after shuffling indicates useful
older word order. On generated texts from the LM1B experiments,
comparisons favor RJD-LOO over RJD, RJD-SA, and EPJ with 4 or 8 recent
GPT-2 tokens fixed, see
Appendix~\ref{app:lm1b-blind-assessment}.
\endgroup

\FloatBarrier
\iftrue
\begingroup
\color{black}
\section{Conclusion}
\label{sec:discussion}
Across Text8, LM1B, and ZINC250k, LOO-based variants improve key
generation metrics over matched unrestricted controls, with the choice
between raw and Bayesian-completed predictions depending on the task
and sampler. In the LM1B interventions, unrestricted RJD relies heavily
on its own-position input at the tested intermediate and late times,
retaining incorrect local tokens despite conflicting context.

\textbf{Limitations and future work.}
Our fixed-source architecture is restrictive: ordinary self-attention
improves the paired, single-seed EPJ control
(Table~\ref{tab:text8-ddit-pure-jump}). LOO's existing gains over matched
unrestricted controls make more expressive architectures preserving
LOO invariance particularly promising.

Our repeated-reset sampler uses state-independent rates and full coordinate resets. Rates
and jump kernels adapted to disagreement between $x_{i,t}$ and contextual
predictions could target likely errors more selectively, but would
require new marginal-preservation conditions.

Recovery is tested from prescribed corrupted states, and LM1B uses
short sequences and one training seed. More seeds, longer contexts, and
interventions along generated trajectories would assess robustness and
how local errors affect ordinary generation. Comparisons with exact
posteriors on tractable joint distributions would help separate changes
to the probability path from learning and numerical errors.
\endgroup
\fi

\subsection*{Acknowledgments}

GENCI provided computing and storage resources on Jean Zay at IDRIS
(grant 2026-AD011017753) and on Adastra at CINES (grant 2026-AD011018512).

\begingroup
\expandafter\def\csname rebuttalbib@chelba2014billion\endcsname{}
\expandafter\def\csname rebuttalbib@devlin2019bert\endcsname{}
\expandafter\def\csname rebuttalbib@wu2016wordpiece\endcsname{}
\expandafter\def\csname rebuttalbib@radford2019language\endcsname{}
\expandafter\def\csname rebuttalbib@pillutla2021mauve\endcsname{}
\expandafter\def\csname rebuttalbib@shannon1948communication\endcsname{}
\expandafter\def\csname rebuttalbib@chemseddine2026spherical\endcsname{}
\let\rebuttalOriginalBibitem\bibitem
\renewcommand{\bibitem}[2][]{%
  \par
  \ifcsname rebuttalbib@#2\endcsname\color{black}\else\color{black}\fi
  \rebuttalOriginalBibitem[#1]{#2}}
\bibliography{references,references_rebuttle}
\endgroup
\bibliographystyle{iclr2027_conference}
\appendix

\begingroup
\color{black}
\section{Extended related work}
\label{app:extended-related-work}

\paragraph{Discrete and continuous representations.}
Masked diffusion models such as MDLM and MD4
\citep{sahoo2024mdlm,shi2024md4}, and discrete flow matching
\citep{gat2024dfm}, evolve discrete states. A masked position supplies no
target identity, so prediction uses other positions. Continuous models
diffuse embeddings, as in
Diffusion-LM \citep{li2022diffusionlm}, continuous diffusion for categorical
data \citep{dieleman2022continuous}, and Plaid \citep{gulrajani2023plaid}.
LangFlow and RePlaid report competitive language-modeling performance
\citep{chen2026langflow,yang2026replaid}. Our motivation concerns training
information: an identifiable noisy embedding can permit accurate local
prediction before the model learns dependencies needed for joint generation.
\paragraph{Hybrid processes and repeated resets.}
CADD, CANDI, and CCDD combine discrete states with continuous
representations \citep{zheng2026cadd,pynadath2026candi,zhou2026ccdd}.
CANDI highlights a relevant tension: preserving continuous context can
leave denoising too easy, while stronger corruption can remove that context
\citep{pynadath2026candi}. Sticky Jump Diffusions combine continuous evolution
with absorbing jumps to token anchors and permit cross-position blending
in the corruption process \citep{jutrasdube2026sticky}.
Our endpoint-directed Brownian bridge resets coordinates completely to
their endpoint embeddings, resumes continuous motion, and permits further
resets. Its drift and reset destinations both require predictions, whose
information requirements need not coincide. The CADD-derived experiments
in Appendix~\ref{app:cadd-shared} examine input exclusion within a hybrid
training recipe.

\paragraph{Architectures that exclude local information.}
Blind-spot image denoising excludes a pixel's noisy observation from its
own prediction to enable self-supervised restoration
\citep{batson2019noise2self,laine2019blindspot}.
We adapt T-TA's input-isolated attention \citep{shin2020tta} to noisy
embeddings and path time. Related constructions appear in Hollow
Transformers for uniform discrete corruption \citep{sun2023hollow} and
informed correctors for masked diffusion \citep{zhao2024correctors}.
Input exclusion must hold across the entire network: diagonal attention
masking alone permits information to return through another position in
a later layer. Our fixed token-local sources prevent this route;
SwinIA uses the same principle for blind-spot image denoising
\citep{papkov2025swinia}. 
\paragraph{LOO posteriors in discrete diffusion.}
Input exclusion changes the population prediction target.
\citet{gourevitch2026loo} identify the LOO posterior as the target of the
plug-in ELBO for uniform diffusion and derive conversions between this
posterior, the denoiser, and the score. Their objectives can have a LOO
population minimizer without architectural input exclusion, although
finite training need not yield exact invariance.
\citet{casado2026discrete} give corresponding conversions for general
token-factorizing discrete corruption using the cavity posterior, which
excludes the local corrupted token. \citet{dmitriev2026adaptive} analyze
LOO samplers for uniform and remasking diffusion, with guarantees depending
on the target's dependence structure. We study LOO prediction when
continuous evolution occurs between coordinate resets.

\paragraph{Likelihood completion and the generator.}
Combining contextual predictions with local likelihoods has precedents
in blind-spot denoising \citep{laine2019blindspot}, Plaid's output
parameterization \citep[Section~3.3]{gulrajani2023plaid}, and the
cavity-to-denoiser conversion for discrete corruption
\citep[Eq.~(42)]{casado2026discrete}. For our token-factorizing conditional
paths, Bayes' rule likewise recovers $p_{i,t}$ from $q_{i,t}$ and
the local likelihood (Proposition~\ref{prop:completion}).
\endgroup

\section{Conditional law and analytic likelihood}
\label{app:local-law}

\subsection{Jump rate kernels and Markov generators}
\label{app:jump-terminology}

\paragraph{Jump rate kernels and destination probabilities.}
Let $S$ denote the state space. A \emph{jump rate kernel} $K_t(x,dz)$ is a
nonnegative measure in the destination $dz$ and measurable in $x$. It specifies the instantaneous rates of jump
events into subsets of $S$, rather than probabilities over a fixed time
interval \citep{conforti2022timereversal}. For the processes considered
here, its total mass $\lambda_t(x)=K_t(x,S)$ is finite at each $t<1$ and
gives the total jump intensity. When $\lambda_t(x)>0$,
$J_t(x,dz)=K_t(x,dz)/\lambda_t(x)$ is the destination probability kernel
conditional on an event. When $\lambda_t(x)=0$,
the instantaneous event rate is zero and this destination law may be
chosen arbitrarily.

For a finite-state pure-jump process with rates continuous in time, let
$N_t$ count all jump events, including self-resets. As $h\downarrow0$,
\begin{align*}
 \mathbb P(N_{t+h}-N_t=0\mid X_t=x)
   &=1-\lambda_t(x)h+o(h),\\
 \mathbb P(N_{t+h}-N_t\ge1\mid X_t=x)
   &=\lambda_t(x)h+o(h).
\end{align*}
Thus $\lambda_t(x)$ is a rate, and $\lambda_t(x)h$ approximates the
probability of an event over a short interval of length $h$.
Self-resets may be included in $K_t$; the rate of actual state changes is
then $K_t(x,S\setminus\{x\})=\lambda_t(x)-K_t(x,\{x\})$. Accordingly,
\begin{align*}
 \mathbb P(X_{t+h}=x\mid X_t=x)
   &=1-h\bigl[\lambda_t(x)-K_t(x,\{x\})\bigr]+o(h),\\
 \mathbb P(X_{t+h}=z\mid X_t=x)
   &=hK_t(x,\{z\})+o(h),\qquad z\ne x.
\end{align*}
Without self-resets, the first probability is $1-\lambda_t(x)h+o(h)$;
the second is $h\lambda_t(x)J_t(x,\{z\})+o(h)$ when $\lambda_t(x)>0$.
For a jump diffusion, no reset does not imply an unchanged state, since
the drift and Brownian motion continue between resets.
We use \emph{jump rate kernel} for $K_t$
and \emph{destination probabilities} for normalized laws such as $J_t$.

\paragraph{Discrete transitions and embedding resets.}
On a discrete state space, the masses $K_t(x,\{z\})$, $z\ne x$, are also
called transition rates or transition intensities. In the masked model,
the per-position rates in \eqref{eq:masked-flow-rates} specify
\[
 K_t(x,dz)=\sum_{i=1}^L\sum_{a\in\cV}r_{i,t}(x_i\to a\mid x)
                 \delta_{x^{i\leftarrow a}}(dz),
\]
where $\delta_z$ is the point mass at $z$, and coordinate replacement is
understood here for token sequences. For the embedding-space resets,
the kernels in \eqref{eq:averaged-generator} and
\eqref{eq:contextual-generator} are, respectively,
\begin{align*}
 K_t(x,dz)&=\lambda(t)\sum_{i,a}p_{i,t}(a\mid x)
                       \delta_{x^{i\leftarrow e(a)}}(dz),\\
 K_t^{\mathrm{ctx}}(x,dz)&=\lambda(t)\sum_{i,a}
       q_{i,t}(a\mid x_{-i})\delta_{x^{i\leftarrow e(a)}}(dz).
\end{align*}
Both have total reset intensity $\lambda_t(x)=L\lambda(t)$; $\lambda(t)$ is the
per-position clock intensity. The categorical laws $p_{i,t}$ and
$q_{i,t}$ specify destination tokens conditional on resetting
position $i$; multiplying by the clock intensity gives the corresponding
contribution to the jump rate kernel.

\paragraph{Markov generators.}
The \emph{infinitesimal generator}, or \emph{Markov generator}, describes
the instantaneous change in expected test functions
\citep[Section~4]{holderrieth2025generator}. For our jump diffusions,
\[
 \mathcal A_t f(x)
 =\sum_i\left[b_{i,t}(x)\cdot\nabla_i f(x)
              +\frac{\sigma_d^2}{2}\Delta_i f(x)\right]
   +\int_S[f(z)-f(x)]K_t(x,dz).
\]
For a pure-jump process, the continuous terms are absent. On a finite
state space, the generator matrix has off-diagonal entries
$K_t(x,\{z\})$ and diagonal entry $-K_t(x,S\setminus\{x\})$;
its rows sum to zero. Self-resets contribute zero to the generator.

\subsection{Conditional jump-diffusion SDE}
\label{app:conditional-sde}

For background on stochastic differential equations with jumps and their
generators, see \citet[Chapters~3--4]{kunita2019flows}.
We derive the conditional state law used in Remark~\ref{rem:training-states}
by conditioning on the last reset time, as in last-renewal representations
of stochastic resetting \citep{evans2020resetting}.

As in the main construction, let $\lambda(t)\ge0$ satisfy
$\int_0^t\lambda(s)\,ds<\infty$ for every $t<1$.
Fix $y\in\cV^L$. Let $W_i$ be independent $d$-dimensional Brownian motions
and $N_i$ independent inhomogeneous Poisson counting processes with
 intensity measure $\lambda(t)\,dt$.
Recall the Jump SDE from \eqref{eq:conditional-path} 
\begin{equation}
 dX_{i,t}^y
 =\frac{e(y_i)-X_{i,t-}^y}{1-t}\,dt+\sigma_d\,dW_{i,t}
   +(e(y_i)-X_{i,t-}^y)\,dN_{i,t},
 \qquad X_{i,0}^y\sim\mathcal N(0,\sigma_s^2I_d).
 \label{eq:conditional-jump-sde}
\end{equation}
We take $\sigma_d>0$ and $\sigma_s\ge0$, with a base law $\mathcal N(0,\sigma_s^2 I_d)$ independent of $y$.
Here $X_{i,t-}^y$ is the left limit. At a jump of $N_i$, the increment is
$e(y_i)-X_{i,t-}^y$, so $X_{i,t}^y=e(y_i)$ exactly. The Brownian motion
and counting process continue after this event. On each $[0,T]$, $T<1$,
there are almost surely finitely many resets, and the between-reset linear
SDE has a unique solution. Write $p_t(\cdot\mid y)=\mathcal L(X_t^y)$.

\begin{lemma}[Conditional state law]
\label{lem:conditional-state-law}
Consider \eqref{eq:conditional-jump-sde} with deterministic $\lambda$ and
mutually independent initial states, Brownian motions, and Poisson
processes. For every $0<t<1$ and $y_i=a$, the law
$p_{i,t}(\cdot\mid a)=\mathcal L(X_{i,t}^y)$ is the Gaussian mixture
\begin{align}
 p_{i,t}(dx\mid a)
 &=S(0,t)\mathcal N(dx;t e(a),v_{0,t}I_d)\nonumber\\
 &\quad+\int_0^t\lambda(r)S(r,t)
           \mathcal N(dx;e(a),v_{r,t}I_d)\,dr,
 \label{eq:general-last-reset-law}
\end{align}
where $S(r,t)=\exp(-\int_r^t\lambda(u)\,du)$,
$v_{0,t}=(1-t)^2\sigma_s^2+\sigma_d^2t(1-t)$, and
$v_{r,t}=\sigma_d^2(t-r)(1-t)/(1-r)$ for $0<r<t$.
The coordinates are conditionally
independent, so $p_t(dx\mid y)=\prod_i p_{i,t}(dx_i\mid y_i)$.
\end{lemma}

\begin{proof}
For $0\le s<t<1$, if no reset occurs during $(s,t]$, solving the linear SDE gives
\begin{equation}
 X_{i,t}^y=e(y_i)+\frac{1-t}{1-s}(X_{i,s}^y-e(y_i))
       +\sigma_d(1-t)\int_s^t\frac{dW_{i,u}}{1-u}.
 \label{eq:conditional-bridge-transition}
\end{equation}
The last term is centered Gaussian with covariance
$\sigma_d^2(t-s)(1-t)/(1-s)I_d$.
If there is no reset before $t$, set $s=0$ and use the independent
Gaussian initial state. This gives mean $t e(a)$ and covariance
$v_{0,t}I_d$, where
$v_{0,t}=(1-t)^2\sigma_s^2+\sigma_d^2t(1-t)$.
If the last reset is at $r\in(0,t)$, then $X_{i,r}^y=e(a)$.
Setting $s=r$ therefore gives mean $e(a)$ and covariance $v_{r,t}I_d$,
where $v_{r,t}=\sigma_d^2(t-r)(1-t)/(1-r)$.

It remains to weight these Gaussian laws by the reset probabilities.
Let $\tau_{i,t}$ be the last reset time of coordinate $i$ up to time $t$,
with $\tau_{i,t}=0$ if no reset has occurred. For $0\le r<t$,
\[
 \mathbb P(\tau_{i,t}\le r)
 =\mathbb P(N_{i,t}-N_{i,r}=0)
 =\exp\!\left(-\int_r^t\lambda(u)\,du\right)=S(r,t).
\]
Thus $\tau_{i,t}$ has an atom of mass $S(0,t)$ at zero and density
$\partial_r S(r,t)=\lambda(r)S(r,t)$ for almost every $r\in(0,t)$.
Independence of the Poisson process from the Brownian motion and initial
state ensures that conditioning on these reset times leaves the Gaussian
laws above unchanged. Weighting the no-reset law by $S(0,t)$ and
integrating the last-reset law against $\lambda(r)S(r,t)\,dr$ proves
\eqref{eq:general-last-reset-law}. The weights sum to one because
$S(0,t)+\int_0^t\lambda(r)S(r,t)\,dr=1$.
Finally, each coordinate depends only on its own initial state and drivers,
which proves conditional coordinate independence.
\end{proof}

Since $(t-r)/(1-r)\le t$ for $0\le r<t<1$, the Gaussian mixture above gives
\[
 \E\!\left[\|X_{i,t}^y-e(y_i)\|^2\right]
 \le (1-t)^2\bigl(\|e(y_i)\|^2+d\sigma_s^2\bigr)
       +d\sigma_d^2t(1-t)\longrightarrow0.
\]
Summing over positions and averaging over $Y$ yields
$p_t\Rightarrow\mathcal L(E(Y))$ as $t\uparrow1$.
For a bridge clock $\kappa$, replace $t$ by $\kappa(t)$ in this bound.

\subsection{Closed form for a shared bridge and reset schedule}
\label{app:shared-schedule-law}

The closed form extends beyond the linear clock: it applies whenever the
bridge clock and the first-reset CDF are both $\kappa(t)$. For the baseline
reset schedule, we use it for the repeated-reset training law and for
likelihood completion in RJD-BC-D, with $\kappa(t)=t^2$.

\begin{proposition}[Conditional law and likelihood]
\label{prop:shared-schedule-law}
Let $\kappa$ satisfy the clock conditions in Section~\ref{sec:dfm-background}.
Consider the repeated-reset bridge under the independence and noise
assumptions of Lemma~\ref{lem:conditional-state-law}, with drift factor
$\kappa'(t)/(1-\kappa(t))$, diffusion coefficient
$\sigma_d\sqrt{\kappa'(t)}$, and reset rate
$\lambda(t)=\kappa'(t)/(1-\kappa(t))$.
For $y_i=a$ and $u=\kappa(t)\in(0,1)$, put
$v_u=\sigma_d^2u(1-u)$ and $v_{0,u}=(1-u)^2\sigma_s^2+v_u$.
Then
\begin{equation}
 p_{i,t}(dx\mid a)
 =(1-u)\mathcal N(dx;u e(a),v_{0,u}I_d)
   +u\int_0^1\mathcal N(dx;e(a),v_u z I_d)\,dz.
 \label{eq:local-law}
\end{equation}
Its moments satisfy
\begin{align}
 \E[X_{i,t}\mid Y_i=a]&=u(2-u)e(a),\nonumber\\
 \E[\|X_{i,t}-e(a)\|^2\mid Y_i=a]
 &=(1-u)^3(\|e(a)\|^2+d\sigma_s^2)
       +d\sigma_d^2u(1-u)(1-u/2).
 \label{eq:exact-moment}
\end{align}
For even $d\ge4$ and $x\ne e(a)$, let $n=d/2-1$ and
$\beta=\|x-e(a)\|^2/(2v_u)>0$. Conditional on at least one reset, the density is
\begin{align}
 \int_0^1\mathcal N(x;e(a),v_u z I_d)\,dz
 &=(2\pi v_u)^{-d/2}\beta^{-n}\Gamma(n,\beta),\nonumber\\
 \Gamma(n,\beta)&=(n-1)!e^{-\beta}
                    \sum_{k=0}^{n-1}\frac{\beta^k}{k!}.
 \label{eq:analytic-density}
\end{align}
Here $\Gamma(n,\beta)=\int_\beta^\infty w^{n-1}e^{-w}\,dw$ is the upper
incomplete gamma function.
\end{proposition}

\begin{proof}
The process is the deterministic time change $u=\kappa(t)$ of the
linear-clock bridge with reset rate $(1-u)^{-1}$.
Lemma~\ref{lem:conditional-state-law} therefore gives no-reset weight
$1-u$ and last-reset density $(1-u)/(1-\tau)^2$, for $0<\tau<u$.
Following a last reset at $\tau$, the mean is $e(a)$ and the variance is
$\sigma_d^2(u-\tau)(1-u)/(1-\tau)$.
The substitution $z=(u-\tau)/(u(1-\tau))$ transforms this variance into
$v_u z$ and the last-reset measure into $u\,dz$ on $[0,1]$, with reversed
limits. Together with the no-reset Gaussian, this proves
\eqref{eq:local-law}.

The component means give
$(1-u)u e(a)+u e(a)=u(2-u)e(a)$.
The expected squared distance from $e(a)$ is
$(1-u)[(1-u)^2\|e(a)\|^2+d v_{0,u}]+u d v_u/2$, since
$\int_0^1z\,dz=1/2$. Expanding gives \eqref{eq:exact-moment}.
For the density, substituting $w=\beta/z$ yields
\[
 \int_0^1z^{-d/2}e^{-\beta/z}\,dz
 =\beta^{-n}\int_\beta^\infty w^{n-1}e^{-w}\,dw.
\]
Repeated integration by parts for the positive integer $n$ gives the
finite sum in \eqref{eq:analytic-density}.
\end{proof}

To sample \eqref{eq:local-law}, use the first Gaussian with probability
$1-u$; otherwise draw $z\sim\mathrm{Unif}(0,1)$ and
$\xi\sim\mathcal N(0,I_d)$ independently and set
$x=e(a)+\sqrt{v_u z}\,\xi$. In dimension 64, the density formula uses
$n=31$ and is evaluated in log space without reset-time quadrature.
For even $d\ge4$ and $0<u<1$, the density is singular at an anchor, but
the law has no atom there; a posterior version may be chosen on these
zero-measure points.
At $u=0$, including $\sigma_s=0$, the common base law is handled directly.
The mean squared distance in \eqref{eq:exact-moment} tends to zero at the
data endpoint.

\paragraph{Other jump schedules.}
The proposition covers $\lambda_\gamma(t)=\gamma t^{\gamma-1}/(1-t^\gamma)$
when the bridge also uses $\kappa(t)=t^\gamma$.
If the bridge remains at $\kappa(t)=t^2$ while $\gamma$ changes, the
repeated-reset process has no-reset weight $1-t^\gamma$ and last-reset density
$\gamma r^{\gamma-1}(1-t^\gamma)/(1-r^\gamma)^2$ on $(0,t)$.
For repeated resets, these weights must be combined with the Gaussian
laws in bridge time: the no-reset mean and variance are $\kappa(t)e(a)$
and $v_{0,\kappa(t)}$, and a last reset at physical time $r$ gives mean
$e(a)$ and variance
$\sigma_d^2(\kappa(t)-\kappa(r))(1-\kappa(t))/(1-\kappa(r))$.
This is the last-reset construction of Lemma~\ref{lem:conditional-state-law};
the uniform-mixture formula above does not generally apply.
Our non-absorbing schedule experiments also use $\gamma=1,4$ with the
bridge fixed at $\kappa(t)=t^2$. Training draws from this schedule-specific
last-reset mixture. For completed drift, the local likelihood is evaluated
by numerical quadrature; $\gamma=2$ retains the closed form above.
The quadrature uses 256 generalized Gauss--Laguerre nodes and checks the
completed posterior against 128 nodes at every sampled position, aborting
if the maximum $\ell_1$ difference exceeds $0.002$. This is a numerical
convergence check, not an exact likelihood evaluation. The raw LOO and
unmasked sampling arms do not evaluate the local likelihood.

For analytic completion with a learned predictor, we add
$\log\ell_{i,t}(x_i\mid a)$ to contextual logits and normalize over $a$.
The local likelihood is computed from the conditional training path using
the learned token embeddings, which also define reset destinations and
nearest-embedding decoding.

\section{Proofs and a counterexample}

\subsection{Minimizers of the training losses}
\label{app:posterior-learning}
\label{app:proof-loo-target}

All expectations below use the training pair $(Y,X_t^Y)$.
Logarithms are natural.

\paragraph{Full-posterior prediction.}
Recall the loss \eqref{eq:full-jump-loss},
\[
 \mathcal L_{\mathrm{full}}(\theta)
 =\E_{t,Y,X_t^Y}\left[-\sum_{i=1}^L
       \log p_{\theta,i,t}(Y_i\mid X_t^Y)\right].
\]
Condition on $(t,X_t^Y)=(t,x)$.
At position $i$, the conditional loss from predicting
$p_{\theta,i,t}(\cdot\mid x)$ instead of $p_{i,t}(\cdot\mid x)$ is
\[
 \KL\!\left(p_{i,t}(\cdot\mid x)
       \Vert p_{\theta,i,t}(\cdot\mid x)\right)\ge0.
\]
Averaging over training times and states and summing over positions
therefore shows that the unrestricted minimizers equal $p_{i,t}$ almost
everywhere. The learned drift is
\begin{equation}
 b_{\theta,i,t}(x)
 =\frac{\sum_{a\in\cV}p_{\theta,i,t}(a\mid x)e(a)-x_i}{1-t},
 \label{eq:learned-full-drift}
\end{equation}
and the reset probabilities are $p_{\theta,i,t}$. When $p_{\theta,i,t}=p_{i,t}$,
these coefficients and the corresponding reset clock recover the
posterior-averaged generator for the chosen path.

\paragraph{Leave-one-out prediction.}
Write $X_{-i,t}^Y$ for $X_t^Y$ with position $i$ omitted.
The loss \eqref{eq:loo-loss} is
\[
 \mathcal L_{\mathrm{LOO}}(\theta)
 =\E_{t,Y,X_t^Y}\left[-\sum_{i=1}^L
       \log q_{\theta,i,t}(Y_i\mid X_{-i,t}^Y)\right].
\]
Conditioning on $(t,X_{-i,t}^Y)$ gives the
conditional loss
\[
 \KL\!\left(q_{i,t}(\cdot\mid X_{-i,t}^Y)
       \Vert q_{\theta,i,t}(\cdot\mid X_{-i,t}^Y)\right).
\]
The same argument shows that the unrestricted minimizers equal
$q_{i,t}$ almost everywhere under the training law, establishing
\eqref{eq:loo-target}.

\subsection{Assumptions and proof of Proposition~\ref{prop:contextual-generator}}
\label{app:proof-contextual-generator}

\begin{proposition}\label{prop:contextual-generator-precise}
Consider the conditional path \eqref{eq:conditional-path} under the
assumptions of Appendix~\ref{app:conditional-sde}, with any nonnegative
deterministic reset rate $\lambda(t)$ satisfying
$\int_0^t\lambda(s)\,ds<\infty$ for every $t<1$. Let
$p_t(dx)=\sum_y p_{\mathrm{data}}(y)p_t(dx\mid y)$, and let $p_{i,t}$ and
$q_{i,t}$ be versions of $\mathbb P(Y_i\in\cdot\mid X_t^Y)$ and
$\mathbb P(Y_i\in\cdot\mid X_{-i,t}^Y)$ under this mixture.
On every $[0,T]$, $T<1$, assume that \eqref{eq:contextual-generator}
admits a Markov process started from $p_0$ whose laws satisfy its
integrated weak forward equation against smooth, compactly supported
test functions, and that this equation has a unique probability-law
solution from $p_0$.
Then this Markov process has law $p_t$ for every $t<1$, and
$p_t\Rightarrow\mathcal L(E(Y))$ as $t\uparrow1$.
\end{proposition}

\begin{proof}
Fix $T<1$. We establish two claims for every smooth, compactly supported
test function $f:\R^{dL}\to\R$.

\textbf{Claim 1.} The mixture $p_t$ satisfies the integrated weak forward
equation for the averaged operator $\mathcal A_t$:
\[
 \int f\,dp_t-\int f\,dp_0
 =\int_0^t\int\mathcal A_s f(x)\,p_s(dx)\,ds,
 \qquad 0\le t\le T.
\]

\emph{Proof of Claim 1.} For a fixed endpoint $y\in\cV^L$, the conditional
generator of \eqref{eq:conditional-path} is
\begin{align*}
 \mathcal A_s^y f(x)
 ={}&\sum_{i=1}^L\left[
   \frac{e(y_i)-x_i}{1-s}\cdot\nabla_i f(x)
   +\frac{\sigma_d^2}{2}\Delta_i f(x)\right]\\
 &+\lambda(s)\sum_{i=1}^L
   \left[f(x^{i\leftarrow e(y_i)})-f(x)\right].
\end{align*}
Here $\nabla_i$ and $\Delta_i$ are the gradient and Laplacian with
respect to the block $x_i\in\R^d$, and
$x^{i\leftarrow e(y_i)}$ replaces that block by $e(y_i)$.
The continuous terms are uniformly bounded over $s\le T$, $x$, and $y$:
the derivatives of $f$ have compact support, $1-s\ge1-T$, and the
finite vocabulary makes the embeddings bounded. The jump term is bounded
in absolute value by $2L\lambda(s)\|f\|_\infty$, where
$\|f\|_\infty=\sup_x|f(x)|$. These bounds and
$\int_0^T\lambda(s)\,ds<\infty$ justify taking expectations in the jump
It\^o formula and interchanging the time integral with expectation;
the Brownian and compensated jump integrals have zero expectation.
There are no jumps at any fixed deterministic time almost surely,
since the Poisson intensity measures are $\lambda(s)\,ds$, so the
left limits in that formula can be replaced by $X_s^y$ under expectation.

Now draw $Y\sim p_{\mathrm{data}}$ and write $\mathcal A_s^Y$ for the
conditional generator with endpoint $Y$. Since
$\mathbb P(Y_i=a\mid X_s^Y=x)=p_{i,s}(a\mid x)$, conditioning
$\mathcal A_s^Y f(X_s^Y)$ on $X_s^Y=x$ gives exactly
$\mathcal A_s f(x)$ in \eqref{eq:averaged-generator}.
Consequently, for $0\le t\le T$, taking expectation over $Y$ and its
conditional process gives
\begin{align*}
 \int f\,dp_t-\int f\,dp_0
 &=\int_0^t\E[\mathcal A_s^Y f(X_s^Y)]\,ds\\
 &=\int_0^t\int\mathcal A_s f(x)\,p_s(dx)\,ds.
\end{align*}
This proves Claim 1 and establishes posterior averaging directly, following the
conditional-expectation argument of
\citet[Proposition~1]{holderrieth2025generator}.

\textbf{Claim 2.} Replacing the full-posterior reset probabilities by
the LOO probabilities leaves the generator integral under $p_t$ unchanged:
\[
 \int\mathcal A_t f\,dp_t
 =\int\mathcal A_t^{\mathrm{ctx}}f\,dp_t,
 \qquad 0\le t\le T.
\]

\emph{Proof of Claim 2.} From \eqref{eq:averaged-generator} and
\eqref{eq:contextual-generator},
using $\sum_a p_{i,t}(a\mid x)=\sum_a q_{i,t}(a\mid x_{-i})=1$,
\[
 (\mathcal A_t-\mathcal A_t^{\mathrm{ctx}})f(x)
 =\lambda(t)\sum_{i,a}
   \bigl[p_{i,t}(a\mid x)-q_{i,t}(a\mid x_{-i})\bigr]
   f(x^{i\leftarrow e(a)}).
\]
The tower property gives
\begin{equation}
 \int p_{i,t}(a\mid x)\,p_t(dx_i\mid x_{-i})
 =q_{i,t}(a\mid x_{-i}).
 \label{eq:reset-tower}
\end{equation}
Since $f(x^{i\leftarrow e(a)})$ does not depend on $x_i$,
\begin{multline*}
 \int(\mathcal A_t-\mathcal A_t^{\mathrm{ctx}})f\,dp_t\\
 =\lambda(t)\sum_{i,a}\int f(x^{i\leftarrow e(a)})
   \left[\int p_{i,t}(a\mid x)\,p_t(dx_i\mid x_{-i})
         -q_{i,t}(a\mid x_{-i})\right]p_t(dx_{-i})=0,
\end{multline*}
which proves Claim 2.

Combining Claims 1 and 2, $p_t$ solves the integrated weak forward
equation for $\mathcal A_t^{\mathrm{ctx}}$. By the assumed uniqueness,
the Markov process started from $p_0$ therefore has law $p_t$ on $[0,T]$.
Since $T<1$ was arbitrary, this holds for every $t<1$.
Endpoint convergence follows from the bound in
Appendix~\ref{app:conditional-sde}.
\end{proof}

\subsection{A counterexample for LOO drift}
\label{app:example-proof}

\begin{example}[LOO drift with informative context]
\label{ex:loo-drift-failure}
Take $L=2$, $d=1$, and $Y=(B,B)$, where $B$ is uniform on $\{-1,1\}$;
let $e(a)=a$. Conditional on $B$, let the coordinates follow independent
copies of the repeated-reset bridge \eqref{eq:conditional-path}, with
$X_{1,0}=X_{2,0}=0$, $\sigma_d=1$, and $\lambda(t)=(1-t)^{-1}$.
Both positions therefore carry noisy information about the same endpoint.
Define, under this prescribed conditional-path mixture,
\[
 m_t(x)=\E[B\mid X_{2,t}=x],\qquad
 M_t(x_1,x_2)=\E[B\mid X_{1,t}=x_1,X_{2,t}=x_2].
\]
By symmetry, $m_t$ is also the posterior mean given the first coordinate
alone. The drift $b_{i,t}$ at position $i$ is
$(M_t(x_1,x_2)-x_i)/(1-t)$; its LOO replacement is
$(m_t(x_{3-i})-x_i)/(1-t)$.
Let $\widetilde{\mathcal A}_t$ change only these drifts in
\eqref{eq:averaged-generator}, retaining the exact jump
probabilities $p_{i,t}$, reset clocks, and diffusion coefficients.
The original process with generator $\mathcal A_t$ has marginal law $p_t$.
Write $\widetilde p_t$ for the marginal law of the process with generator
$\widetilde{\mathcal A}_t$, started from the same initial law
$\widetilde p_0=p_0=\delta_{(0,0)}$.

\textbf{Claim.} Replacing the drift in this way does not preserve the
prescribed marginal path $(p_t)_{0\le t<1}$, even though the jump rate
kernel remains exact.
\end{example}

\begin{figure}[t]
 \centering
 \includegraphics[width=.76\linewidth]{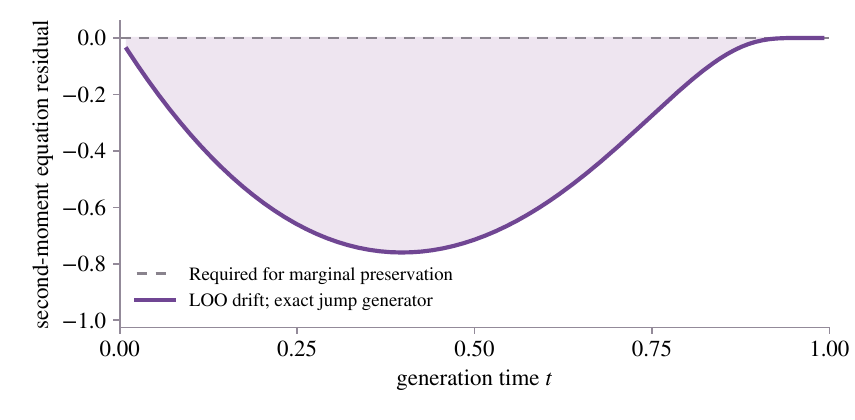}
 \caption{\textbf{LOO drift violates the moment equation with informative context.}
 The residual in \eqref{eq:counterexample-moments} for
 Example~\ref{ex:loo-drift-failure}, with $Y=(B,B)$ and $\sigma_d=1$.
 Preserving the prescribed marginal path requires a zero residual;
 replacing only the drift posterior by LOO gives a negative residual.
 The expectation is evaluated by deterministic quadrature of the exact
 local conditional law, not by trajectory simulation.}
 \label{fig:loo-moment-failure}
\end{figure}

\begin{proof}[Proof]
Take $f(x)=x_1^2$. All second moments are finite on compact time intervals.
Each generator determines the second-moment derivative under its own
marginal law:
\begin{align*}
 \frac{d}{dt}\E_{p_t}[X_1^2]
 &=\int\mathcal A_t f\,dp_t,\\
 \frac{d}{dt}\E_{\widetilde p_t}[X_1^2]
 &=\int\widetilde{\mathcal A}_t f\,d\widetilde p_t.
\end{align*}
Suppose, for contradiction, that $\widetilde p_t=p_t$ for every $t<1$.
Then these second moments, and hence their derivatives, must agree.

To compare these derivatives, we first compute the generator difference.
The continuous diffusion and jump terms of the two generators coincide.
For $f(x)=x_1^2$, we have $\partial_1f=2x_1$ and $\partial_2f=0$,
so only the difference between the first-coordinate drifts contributes:
\[
 (\widetilde{\mathcal A}_t-\mathcal A_t)f(x)
 =\frac{2x_1}{1-t}\bigl(m_t(x_2)-M_t(x_1,x_2)\bigr).
\]
Integrating against $p_t$ gives
\[
 \int(\widetilde{\mathcal A}_t-\mathcal A_t)f\,dp_t
 =\frac{2}{1-t}\left(\E[X_{1,t}m_t(X_{2,t})]
       -\E[X_{1,t}M_t(X_{1,t},X_{2,t})]\right).
\]
It remains to evaluate the two expectations on the right. Both are under
the prescribed joint law of $(B,X_t)$, whose state marginal is $p_t$.
With $\kappa(t)=t$, \eqref{eq:exact-moment} gives
$\E[X_{1,t}\mid B]=t(2-t)B$.

For the first expectation, involving the LOO term, condition on both
$B$ and $X_{2,t}$.
The coordinates are independent conditional on $B$, so observing
$X_{2,t}$ provides no further information about $X_{1,t}$ once $B$ is
known. Thus
\[
 \E[X_{1,t}\mid B,X_{2,t}]
 =\E[X_{1,t}\mid B]=t(2-t)B.
\]
Since $m_t(X_{2,t})$ is determined by $X_{2,t}$, it can be taken outside
this conditional expectation. The tower property therefore gives
\begin{align*}
 \E[X_{1,t}m_t(X_{2,t})]
 &=\E\!\left[\E[X_{1,t}m_t(X_{2,t})\mid B,X_{2,t}]\right]\\
 &=\E\!\left[m_t(X_{2,t})\E[X_{1,t}\mid B,X_{2,t}]\right]\\
 &=t(2-t)\E[Bm_t(X_{2,t})].
\end{align*}
Next condition on $X_{2,t}$ alone and use the definition
$m_t(X_{2,t})=\E[B\mid X_{2,t}]$:
\begin{align*}
 \E[Bm_t(X_{2,t})]
 &=\E\!\left[\E[Bm_t(X_{2,t})\mid X_{2,t}]\right]\\
 &=\E\!\left[m_t(X_{2,t})\E[B\mid X_{2,t}]\right]\\
 &=\E[m_t(X_{2,t})^2].
\end{align*}
Combining these two calculations yields
\[
 \E[X_{1,t}m_t(X_{2,t})]
 =t(2-t)\E[m_t(X_{2,t})^2].
\]

For the second expectation, involving the full-posterior term, recall that
$M_t(X_{1,t},X_{2,t})=\E[B\mid X_{1,t},X_{2,t}]$.
Here $X_{1,t}$ is known when conditioning on both coordinates, so
\begin{align*}
 \E[X_{1,t}M_t(X_{1,t},X_{2,t})]
 &=\E\!\left[X_{1,t}\E[B\mid X_{1,t},X_{2,t}]\right]\\
 &=\E\!\left[\E[X_{1,t}B\mid X_{1,t},X_{2,t}]\right]\\
 &=\E[X_{1,t}B].
\end{align*}
Finally, conditioning on $B$ and using $B^2=1$ gives
\begin{align*}
 \E[X_{1,t}B]
 &=\E\!\left[B\E[X_{1,t}\mid B]\right]\\
 &=t(2-t)\E[B^2]=t(2-t).
\end{align*}

Substituting these two expectations into the generator difference gives
\begin{align}
 \int(\widetilde{\mathcal A}_t-\mathcal A_t)f\,dp_t
 &=\frac{2t(2-t)}{1-t}\left(\E[m_t(X_{2,t})^2]-1\right)\nonumber\\
 &=-\frac{2t(2-t)}{1-t}
       \E\!\left[1-m_t(X_{2,t})^2\right]<0,
 \qquad 0<t<1.
 \label{eq:counterexample-moments}
\end{align}
To justify the strict inequality in \eqref{eq:counterexample-moments},
we must show that $\E[1-m_t(X_{2,t})^2]>0$.
For either endpoint $B=\pm1$, the local law of $X_{2,t}$ contains the
no-reset component $\mathcal N(\pm t,t(1-t))$ with weight $1-t>0$.
For $0<t<1$, both Gaussian densities are positive everywhere. Hence both
values of $B$ retain positive conditional probability after observing
$X_{2,t}$, so $|m_t(X_{2,t})|<1$ almost surely. Since $B^2=1$, this gives
\[
 1-m_t(X_{2,t})^2
 =\operatorname{Var}(B\mid X_{2,t})>0
 \quad\text{almost surely}.
\]
Its expectation is therefore strictly positive, while the prefactor
$-2t(2-t)/(1-t)$ is strictly negative. This proves the strict negativity
of the residual.
Figure~\ref{fig:loo-moment-failure} plots this residual as a function of time.

Under the assumed equality $\widetilde p_t=p_t$, this yields
\[
 \frac{d}{dt}\left(\E_{\widetilde p_t}[X_1^2]-\E_{p_t}[X_1^2]\right)
 =\int(\widetilde{\mathcal A}_t-\mathcal A_t)f\,dp_t<0,
 \qquad 0<t<1,
\]
contradicting equality of the derivatives. Thus the two marginal paths
cannot coincide.
\end{proof}

\subsection{Proof of Proposition~\ref{prop:completion}}
\label{app:proof-completion}

\begin{proof}
Fix $i$ and $t$. Given $Y$, the coordinates are independent, and the law
of $X_{i,t}$ depends on $Y$ only through $Y_i$. Averaging over the other
endpoint tokens therefore gives $X_{i,t}\perp X_{-i,t}\mid Y_i$.
Consequently, for almost every context $x_{-i}$,
\begin{align*}
 \mathbb P(Y_i=a,X_{i,t}\in dx_i\mid X_{-i,t}=x_{-i})
 &=q_{i,t}(a\mid x_{-i})p_{i,t}(dx_i\mid a)\\
 &=q_{i,t}(a\mid x_{-i})
   \ell_{i,t}(x_i\mid a)\mu_{i,t}(dx_i).
\end{align*}
Summing over $a$ shows that the conditional density of $X_{i,t}$ given
$X_{-i,t}=x_{-i}$, with respect to $\mu_{i,t}$, is
\[
 \sum_b q_{i,t}(b\mid x_{-i})\ell_{i,t}(x_i\mid b).
\]
Bayes' rule divides the joint density by this sum, yielding
\eqref{eq:completion} for $p_t$-almost every $x$ where the denominator is
positive.
\end{proof}

\FloatBarrier

\section{\textcolor{black}{Architecture, training, and sampling}}
\label{app:sampler}

Algorithms~\ref{alg:rjd-training} and~\ref{alg:rjd-sampling} summarize
training and generation for the repeated-reset model with bridge clock
$\kappa$ and deterministic physical-time reset rate $\lambda$.
The numerical sampler freezes predictions on each grid interval and uses
the no-reset probability and last-reset distribution described below.
Sampling stops before $t=1$, where the chosen reset schedules have infinite
cumulative intensity.

\subsection{Input-isolated block details}
\label{app:input-isolated-details}

The positionwise update in \eqref{eq:architecture} is
\[
 F_\ell(h,a,t)=h+a+
 \operatorname{FF}_\ell\!\left(\operatorname{LN}_\ell(h+a),t\right).
\]
Here $\operatorname{LN}_\ell$ is positionwise layer normalization and
$\operatorname{FF}_\ell$ is a positionwise feed-forward network.
Time conditioning and residual gates are included in
$\operatorname{Attn}_\ell$ and $\operatorname{FF}_\ell$;
attention also includes positionwise normalization of its queries
and sources.

\subsection{A separate head for predicting $p_{i,t}$}
\label{app:separate-drift-head}

An alternative to analytic likelihood completion is to learn the drift
posterior. Let $h_i^{(D)}(x_{-i},t)$ denote the contextual feature defined by
the input-isolated architecture in \eqref{eq:architecture}. A positionwise
head receives this feature together with the local observation,
\begin{align}
 r_{\theta,i}(\cdot\mid x,t)
 &=\softmax\bigl(g_\theta(h_i^{(D)}(x_{-i},t),x_i,t)\bigr),\nonumber\\
 \mathcal L_{\mathrm{dual}}
 &=\mathcal L_{\mathrm{LOO}}
   +\E\left[-\sum_i\log r_{\theta,i}(Y_i\mid X_t,t)\right].
 \label{eq:dual-loss}
\end{align}
For a fixed conditional path, including fixed embeddings and corruption
schedules, assume that the model class jointly represents both posteriors.
Every global minimizer of the expected loss then satisfies
$q_{\theta,i,t}=q_{i,t}$ and $r_{\theta,i}=p_{i,t}$ almost everywhere
under the training law of $(t,X_t^Y)$, recovering the coefficients of
\eqref{eq:contextual-generator} under that law. This statement does not
constrain predictions outside the support of the training law, in
particular beyond the training-time cutoff. Marginal preservation over a
given time interval still requires exact posterior coefficients on that
interval and the existence and uniqueness assumptions of
Proposition~\ref{prop:contextual-generator-precise}.
The local input and the drift-head activations never enter the jump branch.
Both losses may update the shared contextual parameters.

\subsection{Simulation-free training}
\label{app:training-algorithm}

Algorithm~\ref{alg:rjd-training} samples training states directly from the
conditional law in Lemma~\ref{lem:conditional-state-law}, using bridge time
$\kappa(t)$. Conditional on $Y$ and $t$, reset times and Gaussian draws
are independent across positions; examples are sampled independently.
The atom $\tau_i=0$ represents no reset; the
predictor observes only $x$ and $t$, without reset times or indicators.
Gradients pass through the sampled states to the embedding parameters.

\begin{algorithm}[htbp]
\caption{Simulation-free training of the RJD LOO predictor}
\label{alg:rjd-training}
\begin{algorithmic}[1]
\Require Data law $p_{\mathrm{data}}$; trainable predictor $q_{\theta,i,t}$ and embedding $e$;
\Statex \hspace{\algorithmicindent} bridge clock $\kappa$, reset rate $\lambda$, noise scales $\sigma_s,\sigma_d$;
\Statex \hspace{\algorithmicindent} batch size $B$ and training cutoff $0<t_{\max}<1$.
\While{training budget remains}
  \For{$b=1,\ldots,B$}
    \State Draw $Y\sim p_{\mathrm{data}}$, $t\sim\operatorname{Unif}(0,t_{\max})$; set $u\gets\kappa(t)$.
    \For{$i=1,\ldots,L$}
      \State Draw $\tau_i\in[0,t]$ with $\mathbb P(\tau_i\le h)=S(h,t)$ for $0\le h\le t$.
      \State Draw $\xi_i\sim\mathcal N(0,I_d)$ independently.
      \If{$\tau_i=0$}
        \State $x_i\gets u e(Y_i)+\sqrt{(1-u)^2\sigma_s^2+\sigma_d^2u(1-u)}\,\xi_i$.
      \Else
        \State $x_i\gets e(Y_i)+\sigma_d\sqrt{\dfrac{(u-\kappa(\tau_i))(1-u)}{1-\kappa(\tau_i)}}\,\xi_i$.
      \EndIf
    \EndFor
    \State One forward pass gives $q_{\theta,i,t}(\cdot\mid x_{-i})$ for all $i$.
    \State $\mathcal L_b\gets-\dfrac1L\sum_{i=1}^L\log q_{\theta,i,t}(Y_i\mid x_{-i})$.
  \EndFor
  \State Update $\theta$ and the embedding parameters using $B^{-1}\sum_{b=1}^B\mathcal L_b$.
\EndWhile
\end{algorithmic}
\end{algorithm}

The Text8 experiments use $t_{\max}=0.999$. For
$\lambda(t)=\kappa'(t)/(1-\kappa(t))$, the equivalent mixture
\eqref{eq:local-law} can also be sampled directly. The unrestricted
control uses the same state sampler and replaces $q_{\theta,i,t}$ by
$p_{\theta,i,t}$ in the loss. Pure diffusion uses $\lambda=0$ in both
training and sampling.

\subsection{Frozen-coefficient interval update}

For the linear bridge clock $\kappa(t)=t$, consider an interval
$[a,b]\subset[0,1)$ and freeze the contextual
reset distribution $q_i$ and drift endpoint mean $m_i$ at its left endpoint.
With $S(a,b)=\exp(-\int_a^b\lambda(r)\,dr)$, the probability of no reset
is $S(a,b)$, and the unconditional density of the last reset time
$\tau\in(a,b)$ is $\lambda(\tau)S(\tau,b)$. Its destination is sampled from $q_i$,
independently of the clock. Earlier resets can be integrated out because
the last reset replaces the state completely.

Set $\tau=a$ and $z_i=X_{i,a}$ if no reset occurs; otherwise let $\tau$ be the
last reset time and $z_i$ its destination embedding. The exact update for
this frozen-coefficient process is
\begin{equation}
 X_{i,b}=m_i+\alpha(z_i-m_i)
          +\sigma_d\sqrt{(b-\tau)\alpha}\,\xi_i,
 \qquad \alpha=\frac{1-b}{1-\tau},\quad
 \xi_i\sim\mathcal N(0,I_d).
 \label{eq:interval-update}
\end{equation}
For a general bridge clock $\kappa$, keep $a,b,\tau$ in physical time
when computing $S$ and sampling resets, and apply \eqref{eq:interval-update}
with $b,\tau$ replaced by $\kappa(b),\kappa(\tau)$ in $\alpha$ and the
Gaussian variance. This is the construction used in
Algorithm~\ref{alg:rjd-sampling} and does not require $\kappa$ to be invertible.
Alternatively, to sample resets directly in bridge time $u=\kappa(t)$,
suppose $\kappa$ is strictly increasing with $\kappa'(t)>0$ for $0<t<1$.
The reset rate per unit bridge time is then
\begin{equation}
 \lambda_{\mathrm{bridge}}(u)
 =\frac{\lambda(\kappa^{-1}(u))}{\kappa'(\kappa^{-1}(u))},
 \qquad 0<u<1,
 \label{eq:reset-rate-clock-conversion}
\end{equation}
where $\lambda$ remains the physical-time rate. In particular,
$\int_{\kappa(a)}^{\kappa(b)}\lambda_{\mathrm{bridge}}(u)\,du
=\int_a^b\lambda(t)\,dt$, so both clocks give the same no-reset probability.
For $\kappa(t)=t^2$, the baseline physical rate $2t/(1-t^2)$ becomes
$\lambda_{\mathrm{bridge}}(u)=1/(1-u)$, whereas the physical rate
$1/(1-t)$ becomes $1/[2\sqrt{u}(1-\sqrt{u})]$.

The post-reset drift uses $m_i$, not the last sampled embedding as an
additional hidden endpoint. Predictions are recomputed at the next grid
point. This uses one contextual backbone call per interval and no terminal
prediction call. Freezing state-dependent predictions, truncating before
the endpoint, and decoding at the cutoff are numerical approximations.

\subsection{Repeated-reset generation}
\label{app:generation-algorithm}

Algorithm~\ref{alg:rjd-sampling} uses physical reset times and applies
\eqref{eq:interval-update} in bridge time. The local likelihood
$\ell_{i,t}$ is defined in Proposition~\ref{prop:completion} and must match
the training path and embedding. Appendices~\ref{app:conditional-sde}
and~\ref{app:shared-schedule-law} give its Gaussian-mixture form and
evaluation. All predictions are computed from
the state at the interval's left endpoint and held fixed throughout that
interval. Reset times, destination draws, and Gaussian noises are
independent across positions conditional on those predictions.

\begin{algorithm}[H]
\caption{RJD generation with LOO resets and completed or raw LOO drift}
\label{alg:rjd-sampling}
\begin{algorithmic}[1]
\Require Trained $q_{\theta,i,t}$ and $e$; $\kappa,\lambda,\sigma_s,\sigma_d$;
\Statex \hspace{\algorithmicindent} grid $0=t_0<\cdots<t_M<1$; drift choice; local likelihood $\ell_{i,t}$ if completed.
\State Draw $x_i\sim\mathcal N(0,\sigma_s^2I_d)$ independently for $i=1,\ldots,L$.
\For{$j=0,\ldots,M-1$}
  \State Set $s\gets t_j$, $r\gets t_{j+1}$.
  \State One forward pass gives $q_i(a)\gets q_{\theta,i,s}(a\mid x_{-i})$ for all $i,a$.
  \If{using completed drift (RJD-BC-D)}
    \State $m_i\gets\dfrac{\sum_{a\in\cV}q_i(a)\ell_{i,s}(x_i\mid a)e(a)}{\sum_{a\in\cV}q_i(a)\ell_{i,s}(x_i\mid a)}$ for all $i$.
  \Else \Comment{raw LOO drift (RJD-LOO)}
    \State $m_i\gets\sum_{a\in\cV}q_i(a)e(a)$ for all $i$.
  \EndIf
  \For{$i=1,\ldots,L$}
    \State Draw $\tau_i\in[s,r]$ with $\mathbb P(\tau_i\le h)=S(h,r)$ for $s\le h\le r$.
    \If{$\tau_i=s$} \Comment{no reset}
      \State $z_i\gets x_i$.
    \Else
      \State Draw $A_i\sim q_i$; set $z_i\gets e(A_i)$.
    \EndIf
    \State Set $\alpha_i\gets\dfrac{1-\kappa(r)}{1-\kappa(\tau_i)}$; draw $\xi_i\sim\mathcal N(0,I_d)$ independently.
    \State $x_i'\gets m_i+\alpha_i(z_i-m_i)+\sigma_d\sqrt{(\kappa(r)-\kappa(\tau_i))\alpha_i}\,\xi_i$.
  \EndFor
  \State $x\gets x'$.
\EndFor
\State Decode $\widehat Y_i\in\arg\min_{a\in\cV}\|x_i-e(a)\|^2$ for each $i$; \Return $\widehat Y$.
\end{algorithmic}
\end{algorithm}

The law of $\tau_i$ has an atom of mass $S(s,r)$ at $s$ and
unconditional density $\lambda(h)S(h,r)$ on $(s,r)$. Thus each interval includes any number of
resets through its last-reset law, and every coordinate remains eligible
for later resets. The algorithm uses $M$ network evaluations and no
terminal prediction call. Raw LOO drift is the empirical variant and is
not covered by Proposition~\ref{prop:contextual-generator}.

\FloatBarrier

\section{Baseline paths and information access}
\label{app:baselines}

\begingroup
\color{black}
For Text8, masked discrete flow matching (DFM) \citep{gat2024dfm} uses
$\mathsf m\to A_i$ and embedded pure jumps (EPJ) use $0\to e(A_i)$.
Both use $q_{\theta,i,t}$ and the same backbone after the input encoder;
each position jumps once and remains fixed, with reveal probability
$\kappa(t)=t^2$.
\endgroup

The pure-jump reference starts with every coordinate unresolved, commits
to $e(Y_i)$ at rate $\kappa'(t)/(1-\kappa(t))$, and remains fixed after commitment.
The baseline experiments use $\kappa(t)=t^2$. Its augmented state records
commitment. At an unresolved position the local
state carries no token information, so $p_{i,t}$ and $q_{i,t}$
agree there in the idealized model. Their learned architectures
and losses can nevertheless behave differently.

Pure diffusion follows \eqref{eq:conditional-path} with $\lambda=0$.
Its conditional law is
$\mathcal N(t e(Y_i),[(1-t)^2\sigma_s^2+\sigma_d^2t(1-t)]I_d)$.
A predictor of $p_{i,t}$ or exact likelihood-completed LOO supplies its drift
mean; raw LOO is an empirical ablation without that justification.
Absorbing jump diffusion follows this bridge until its first reset, then
holds both the anchor and an observed commitment indicator fixed. Completion
for that control includes the observed status. Proposition~\ref{prop:contextual-generator}
concerns repeated resets without such an indicator and is not its proof.

The numerical experiments use the time change $\kappa(t)=t^2$ for the
continuous bridge.

\section{Four-token parity diagnostic}
\label{app:parity-learning}

For Table~\ref{tab:local-shortcut}, we use $\cV=\{0,1\}$, $L=4$,
scalar embeddings $e(a)=2a-1$, and $p_{\mathrm{data}}$ uniform on
$\{y\in\cV^4:\sum_i y_i\equiv0\pmod 2\}$.
Each token is marginally uniform, but the other three determine it
exactly. A generated string is invalid if it has odd parity;
independently sampling the four marginal distributions gives
$50\%$ invalid strings.
We use $\lambda(t)=(1-t)^{-1}$ and train equally sized MLPs for
250 updates. Table~\ref{tab:local-shortcut} reports cross-entropy
and the percentage of invalid strings, averaged over three seeds
with 8,192 generated strings per seed.

\begin{figure}[!htb]
 \color{black}
 \centering
 \includegraphics[width=\linewidth]{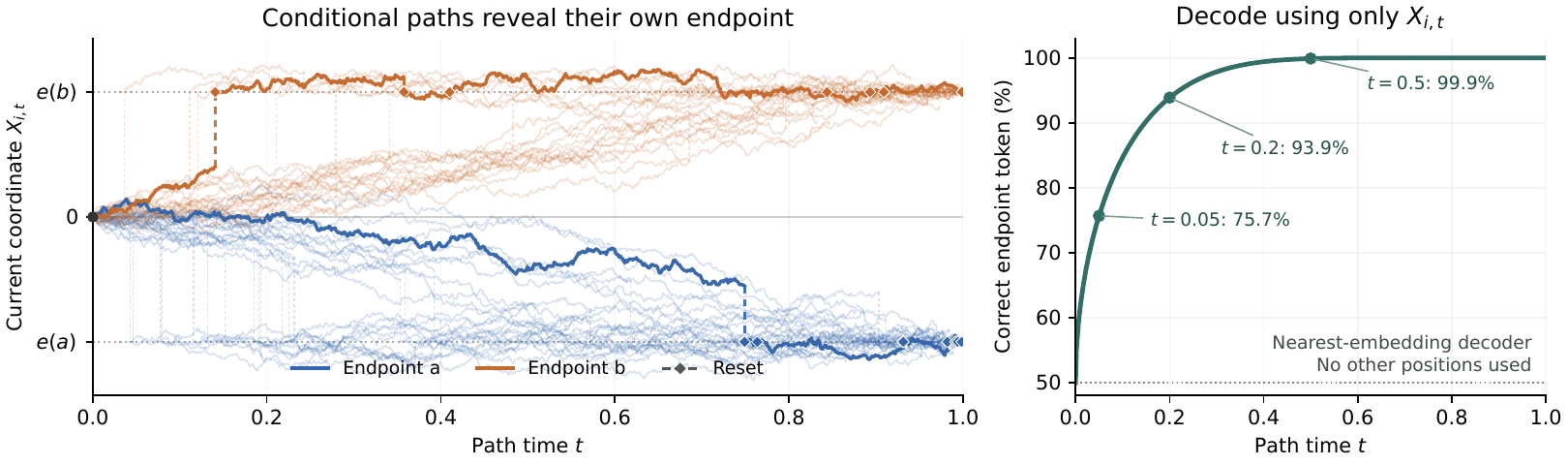}
 \caption{\textbf{Local information in conditional paths.}
 Repeated-reset bridges with $e(a)=-1$, $e(b)=1$, $X_0=0$,
 $\sigma_d=0.35$, and $\lambda(t)=(1-t)^{-1}$.
 Left: 48 coordinate paths, colored by endpoint; dashed segments
 indicate jumps, after which diffusion continues.
 Right: endpoint recovery using only the current local state.}
 \label{fig:local-information}
\end{figure}
\FloatBarrier

\section{Text8 experimental details and additional results}
\label{app:numerical-details}

\subsection{Training and sampling protocol}

We use 90 million training characters and a nominal five-million-character
validation split of Text8 \citep{mahoney2006text8}, divided into length-128 sequences with incomplete
chunks discarded. The alphabet consists of the 26 lower-case letters and
space. Each model is trained for 60,000 updates with batch size 512 and
uniform training times on $[0,.999]$. The endpoint cross-entropy includes
all positions. AdamW uses learning rate $10^{-3}$, no warmup or weight decay,
coefficients $(0.9,0.999)$, numerical constant $10^{-8}$, and gradient-norm
clipping at one. Evaluation uses the network and embedding EMA with decay
$0.9999$. Training and evaluation use FP32 and the math attention backend.

The shared backbone has eight blocks of width 256, eight attention heads,
32 time features, AdaLN-Zero conditioning \citep{peebles2023dit},
SwiGLU width 704 \citep{shazeer2020glu}, and dropout
$0.1$. The LOO and unrestricted variants have the same 6,878,880 trainable
parameters, including the normalized $27\times64$ embedding matrix. The
LOO construction excludes the local input through the fixed-source
keys and values in \eqref{eq:architecture}; it is not
obtained by masking only the diagonal of ordinary multilayer self-attention.
DFM replaces the geometric input encoder with a token-plus-mask lookup and
has 6,867,616 parameters, approximately $0.164\%$ fewer. Its backbone after
the input encoder and its token-level conditional path match EPJ.

The continuous bridge uses the fixed schedule $\kappa(t)=t^2$. During
continuous motion, the conditional drift in physical time is
$2t(e(y_i)-x_i)/(1-t^2)$ and the diffusion coefficient is
$\sigma_d\sqrt{2t}$. The jump-schedule comparisons keep this bridge fixed
and vary only the physical-time reset rate
$\lambda_\gamma(t)=\gamma t^{\gamma-1}/(1-t^\gamma)$,
$\gamma\in\{1,2,4\}$.

Unless specified otherwise, we use the cosine grid
\[
 t_j=.999\left[1-\cos\left(\frac{\pi j}{2M}\right)\right],
 \qquad j=0,\ldots,M,
\]
where $M$ is the number of predictor intervals. NFE counts actual backbone
evaluations, including terminal predictions for unabsorbed coordinates.
Pure diffusion and RJD are decoded to the nearest token embedding at the
cutoff; absorbing jump models use the stated token rule for remaining
unabsorbed positions.
The conditional-path comparisons use temperature one and CAT for jumps;
the separately labelled TOPP variant changes only reset
destinations.
AJD-BC-D and AJD-BC reuse the AJD-LOO checkpoints; they change posterior
routing without additional training or backbone evaluations. Their local
likelihood is Gaussian at an unabsorbed coordinate and includes the token
atom and observed status after absorption. PD-BC similarly reuses PD-LOO.

\paragraph{Repeated resets.}
\begingroup
\color{black}
RJD uses \eqref{eq:conditional-path} with $(\sigma_s,\sigma_d)=(0,1)$
and independent reset clocks of rate $1/(1-u)$ in path time
$u=\kappa(t)=t^2$, or physical-time intensity
$\kappa'(t)/(1-\kappa(t))=2t/(1-t^2)$.
\endgroup Training samples follow
\eqref{eq:local-law} with $u=\kappa(t)$; no absorption indicator is
observed. RJD-LOO and RJD-BC-D reuse the same repeated-reset LOO checkpoints.
For RJD-BC-D, the local likelihood is evaluated using
\eqref{eq:analytic-density}, without reset-time quadrature. Sampling freezes
the predictions on each grid interval and uses the last-reset update
\eqref{eq:interval-update} between $\kappa(t_j)$ and $\kappa(t_{j+1})$.
The last-reset law integrates out any earlier resets within that interval;
diffusion continues afterwards. Each interval costs one network evaluation,
and nearest-embedding decoding costs none. RJD-BC-D has the prescribed
marginals with exact posteriors and dynamics; learned prediction
and finite-step sampling introduce approximations. RJD-LOO additionally
replaces the drift posterior by $q_{\theta,i,t}$ and is an empirical ablation.
RJD and AJD use path-specific training, different observed states, and
different terminal rules, so their comparison does not isolate the effect
of absorption alone.

\paragraph{Main evaluation.}
\begingroup
\color{black}
The eight settings in Table~\ref{tab:text8-paths} and RJD-LOO TOPP in
Table~\ref{tab:text8-calibration} use 4,096 sequences per training seed
at 128 NFE, generated in batches of 64 on A100 GPUs with TF32 disabled.
Both tables round $D_5$ and $E_5$ to integers.
\endgroup Batch $b\in\{0,\ldots,63\}$ for training seed $s$ uses
generation seed $625000+1000s+b$. RJD and PD use 128 predictor intervals;
DFM and EPJ use 127 intervals and one terminal prediction. Actual backbone
calls are checked in every batch. The checkpoints are unchanged across
sampling variants; fresh generation adds no training replicates.

The main comparison uses validation characters
$[3{,}000{,}000:4{,}000{,}000]$ as the 4-gram reference. The separate
1,024-sample ablations use the first million validation characters unless
stated otherwise. The fixed word dictionary is derived from the test
split and has been reused during development; this is not an untouched
test evaluation. Dictionary membership is a lexical metric, not a measure
of sentence grammaticality. We compute each metric separately for each seed,
then report its arithmetic mean and sample standard deviation across seeds
0, 1, and 2. Repeated evaluations of one checkpoint are not additional
training seeds. The path, decoding, and clock comparisons have separate
evaluation records, which accounts for small differences between their
CAT results.

\subsection{\textcolor{black}{Evaluation metrics}}
\label{app:metrics}
\paragraph{Word validity and diversity.}
Words are sequences of lowercase letters a--z, separated by spaces or other
non-letter characters. Word fragments at sample boundaries are retained.
For each seed, let $\mathcal W_\ell$ be the multiset of generated
words of length at least $\ell\in\{5,6\}$, and $\mathcal D$ the fixed
dictionary from the Text8 test split. Define
\begin{align}
 V_\ell
 &=\frac{\sum_{w\in\mathcal W_\ell}\mathbf 1\{w\in\mathcal D\}}
         {|\mathcal W_\ell|},
 \label{eq:word-metric}\\
 D_\ell&=\bigl|\{w:w\in\mathcal W_\ell,\ w\in\mathcal D\}\bigr|.
 \nonumber
\end{align}
Compare $D_\ell$ only at matched sample counts and sequence lengths.

\paragraph{Effective valid-word count.}
If $f_\ell(w)$ is the relative
frequency of $w$ among valid words in $\mathcal W_\ell$, define
\[
 E_\ell=\exp\!\left(-\sum_{w\in\mathcal D:\,f_\ell(w)>0}
 f_\ell(w)\log f_\ell(w)\right),
\]
setting $E_\ell=0$ when no valid words occur.

\paragraph{Exact duplicates and training overlap.}
The duplicate fraction is one minus the number of distinct complete
sequences divided by the sample count. The exact training-match fraction
is the fraction of generated sequences occurring as contiguous length-128
segments anywhere in the full training text, including across stored row boundaries.
It does not test partial or approximate copying.

\paragraph{Character 4-gram divergence.}
Let $C_{\mathrm{gen}}(g)$ and $C_{\mathrm{ref}}(g)$ count overlapping
character 4-grams $g$ (including spaces) in generated samples and the
validation reference. With
\[
 \mathcal S=\{g:C_{\mathrm{gen}}(g)+C_{\mathrm{ref}}(g)>0\},
 \qquad \varepsilon=10^{-8},
\]
use the smoothed distributions
\[
 P(g)=\frac{C_{\mathrm{gen}}(g)+\varepsilon}
 {\sum_{h\in\mathcal S}C_{\mathrm{gen}}(h)+\varepsilon|\mathcal S|},
 \qquad
 Q(g)=\frac{C_{\mathrm{ref}}(g)+\varepsilon}
 {\sum_{h\in\mathcal S}C_{\mathrm{ref}}(h)+\varepsilon|\mathcal S|}
\]
and report
\[
 \operatorname{JSD}_4=\frac12\sum_{g\in\mathcal S}P(g)\log\frac{P(g)}{M(g)}
 +\frac12\sum_{g\in\mathcal S}Q(g)\log\frac{Q(g)}{M(g)},
 \qquad M(g)=\frac{P(g)+Q(g)}2.
\]
All logarithms are natural, so $0\le\operatorname{JSD}_4\le\log 2$.
Counts never cross generated-sample boundaries; the reference is contiguous
across stored chunks. The main 4,096-sample comparison and
Appendix~\ref{app:text8-confirmation} use validation characters
$[3{,}000{,}000:4{,}000{,}000]$. Unless stated otherwise, the separate
1,024-sample ablations use the first $10^6$ validation characters.
Compare scores only at matched reference slices and sample counts.

\paragraph{Aggregation across seeds.}
Compute each metric on each seed's complete sample set. For values $r_s$
from training seeds $s=0,1,2$, report $\bar r\pm\operatorname{SD}(r)$, where
\[
 \bar r=\frac13\sum_{s=0}^2r_s,
 \qquad
 \operatorname{SD}(r)=\sqrt{\frac12\sum_{s=0}^2(r_s-\bar r)^2}.
\]
All transformations, including the exponential defining $E_\ell$, precede
averaging. Report $V_\ell$ as percentages and round count
statistics only for display.

\subsection{Real-text calibration and supplementary diagnostics}
\label{app:text8-calibration}

Table~\ref{tab:text8-calibration} adds $V_6$ and real-text controls to the
main 4,096-sample comparison. \textcolor{black}{The eight main rows reuse samples from
Table~\ref{tab:text8-paths}; the additional RJD-LOO row applies TOPP
($p=0.9$) to reset destinations, retaining the cosine grid and
temperature-one raw LOO drift.} For real text, we take three
disjoint blocks of $128\times4096$ validation characters, starting at
offsets $1{,}100{,}000$, $1{,}700{,}000$, and $2{,}300{,}000$, and divide
each into length-128 passages. These blocks are disjoint from the common
4-gram reference $[3{,}000{,}000:4{,}000{,}000]$. We report mean and sample
standard deviation across training seeds for models and across text
blocks for the real controls.

\begin{table}[htbp]
\centering
\footnotesize
\setlength{\tabcolsep}{2.5pt}
\caption{Calibration on 4,096 length-128 sequences per replicate. Model
rows use three training seeds at 128 NFE; real text uses three validation
blocks. All rows share the dictionary and validation reference. Counts
are rounded to integers. \textcolor{black}{Model rows add $V_6$ to the main comparison and include
RJD-LOO with TOPP.}}
\label{tab:text8-calibration}
\label{tab:text8-cat-topp}
\begin{tabular}{lccccc}
\toprule
Method & $V_5$ (\%) & $V_6$ (\%) & $D_5$ & $E_5$ & $\operatorname{JSD}_4$ \\
\midrule
RJD & $28.96\pm 0.62$ & $17.54\pm 0.68$ & $3051\pm 43$ & $911\pm 21$ & $0.0657\pm 0.0007$ \\
RJD-LOO & $75.60\pm 0.89$ & $70.26\pm 1.08$ & $6596\pm 66$ & $2411\pm 48$ & $0.0580\pm 0.0011$ \\
RJD-BC-D & $60.16\pm 1.18$ & $51.59\pm 1.43$ & $6783\pm 13$ & $2843\pm 54$ & $0.0672\pm 0.0015$ \\
DFM & $44.62\pm 0.56$ & $34.34\pm 0.96$ & $4798\pm 159$ & $1501\pm 150$ & $0.0536\pm 0.0002$ \\
EPJ & $44.96\pm 0.47$ & $35.11\pm 0.56$ & $4971\pm 75$ & $1703\pm 73$ & $0.0540\pm 0.0002$ \\
PD & $29.47\pm 1.49$ & $18.29\pm 1.74$ & $3137\pm 191$ & $964\pm 83$ & $0.0664\pm 0.0005$ \\
PD-LOO & $85.49\pm 0.67$ & $80.55\pm 0.81$ & $5696\pm 58$ & $1944\pm 36$ & $0.0653\pm 0.0012$ \\
PD-BC & $36.34\pm 0.51$ & $24.27\pm 0.65$ & $4499\pm 7$ & $1637\pm 23$ & $0.0691\pm 0.0004$ \\
RJD-LOO, TOPP & $79.70\pm 0.68$ & $74.22\pm 0.88$ & $6611\pm 68$ & $2368\pm 29$ & $0.0545\pm 0.0009$ \\
\midrule
Real Text8 & $88.28\pm 0.85$ & $86.83\pm 1.35$ & $7204\pm 1055$ & $2380\pm 415$ & $0.0585\pm 0.0219$ \\
\bottomrule
\end{tabular}
\end{table}

RJD-BC-D has higher $D_5$ and $E_5$ than RJD-LOO, while RJD-LOO has
higher validity and lower four-gram divergence. TOPP improves RJD-LOO's
validity and divergence with similar $D_5$ but slightly lower $E_5$.
Real-text scores vary across blocks, and their four-gram divergence can
exceed that of generated sets; they do not define a universal lower bound.

Within each of the 27 generated sets of 4,096 samples, all complete
sequences are distinct, and none exactly matches a contiguous
128-character segment of the training text. This does not exclude partial
or approximate copying. Real-text controls have an exact training-match
fraction of $0.285\%\pm0.322\%$ across blocks.
These diagnostics concern lexical statistics, short-range character
agreement, and exact repetition, rather than grammaticality or coherence.

\FloatBarrier

\begingroup
\color{black}
\subsection{Text8 interventions with full-vocabulary completion}
\label{app:text8-interventions}

Using 60k-update PD-LOO and RJD-LOO predictors, we compare raw drift with
Bayesian completion. We replace a selected noisy
state with an independently corrupted state for a different donor
character $d$, initially preserving its surroundings, then continue all
positions to $t=0.999$ with 128 network calls.
Raw and completed variants share a LOO checkpoint. We use three training
seeds and 957 replacements per seed, each using a donor character different
from the original.

The immediate argmax uses the drift probabilities just after replacement:
$q_{\theta,i,t}$ for PD-LOO and RJD-LOO, or its likelihood completion
via \eqref{eq:completion} for PD-BC and RJD-BC-D. Both RJD variants
use raw LOO jump probabilities.

The local word check uses the Text8 test lexicon of all word lengths.
A letter's complete containing word, or both nonempty complete words
beside a space, must belong to it. Clipped or missing required words
remain in the denominator as unresolved cases. 

\begin{table}[htbp]
\color{black}
\centering
\small
\setlength{\tabcolsep}{3.5pt}
\caption{Text8 local-state replacement: percentages, mean $\pm$ sample
SD over three training seeds.
Bold marks the higher final word-check pass rate within each path family.
}
\label{tab:text8-intervention-completion}
\begin{tabular}{lrrrr}
\toprule
Outcome & PD-LOO & PD-BC & RJD-LOO & RJD-BC-D \\
\midrule
\multicolumn{5}{l}{Start $t=0.3013$} \\
Immediate argmax is $d$ & $1.95\pm0.61$ & $40.26\pm1.93$ & $1.67\pm0.38$ & $42.60\pm2.12$ \\
Final character is $d$ & $13.65\pm0.40$ & $31.52\pm1.09$ & $15.05\pm1.01$ & $16.06\pm0.34$ \\
Final local word check passes & $\mathbf{86.42}\pm0.83$ & $56.32\pm0.58$ & $\mathbf{78.16}\pm0.31$ & $69.80\pm1.10$ \\
\midrule
\multicolumn{5}{l}{Start $t=0.6973$} \\
Immediate argmax is $d$ & $0.21\pm0.10$ & $99.79\pm0.10$ & $0.07\pm0.06$ & $100.00\pm0.00$ \\
Final character is $d$ & $8.50\pm0.59$ & $99.83\pm0.06$ & $15.57\pm1.31$ & $20.72\pm2.28$ \\
Final local word check passes & $\mathbf{88.89}\pm0.54$ & $9.40\pm0.10$ & $\mathbf{75.51}\pm1.46$ & $63.22\pm0.79$ \\
\bottomrule
\end{tabular}
\end{table}

Replacing the local state leaves the immediate LOO prediction unchanged,
while completion can favor the donor through its new local likelihood
(Table~\ref{tab:text8-intervention-completion}). PD-BC tends to retain this
conflicting character, whereas RJD-BC-D can replace it through further resets.
Completion lowers final word-check pass rates in both families.
Nearly perfect recovery in unchanged and same-character controls
implicates conflicting character evidence rather than replacement noise
alone. These interventions suggest that Bayesian completion can hinder
recovery by reinforcing misleading local evidence.
\endgroup
\FloatBarrier

\subsection{Absorbing jump diffusion}
\label{app:text8-absorbing}

Absorbing jump diffusion (AJD) follows the time-changed Brownian bridge
until its first jump, then remains at the sampled embedding. Its rate is
$2t/(1-t^2)$ while unabsorbed. We compare unrestricted prediction in both
drift and jump probabilities (AJD), LOO prediction in both (AJD-LOO),
completion only in the drift (AJD-BC-D), and completion in both terms
(AJD-BC). The full predictor observes the absorption indicators;
$q_{\theta,i,t}$ excludes both $X_{i,t}$ and its indicator. Completion uses
the local likelihood of this absorbing path.

Table~\ref{tab:text8-paths-complete} compares the four AJD variants at
128 NFE, using 1,024 samples per seed and the first million validation
characters as reference. LOO improves all four reported metrics over
unrestricted AJD and both completion variants. This is a separate
prediction ablation; its $D_5$ and $\operatorname{JSD}_4$ values should not
be compared directly with the 4,096-sample main table.

\begin{table}[htbp]
\centering

\setlength{\tabcolsep}{4pt}
\caption{AJD prediction ablation at 128 NFE and 1,024 samples per seed. We report mean $\pm$ sample standard deviation over three training seeds; $D_5$ counts are rounded to the nearest integer. All paths use $\kappa(t)=t^2$, jumps use CAT, and predictions have temperature one.}
\label{tab:text8-paths-complete}
\begin{tabular}{lcccc}
\toprule
Method & $V_5$ (\%) $\uparrow$ & $V_6$ (\%) $\uparrow$ & $D_5$ $\uparrow$ & $\operatorname{JSD}_4$ $\downarrow$ \\
\midrule
AJD & $29.37\pm 2.36$ & $18.38\pm 2.70$ & $1339\pm 97$ & $0.0919\pm 0.0020$ \\
AJD-LOO & $\mathbf{68.62}\pm 0.34$ & $\mathbf{58.75}\pm 0.43$ & $\mathbf{2910}\pm 77$ & $\mathbf{0.0761}\pm 0.0014$ \\
AJD-BC-D & $53.68\pm 0.89$ & $41.71\pm 1.08$ & $2722\pm 42$ & $0.0879\pm 0.0009$ \\
AJD-BC & $35.86\pm 0.08$ & $23.70\pm 0.46$ & $1816\pm 16$ & $0.0947\pm 0.0003$ \\
\bottomrule
\end{tabular}
\end{table}

\FloatBarrier

\subsection{Jump-time schedules for absorbing jump diffusion}
\label{app:text8-clock-details}
\label{sec:text8-clocks}

Keeping the continuous schedule $\kappa(t)=t^2$ fixed, we compare independent
first-jump times $T_i$ with
\[
 \mathbb P(T_i\le t)=t^\gamma,
 \qquad \lambda_\gamma(t)=\frac{\gamma t^{\gamma-1}}{1-t^\gamma},
 \qquad \gamma\in\{1,2,4\}.
\]
The baseline is $\gamma=2$; larger $\gamma$ delays jumps. For absorbing jump
diffusion, each model is trained and sampled with the same $\gamma$.
An unabsorbed coordinate jumps over $[s,t]$ with probability
$(t^\gamma-s^\gamma)/(1-s^\gamma)$. We keep the cosine time grid and network
budget fixed, stop at $t=.999$, and apply the chosen token rule to any
remaining unabsorbed positions, counting terminal network evaluations.

At 128 NFE, delaying absorption increases validity and diversity:
$V_5$ rises from $58.07\%$ to $77.10\%$ between $\gamma=1$ and $4$.
The lowest mean $\operatorname{JSD}_4$ occurs at $\gamma=2$, close to
$\gamma=4$ (Table~\ref{tab:text8-clocks}).

\begin{table}[htbp]
\centering

\setlength{\tabcolsep}{4pt}
\caption{AJD-LOO with CAT, 128 NFE, and 1,024 samples per seed (mean $\pm$ sample standard deviation over three seeds). $D_5$ counts are rounded to the nearest integer. The continuous schedule is fixed at $\kappa(t)=t^2$; the $\gamma=1,4$ models are trained with their respective jump clocks.}
\label{tab:text8-clocks}
\begin{tabular}{ccccc}
\toprule
$\gamma$ & $V_5$ (\%) $\uparrow$ & $V_6$ (\%) $\uparrow$ & $D_5$ $\uparrow$ & $\operatorname{JSD}_4$ $\downarrow$ \\
\midrule
$1$ & $58.07\pm 1.15$ & $46.86\pm 0.71$ & $2749\pm 64$ & $0.0792\pm 0.0013$ \\
$2$ & $68.62\pm 0.34$ & $58.75\pm 0.43$ & $2910\pm 77$ & $\mathbf{0.0761}\pm 0.0014$ \\
$4$ & $\mathbf{77.10}\pm 0.83$ & $\mathbf{69.44}\pm 1.08$ & $\mathbf{3064}\pm 13$ & $0.0766\pm 0.0007$ \\
\bottomrule
\end{tabular}
\end{table}

Table~\ref{tab:text8-clock-budgets} supplements the 128-NFE clock comparison
in Table~\ref{tab:text8-clocks}. Increasing $\gamma$ improves $V_5$ at both
additional budgets. At 32 NFE, $\gamma=4$ gives a slightly lower mean
$\operatorname{JSD}_4$ than $\gamma=2$; at 512 NFE, $\gamma=2$ gives the
lowest mean. The continuous schedule and numerical grid remain
fixed. At the common cutoff, the theoretical unabsorbed fractions are
$1-.999^\gamma$, approximately $0.001$, $0.002$, and $0.004$ for
$\gamma=1,2,4$; terminal decoding therefore affects different fractions.

\begin{table}[htbp]
\centering
\small
\setlength{\tabcolsep}{4pt}
\caption{Additional budgets for AJD-LOO with CAT and trained jump CDF $t^\gamma$. Mean $\pm$ sample standard deviation over three seeds, with 1,024 samples per seed and fixed continuous schedule $\kappa(t)=t^2$.}
\label{tab:text8-clock-budgets}
\begin{tabular}{ccccc}
\toprule
NFE & $\gamma$ & $V_5$ (\%) $\uparrow$ & $V_6$ (\%) $\uparrow$ & $\operatorname{JSD}_4$ $\downarrow$ \\
\midrule
32 & $1$ & $50.72\pm 0.41$ & $39.07\pm 0.70$ & $0.0887\pm 0.0012$ \\
32 & $2$ & $59.05\pm 0.62$ & $48.39\pm 0.77$ & $0.0833\pm 0.0013$ \\
32 & $4$ & $\mathbf{67.85}\pm 1.35$ & $\mathbf{59.32}\pm 1.85$ & $\mathbf{0.0830}\pm 0.0010$ \\
\midrule
512 & $1$ & $60.32\pm 0.10$ & $49.04\pm 0.17$ & $0.0781\pm 0.0002$ \\
512 & $2$ & $70.06\pm 1.04$ & $60.35\pm 1.04$ & $\mathbf{0.0750}\pm 0.0005$ \\
512 & $4$ & $\mathbf{78.53}\pm 0.72$ & $\mathbf{71.09}\pm 1.11$ & $0.0765\pm 0.0007$ \\
\bottomrule
\end{tabular}
\end{table}

\FloatBarrier

\subsection{Jump sampling rules and complete comparisons}
\label{app:text8-jump-sampling}
\label{sec:text8-sampling}

\begingroup
\color{black}
Five jump rules reuse each path's 60,000-update EMA checkpoints.
Destination rules are categorical sampling
$A_i\sim q_{\theta,i,t}$ (CAT), $A_i=\arg\max_a q_{\theta,i,t}(a)$ (MAX),
probabilities proportional to $q_{\theta,i,t}(a)^{1/0.7}$ (TEMP), and
categorical sampling from the smallest set of highest-probability tokens
with total mass at least $0.9$ (TOPP, or nucleus sampling)
\citep[Sections~3.1 and~3.3]{holtzman2020degeneration}.
Confidence selection (CONF) reassigns the positions selected by independent
clocks to those with largest $\max_a q_{\theta,i,t}(a)$, then samples
destinations categorically, adapting Mask-Predict
\citep{ghazvininejad2019maskpredict} and MaskGIT
\citep{chang2022maskgit}. AJD-LOO and RJD-LOO retain the temperature-one raw
LOO drift for every rule. The modified rules change the generator.

Comparisons use 1,024 length-128 sequences per seed, the same dictionary,
the first million validation characters as reference, and the cosine grid.
NFE includes EPJ and AJD-LOO terminal prediction calls for unabsorbed
positions; RJD-LOO decodes to nearest embeddings at $t=.999$ without
an extra call. TOPP improves validity and four-gram agreement across the
tested paths; the effects of TEMP and CONF depend on the path and budget.
Appendices~\ref{app:text8-budgets} and~\ref{app:text8-confirmation} give
additional NFE budgets and other samplers with 4,096 samples per seed.

\endgroup

\paragraph{Confidence selection.}
For EPJ and AJD-LOO, CONF draws the number of jumps from the original
independent clocks and assigns them to the most confident unabsorbed
positions; committed positions remain fixed. For RJD-LOO, all positions
remain eligible on every interval. On the path-time interval $[u,v]$,
we first sample the independent reset indicators and last-reset times from
Appendix~\ref{app:sampler}. Their count satisfies
\[
 K\sim\operatorname{Binomial}\!\left(L,\frac{v-u}{1-u}\right).
\]
Here $K$ counts positions with at least one reset, not the total number of
Poisson events. CONF chooses the $K$ positions with largest
$\max_a q_{\theta,i,t}(a)$, breaking ties with fresh independent uniforms,
and reassigns the sampled last-reset times to them. The multiset of times
is preserved, and their assignment does not depend on the time values.
The interval update \eqref{eq:interval-update} then includes diffusion
after each reset. The drift uses the unchanged LOO mean, rather than the
last sampled token as a hidden endpoint. This finite-grid rule preserves
the sampled count but changes the position-specific reset probabilities;
it is not a marginal-preserving modification.

At 128 NFE, CAT and CONF both produce about $4.19$ reset-containing
intervals per position. The fraction of positions reset in multiple
intervals is $96.5\%$ for CAT and $48.9\%$ for CONF; CONF's selected
positions have mean predicted confidence $0.994$. Thus CONF concentrates
resets on fewer positions. This is consistent with, but does not establish
the cause of, the lower word diversity. These diagnostics count intervals
with resets, not individual Poisson events within an interval.

\paragraph{Complete sampling results.}
Tables~\ref{tab:text8-epj-sampling-all}--\ref{tab:text8-rjd-sampling-all}
report all five rules, including the effective word count $E_5$.
At 128 NFE, AJD-LOO with TOPP has lower $\operatorname{JSD}_4$ than
RJD-LOO with TOPP, but lower validity and diversity; neither path improves
every metric.
EPJ and AJD-LOO have all three training seeds at every budget.
For the RJD-LOO sampling study, 32 and 128 NFE have seeds 0, 1, and 2;
the 512-NFE results use only seeds 0 and 1.
The separate CAT path comparison in
Table~\ref{tab:text8-additional-budgets} has all three seeds.

At 128 NFE, MAX raises RJD-LOO validity over CAT but reduces $E_5$ by
about $25\%$; CONF reduces it by about $19\%$.
None of the RJD-LOO evaluations contains duplicate full sequences;
the reduction concerns word diversity.

For budget comparisons involving 512 NFE, we match seeds 0 and 1 at every
budget. RJD-LOO with TOPP has $V_5=70.27\%,79.96\%,82.31\%$ and
$\operatorname{JSD}_4=0.0874,0.0771,0.0750$ at 32, 128, and 512 NFE.
At 512 NFE, PD-LOO on these same two seeds has $V_5=85.37\%$,
$D_5=2904$, $E_5=1432$, and $\operatorname{JSD}_4=0.0811$, compared with
$D_5=3210$ and $E_5=1654$ for RJD-LOO with TOPP. The tradeoff remains;
these two-seed estimates should not be compared directly with the
three-seed means as if the seed sets were identical.

\begin{table}[htbp]
\centering
\small
\setlength{\tabcolsep}{4pt}
\caption{EPJ sampling at 32, 128, and 512 NFE with 1,024 sequences per seed. Mean $\pm$ sample standard deviation; count metrics are rounded to integers. Best means within each budget are bold. All budgets use three training seeds.}
\label{tab:text8-epj-sampling-all}
\begin{tabular}{lccccc}
\toprule
Rule & $V_5$ (\%) $\uparrow$ & $V_6$ (\%) $\uparrow$ & $D_5$ $\uparrow$ & $E_5$ $\uparrow$ & $\operatorname{JSD}_4$ $\downarrow$ \\
\midrule
\multicolumn{6}{l}{32 NFE, 3 seeds} \\
CAT & $35.22\pm 0.50$ & $26.01\pm 0.32$ & $1804\pm 20$ & $967\pm 8$ & $0.0996\pm 0.0007$ \\
TEMP & $53.05\pm 0.47$ & $41.59\pm 0.66$ & $1915\pm 60$ & $695\pm 46$ & $0.0875\pm 0.0015$ \\
TOPP & $44.19\pm 0.45$ & $34.60\pm 0.59$ & $\mathbf{2021}\pm 29$ & $\mathbf{1020}\pm 47$ & $\mathbf{0.0856}\pm 0.0005$ \\
MAX & $\mathbf{90.58}\pm 0.74$ & $\mathbf{76.16}\pm 1.47$ & $363\pm 14$ & $13\pm 1$ & $0.3708\pm 0.0034$ \\
CONF & $9.78\pm 0.15$ & $5.86\pm 0.29$ & $665\pm 37$ & $452\pm 42$ & $0.2223\pm 0.0016$ \\
\midrule
\multicolumn{6}{l}{128 NFE, 3 seeds} \\
CAT & $44.75\pm 0.41$ & $34.85\pm 0.59$ & $2139\pm 47$ & $\mathbf{1098}\pm 46$ & $0.0817\pm 0.0006$ \\
TEMP & $64.52\pm 0.49$ & $53.17\pm 0.64$ & $2117\pm 25$ & $712\pm 42$ & $0.0825\pm 0.0009$ \\
TOPP & $55.95\pm 0.91$ & $45.92\pm 1.18$ & $\mathbf{2323}\pm 42$ & $1092\pm 31$ & $\mathbf{0.0722}\pm 0.0006$ \\
MAX & $\mathbf{96.87}\pm 0.29$ & $\mathbf{89.50}\pm 1.53$ & $329\pm 36$ & $11\pm 2$ & $0.3932\pm 0.0104$ \\
CONF & $26.24\pm 0.82$ & $19.23\pm 0.63$ & $1393\pm 45$ & $720\pm 25$ & $0.1397\pm 0.0002$ \\
\midrule
\multicolumn{6}{l}{512 NFE, 3 seeds} \\
CAT & $47.58\pm 0.57$ & $37.56\pm 0.88$ & $2220\pm 42$ & $1131\pm 86$ & $0.0776\pm 0.0005$ \\
TEMP & $68.25\pm 0.37$ & $57.23\pm 0.69$ & $2192\pm 45$ & $718\pm 54$ & $0.0817\pm 0.0024$ \\
TOPP & $58.78\pm 0.14$ & $49.03\pm 0.34$ & $\mathbf{2427}\pm 36$ & $\mathbf{1149}\pm 32$ & $\mathbf{0.0705}\pm 0.0005$ \\
MAX & $\mathbf{98.22}\pm 0.23$ & $\mathbf{93.09}\pm 0.26$ & $309\pm 3$ & $10\pm 0$ & $0.4025\pm 0.0058$ \\
CONF & $49.74\pm 0.11$ & $41.54\pm 0.93$ & $2046\pm 48$ & $842\pm 95$ & $0.0899\pm 0.0018$ \\
\bottomrule
\end{tabular}
\end{table}

\FloatBarrier

\begin{table}[htbp]
\centering
\small
\setlength{\tabcolsep}{4pt}
\caption{AJD-LOO sampling at 32, 128, and 512 NFE with 1,024 sequences per seed. Mean $\pm$ sample standard deviation; count metrics are rounded to integers. Best means within each budget are bold. All budgets use three training seeds.}
\label{tab:text8-ajd-sampling-all}
\begin{tabular}{lccccc}
\toprule
Rule & $V_5$ (\%) $\uparrow$ & $V_6$ (\%) $\uparrow$ & $D_5$ $\uparrow$ & $E_5$ $\uparrow$ & $\operatorname{JSD}_4$ $\downarrow$ \\
\midrule
\multicolumn{6}{l}{32 NFE, 3 seeds} \\
CAT & $58.94\pm 0.68$ & $48.19\pm 0.95$ & $2796\pm 29$ & $1567\pm 23$ & $0.0833\pm 0.0013$ \\
TEMP & $66.04\pm 1.12$ & $55.44\pm 1.82$ & $2859\pm 90$ & $1481\pm 70$ & $\mathbf{0.0774}\pm 0.0015$ \\
TOPP & $63.48\pm 0.41$ & $52.89\pm 1.08$ & $\mathbf{2904}\pm 29$ & $\mathbf{1578}\pm 43$ & $0.0776\pm 0.0011$ \\
MAX & $74.52\pm 0.21$ & $64.19\pm 0.31$ & $2640\pm 68$ & $1162\pm 27$ & $0.0845\pm 0.0006$ \\
CONF & $\mathbf{77.41}\pm 0.93$ & $\mathbf{70.26}\pm 1.01$ & $2828\pm 65$ & $1300\pm 81$ & $0.0813\pm 0.0013$ \\
\midrule
\multicolumn{6}{l}{128 NFE, 3 seeds} \\
CAT & $68.34\pm 0.42$ & $58.47\pm 0.39$ & $2924\pm 42$ & $\mathbf{1508}\pm 31$ & $0.0758\pm 0.0003$ \\
TEMP & $73.41\pm 0.52$ & $63.87\pm 0.42$ & $2896\pm 23$ & $1392\pm 20$ & $0.0743\pm 0.0012$ \\
TOPP & $71.75\pm 0.73$ & $62.17\pm 0.75$ & $\mathbf{2942}\pm 41$ & $1484\pm 24$ & $\mathbf{0.0729}\pm 0.0007$ \\
MAX & $80.26\pm 0.17$ & $71.41\pm 0.67$ & $2650\pm 36$ & $1078\pm 29$ & $0.0830\pm 0.0021$ \\
CONF & $\mathbf{82.75}\pm 0.20$ & $\mathbf{76.12}\pm 0.28$ & $2882\pm 55$ & $1264\pm 52$ & $0.0807\pm 0.0025$ \\
\midrule
\multicolumn{6}{l}{512 NFE, 3 seeds} \\
CAT & $69.80\pm 1.26$ & $60.05\pm 1.27$ & $2929\pm 28$ & $\mathbf{1443}\pm 30$ & $0.0745\pm 0.0006$ \\
TEMP & $75.17\pm 0.98$ & $65.90\pm 1.16$ & $2869\pm 5$ & $1339\pm 20$ & $0.0740\pm 0.0007$ \\
TOPP & $73.35\pm 0.95$ & $63.87\pm 1.19$ & $\mathbf{2958}\pm 8$ & $1424\pm 37$ & $\mathbf{0.0720}\pm 0.0009$ \\
MAX & $81.75\pm 1.10$ & $73.18\pm 1.29$ & $2625\pm 35$ & $1052\pm 44$ & $0.0845\pm 0.0017$ \\
CONF & $\mathbf{83.69}\pm 0.60$ & $\mathbf{76.83}\pm 0.68$ & $2886\pm 61$ & $1216\pm 59$ & $0.0818\pm 0.0022$ \\
\bottomrule
\end{tabular}
\end{table}

\FloatBarrier

\begin{table}[htbp]
\centering
\small
\setlength{\tabcolsep}{4pt}
\caption{RJD-LOO sampling at 32, 128, and 512 NFE with 1,024 sequences per seed. Mean $\pm$ sample standard deviation; count metrics are rounded to integers. Best means within each budget are bold. The 32- and 128-NFE results use three seeds; 512 NFE uses only seeds 0 and 1. The nonzero CAT standard deviation for $\operatorname{JSD}_4$ at 512 NFE is shown in scientific notation.}
\label{tab:text8-rjd-sampling-all}
\label{tab:text8-decoding}
\begin{tabular}{lccccc}
\toprule
Rule & $V_5$ (\%) $\uparrow$ & $V_6$ (\%) $\uparrow$ & $D_5$ $\uparrow$ & $E_5$ $\uparrow$ & $\operatorname{JSD}_4$ $\downarrow$ \\
\midrule
\multicolumn{6}{l}{32 NFE, 3 seeds} \\
CAT & $64.39\pm 1.34$ & $58.06\pm 1.28$ & $2819\pm 29$ & $1597\pm 48$ & $0.0979\pm 0.0024$ \\
TEMP & $72.70\pm 1.00$ & $65.82\pm 1.12$ & $\mathbf{2963}\pm 33$ & $1601\pm 34$ & $0.0857\pm 0.0013$ \\
TOPP & $69.90\pm 0.86$ & $63.34\pm 1.06$ & $2960\pm 36$ & $\mathbf{1654}\pm 58$ & $0.0882\pm 0.0013$ \\
MAX & $\mathbf{80.64}\pm 0.71$ & $\mathbf{73.55}\pm 0.96$ & $2734\pm 29$ & $1282\pm 65$ & $0.0865\pm 0.0017$ \\
CONF & $77.20\pm 0.91$ & $71.03\pm 1.64$ & $2718\pm 8$ & $1312\pm 42$ & $\mathbf{0.0847}\pm 0.0009$ \\
\midrule
\multicolumn{6}{l}{128 NFE, 3 seeds} \\
CAT & $75.76\pm 1.04$ & $70.61\pm 1.26$ & $3118\pm 41$ & $\mathbf{1669}\pm 42$ & $0.0824\pm 0.0018$ \\
TEMP & $81.23\pm 0.38$ & $75.63\pm 0.60$ & $3100\pm 43$ & $1576\pm 40$ & $\mathbf{0.0775}\pm 0.0016$ \\
TOPP & $79.52\pm 0.89$ & $74.10\pm 0.77$ & $\mathbf{3149}\pm 13$ & $1659\pm 21$ & $0.0776\pm 0.0010$ \\
MAX & $\mathbf{86.29}\pm 0.38$ & $\mathbf{80.20}\pm 0.40$ & $2814\pm 66$ & $1250\pm 103$ & $0.0834\pm 0.0010$ \\
CONF & $84.21\pm 0.41$ & $78.23\pm 0.65$ & $2876\pm 12$ & $1352\pm 26$ & $0.0801\pm 0.0005$ \\
\midrule
\multicolumn{6}{l}{512 NFE, 2 seeds} \\
CAT & $79.22\pm 0.10$ & $74.47\pm 0.22$ & $3203\pm 1$ & $\mathbf{1689}\pm 4$ & $0.0781\pm 4.19\!\times\!10^{-6}$ \\
TEMP & $83.06\pm 0.37$ & $77.82\pm 0.27$ & $3143\pm 18$ & $1558\pm 22$ & $0.0755\pm 0.0011$ \\
TOPP & $82.31\pm 0.76$ & $77.29\pm 0.85$ & $\mathbf{3210}\pm 4$ & $1654\pm 9$ & $\mathbf{0.0750}\pm 0.0004$ \\
MAX & $\mathbf{86.85}\pm 0.54$ & $\mathbf{80.94}\pm 0.38$ & $2776\pm 83$ & $1209\pm 116$ & $0.0826\pm 0.0016$ \\
CONF & $85.92\pm 0.28$ & $80.36\pm 0.85$ & $2852\pm 62$ & $1330\pm 88$ & $0.0793\pm 0.0016$ \\
\bottomrule
\end{tabular}
\end{table}

\FloatBarrier

\subsection{Additional network budgets}
\label{app:text8-budgets}

Table~\ref{tab:text8-additional-budgets} gives the complete comparison at
512 NFE and a partial comparison at 2,048 NFE. DFM, AJD-BC-D, and RJD-LOO were not
evaluated at 2,048 NFE. There is no complete comparison at
32 NFE: PD, PD-BC, AJD, AJD-BC, AJD-BC-D, RJD, and RJD-BC-D have no such
evaluation in this set; the separate RJD-LOO sampling results are in
Table~\ref{tab:text8-rjd-sampling-all}. All values below use 1,024 samples per seed.
At 512 NFE, RJD-LOO has $D_5=3193\pm17$, compared with $2939\pm23$ for
AJD-LOO and $3209\pm23$ for RJD-BC-D. Thus the validity advantage of
RJD-LOO over RJD-BC-D does not imply higher word diversity: RJD-BC-D has
slightly higher $D_5$ and higher $E_5$ at this budget
($1911\pm22$ versus $1686\pm6$).

\begin{table}[htbp]
\centering
\small
\setlength{\tabcolsep}{4pt}
\caption{Additional budgets for the conditional-path comparison. Mean $\pm$ sample standard deviation over three seeds, with 1,024 samples per seed, temperature one, CAT, and $\kappa(t)=t^2$. All checkpoints are evaluated after 60,000 training updates.}
\label{tab:text8-additional-budgets}
\begin{tabular}{clccc}
\toprule
NFE & Method & $V_5$ (\%) $\uparrow$ & $V_6$ (\%) $\uparrow$ & $\operatorname{JSD}_4$ $\downarrow$ \\
\midrule
512 & DFM & $47.09\pm 0.37$ & $36.47\pm 0.83$ & $0.0787\pm 0.0018$ \\
512 & EPJ & $47.52\pm 0.92$ & $37.42\pm 1.27$ & $0.0778\pm 0.0002$ \\
512 & PD & $31.48\pm 1.30$ & $20.10\pm 1.71$ & $0.0898\pm 0.0016$ \\
512 & PD-LOO & $\mathbf{86.10}\pm 1.29$ & $\mathbf{81.12}\pm 1.70$ & $0.0817\pm 0.0012$ \\
512 & PD-BC & $37.59\pm 0.34$ & $25.78\pm 0.40$ & $0.0938\pm 0.0003$ \\
512 & AJD & $31.51\pm 1.21$ & $20.06\pm 1.84$ & $0.0885\pm 0.0023$ \\
512 & AJD-LOO & $70.06\pm 1.04$ & $60.35\pm 1.04$ & $\mathbf{0.0750}\pm 0.0005$ \\
512 & AJD-BC-D & $54.96\pm 0.49$ & $43.26\pm 0.20$ & $0.0856\pm 0.0004$ \\
512 & AJD-BC & $37.62\pm 0.82$ & $25.47\pm 1.04$ & $0.0929\pm 0.0007$ \\
512 & RJD & $29.89\pm 0.66$ & $18.76\pm 1.21$ & $0.0897\pm 0.0003$ \\
512 & RJD-LOO & $78.57\pm 1.12$ & $73.56\pm 1.59$ & $0.0793\pm 0.0021$ \\
512 & RJD-BC-D & $65.31\pm 1.59$ & $57.24\pm 1.84$ & $0.0909\pm 0.0027$ \\
\midrule
2048 & EPJ & $48.57\pm 0.14$ & $38.50\pm 0.61$ & $0.0764\pm 0.0002$ \\
2048 & PD & $31.24\pm 1.67$ & $19.63\pm 2.09$ & $0.0891\pm 0.0014$ \\
2048 & PD-LOO & $\mathbf{85.70}\pm 0.82$ & $\mathbf{80.88}\pm 1.07$ & $0.0808\pm 0.0021$ \\
2048 & PD-BC & $38.58\pm 0.60$ & $26.53\pm 0.19$ & $0.0934\pm 0.0005$ \\
2048 & AJD & $32.39\pm 2.10$ & $20.44\pm 2.31$ & $0.0891\pm 0.0025$ \\
2048 & AJD-LOO & $70.21\pm 0.52$ & $60.76\pm 0.75$ & $\mathbf{0.0739}\pm 0.0010$ \\
2048 & AJD-BC & $38.26\pm 0.38$ & $25.61\pm 0.43$ & $0.0929\pm 0.0003$ \\
2048 & RJD & $30.70\pm 0.58$ & $19.05\pm 0.64$ & $0.0891\pm 0.0006$ \\
2048 & RJD-BC-D & $68.77\pm 0.91$ & $60.97\pm 1.58$ & $0.0864\pm 0.0026$ \\
\bottomrule
\end{tabular}
\end{table}

At 128 NFE, EPJ with MAX gives $96.87\%$ word validity but only about 11
effective valid-word types, compared with about 1,098 for CAT. Here the
effective count is $\exp(H)$, where $H$ is the empirical entropy of valid
words of length at least five; it is calculated per seed before averaging.
This concentration explains why a high $V_5$ need not imply a close match
to the reference text distribution.

\FloatBarrier

\subsection{Additional samplers with 4,096 samples per seed}
\label{app:text8-confirmation}

This separate evaluation covers five sampler settings, the same three
training seeds, and budgets of 128 and 512 NFE. All checkpoints are the
60,000-update EMA checkpoints. We generate 4,096 sequences per seed with
generation seeds $51000+s$, where $s\in\{0,1,2\}$ is the training seed.
The dictionary and validation reference
$[3{,}000{,}000:4{,}000{,}000]$ match the main comparison, but the grids,
jump rules, and correction steps below differ. These are complete-sampler
comparisons using existing checkpoints, not additional training seeds.

PD-LOO uses a grid uniform in $\kappa(t)=t^2$; the other settings use the
cosine grid. All AJD variants retain raw LOO drift. The final setting
adds 16 LOO updates at times in $[2/3,1]$: each update chooses one position
per sequence, draws its token from $q_{\theta,i,t}$, and redraws its complete
conditional state, including its absorption indicator. These are 16
single-position updates, not 16 sweeps over the sequence. At total budget
$B$, this setting uses $B-17$ predictor intervals, 16 correction calls,
and one terminal call. The selected settings are fixed before this
confirmation; they compare complete samplers rather than isolating one
change to the conditional path.

\begin{table}[htbp]
\centering
\small
\setlength{\tabcolsep}{4pt}
\caption{Additional samplers with 4,096 generated sequences per seed. Mean $\pm$ sample standard deviation over three training seeds; the dictionary and 4-gram reference match the main comparison. All correction and terminal calls count toward NFE.}
\label{tab:text8-confirmation}
\begin{tabular}{clcc}
\toprule
NFE & Sampler & $V_5$ (\%) $\uparrow$ & $\operatorname{JSD}_4$ $\downarrow$ \\
\midrule
128 & PD-LOO, uniform $\kappa$ & $\mathbf{85.48}\pm 0.75$ & $0.0657\pm 0.0011$ \\
128 & EPJ, TOPP & $55.72\pm 0.58$ & $0.0508\pm 0.0009$ \\
128 & AJD-LOO, $\gamma=2$, TOPP & $72.05\pm 0.35$ & $\mathbf{0.0487}\pm 0.0002$ \\
128 & AJD-LOO, $\gamma=4$, TOPP & $79.00\pm 0.93$ & $0.0543\pm 0.0008$ \\
128 & AJD-LOO, $\gamma=4$, CONF + 16 updates & $82.87\pm 0.78$ & $0.0582\pm 0.0012$ \\
\midrule
512 & PD-LOO, uniform $\kappa$ & $\mathbf{85.86}\pm 0.72$ & $0.0646\pm 0.0019$ \\
512 & EPJ, TOPP & $58.81\pm 0.56$ & $0.0487\pm 0.0001$ \\
512 & AJD-LOO, $\gamma=2$, TOPP & $73.82\pm 0.60$ & $\mathbf{0.0486}\pm 0.0005$ \\
512 & AJD-LOO, $\gamma=4$, TOPP & $80.33\pm 0.67$ & $0.0542\pm 0.0007$ \\
512 & AJD-LOO, $\gamma=4$, CONF + 16 updates & $83.54\pm 0.95$ & $0.0576\pm 0.0015$ \\
\bottomrule
\end{tabular}
\end{table}

\FloatBarrier

\clearpage
\subsection{Generated Text8 samples}
\label{app:text8-samples}

The panels below show the first four generated sequences from training
seed~0 for each model in Table~\ref{tab:text8-calibration}, using the
60,000-update EMA checkpoints and 128 NFE. They come from the first batch,
with generation seed $625000$, of the main 4,096-sample evaluations.
The real-text panel shows the first four passages from the first
calibration block. Every sequence has 128 characters; spelling and spacing
are retained, with line wrapping only for display. No sample is selected
by quality. Text8 chunks may begin or end inside a word and need not
contain complete sentences.

\begingroup
\lstset{basicstyle=\footnotesize\ttfamily,columns=fullflexible,
  keepspaces=true,showstringspaces=false,breaklines=true,
  breakatwhitespace=true,numbers=left,firstnumber=1,numberstyle=\tiny,
  numbersep=7pt,xleftmargin=1.5em,aboveskip=4pt,belowskip=0pt}

\noindent\begin{minipage}{\linewidth}
\textbf{RJD}
\begin{lstlisting}
e dumb and viner moders in londom ishue eight seven true tradicated in rather from the condect one nine isles that citigating an
 sex mperet copics miles the hzrio seven with e glor out and afforms what pertihally obstylor dissogovernation has game millions
ore s for scends are and agoin profound whone in the work irs and it christ of alkind with the pus assi son the moduls awice sho
s soon an ive a rind th for diale orseural our eight freed basc time samier unite forcrection was indual numes two zerological w
\end{lstlisting}
\end{minipage}
\par\medskip

\noindent\begin{minipage}{\linewidth}
\textbf{RJD-LOO}
\begin{lstlisting}
lm an increased at never until one nine zero zero s missed into the from the germans seven the size with their but conditions in
ne nine from the similar mey lije finding in treaty are federatorial from master others leading system and and so two spirit in 
c wicked as they three poland beyond lanes to with city award and stand its black ards s hulm premiere of kelrey could arrive wo
lt at the bbc s case of diamond one four three singer had a links situationes the bhm trishment both saturated in twins in the t
\end{lstlisting}
\end{minipage}
\par\medskip

\noindent\begin{minipage}{\linewidth}
\textbf{RJD-BC-D}
\begin{lstlisting}
dm and the six nine microbably the much five zero s protestants the from the one one depart on mars with their dint eduction sym
ne five root the similar taxation finding of traits are need italian crime mo orname ballistine system appsara a yff cld nations
nglabour is that finally have a sack antident with no times terrain him one which own a coast julians cpu of continues refer com
t wae the end a core or ted airlinear used to exclaim nsiek manuse igration scales greenish will tried segwod war complete the r
\end{lstlisting}
\end{minipage}
\par\medskip

\noindent\begin{minipage}{\linewidth}
\textbf{DFM}
\begin{lstlisting}
the geoglopes have cpublic grand children cummon from any alternet tee also goes a mitchem abablical hugaleth three of covement 
lly even products while is olderchears and dandry minstur five last invaserabolations und novel tamp artial and theory prejertiv
ked place the icon unsoruence a like sotrague are excell museumus very proper j five zero fivr kx three eighteen agree mcass as 
gly class for no fish race humans batant tu merolesc french chances of inderstand s s movonecy but estment of the acid and as e 
\end{lstlisting}
\end{minipage}
\par\medskip

\noindent\begin{minipage}{\linewidth}
\textbf{EPJ}
\begin{lstlisting}
ch proootypes whom the diccs and children cummon lock and alternet the also goes a discreas biblical quesleth three of constant 
lly john products would be kide cietre and landon the tur five last invoker word it sound constant pourthel in its of t rkjers a
ked electrolysi of unborusics a like sourage earloes allo usedmus very proper s live articuiar exported either aka meen carious 
g include for no fish race humans balant toe epollic speech chances of inderstand s complexacy bettes term of the acid and acle 
\end{lstlisting}
\end{minipage}
\par\medskip

\noindent\begin{minipage}{\linewidth}
\textbf{PD}
\begin{lstlisting}
ellow willing printers used an a host qz o rge work wood the suveriations in use at mouke the devells web or the taffairable pun
ines technical uperators as a justicially electrbst delphan s also litylcha tustman pathinal his most zero six somes posite diag
n to athmic as long a can carediamed in waftgaivil that all ardent of whichment of comleptitiea buduals targe base it beespresen
rm nately of brone for a i to st author of bell or or terrable or a masteri s anotherain of two time construttle golgia of aomon
\end{lstlisting}
\end{minipage}
\par\medskip

\noindent\begin{minipage}{\linewidth}
\textbf{PD-LOO}
\begin{lstlisting}
ther way if and is required and his three in the much lead the unable two zero with fresco known explants botany where bible by 
in combinatings and known incomes with more volume turkiest south the faced back and from a pictures functive in a zero six foll
 family concerned with led the handpoint and of break train later and german is in nine door is violence west for three the part
 selling roman holding deputies the threats the root of british unificialized in the official companion islanding man and four f
\end{lstlisting}
\end{minipage}
\par\medskip

\noindent\begin{minipage}{\linewidth}
\textbf{PD-BC}
\begin{lstlisting}
ry arvasion god be nasourew as whichborre with robbonder external planing were highening of sean explated but onothers libdiness
ided mmigh mary and brown in one white clocter over rarring years howevered contrast s am ibrans intemulceptic to where firstwir
eft day converses logist diated a insult and required for foltwerna decamillies on the amate native abcoaster togethrew the land
mportational civil its families one three e well of hopes uzes it inviding to in the old that cormits is standard but ael founli
\end{lstlisting}
\end{minipage}
\par\medskip

\noindent\begin{minipage}{\linewidth}
\textbf{RJD-LOO, TOPP}
\begin{lstlisting}
cd and spreaded an generatical one nine zero zero somerged into the from the germans seven the life with their all conditions in
ne nine from the similar may life finding in treaty are federatorial from master others leading system and and so two shend cond
d passed as they three poland beyond lined to with city award and stand its black ards a hulm premiere on the us could argued wo
temit the abdul work os studies one four three loved from a livks in party by these largestment both saturated in twins in the t
\end{lstlisting}
\end{minipage}
\par\medskip

\noindent\begin{minipage}{\linewidth}
\textbf{Real Text8}
\begin{lstlisting}
n nine seven five ad also malay sailors knew of mauritius as early as the one zero th century portuguese sailors one five zero s
even one five one three mauritius was discovered and visited by the portuguese between one five zero seven and one five one thre
e an official world map by diego ribeiro described from west to east the first island mascarenhas the second santa apolonia and 
the third domingo friz one the three islands r union mauritius and rodrigues were discovered some years earlier by chance during
\end{lstlisting}
\end{minipage}
\par\medskip
\endgroup

\clearpage
\section{\textcolor{black}{Additional Text8 controls}}
\label{app:text8-two-head}
\label{sec:text8-two-head}

\subsection{Pure-jump control with ordinary self-attention}
\label{app:text8-ddit-pure-jump}

We additionally evaluate a pure-jump predictor with ordinary self-attention,
using the matched-layer DDiT control. Queries, keys, and values are
recomputed from the evolving hidden states at every layer, and attention
includes the diagonal. This differs from both the LOO and unmasked
fixed-source predictors in the main comparison. The remaining layers match
our backbone: eight blocks of width 256, eight heads of dimension 32,
SwiGLU width 704, time conditioning, rotary positional embeddings,
normalization, dropout, and learned 64-dimensional token embeddings. The control
has 6,878,624 trainable parameters, versus 6,878,880 for structural LOO;
the difference is the latter's learned 256-dimensional initial query.

The run uses the same embedded pure-jump path as EPJ: coordinates start
at zero, jump with CDF $t^2$, and remain at their sampled token embeddings.
It is trained from scratch on an A100 for 60,000 updates with seed 0,
length-128 sequences, batch size 512, uniform training times on $[0,.999]$,
and ordinary cross-entropy over all positions. AdamW, learning rate
$10^{-3}$, FP32/math attention, and EMA decay $0.9999$ match the main
protocol. No likelihood completion or LOO loss correction is used.
The mean training loss over the final 1,000 updates is $1.5089$;
final EMA validation CE is $1.4885$ over all positions, $1.8254$ for the
per-sequence average over unabsorbed positions, and $5.20\times10^{-5}$
over absorbed positions. These are denoising losses, not language-model
likelihoods or generation-quality scores.

Table~\ref{tab:text8-ddit-pure-jump} compares this control with the
EPJ structural-LOO checkpoint for the same training seed.
Each evaluation uses the 60,000-update EMA, sampling seed 0, 1,024 sequences, the cosine
grid, and categorical destinations at temperature one or $0.7$. NFE
includes terminal classification of any remaining unabsorbed positions.
The validation-reference, lexicon, and character-map hashes agree across
all twelve evaluations. These evaluations share two trained checkpoints;
there is no training-seed standard deviation.

Ordinary self-attention gives higher $V_5$ and $V_6$ and lower
$\operatorname{JSD}_4$ in all six matched sampling settings, while the
word-diversity comparisons are mixed. At 128 NFE and temperature one,
$V_5$ increases from $44.35\%$ to $54.08\%$ and
$\operatorname{JSD}_4$ decreases from $0.08215$ to $0.07284$; both $D_5$
and $E_5$ also increase. Lowering the self-attention predictor's sampling
temperature to $0.7$ raises $V_5$ to $73.41\%$, but reduces diversity and
increases divergence. At temperature one, increasing the budget from
512 to 2,048 NFE gives little additional validity and reduces $D_5$ and
$E_5$. No evaluation contains duplicate full sequences.

This provides a stronger pure-jump reference than the structural-LOO EPJ
predictor in this seed. At an unabsorbed pure-jump position, the local
zero state carries no endpoint information, so the local denoising shortcut
that motivates LOO for continuous paths is absent. Structural LOO therefore
need not improve this baseline. The result is specific to this single-seed
comparison and does not establish an ordering across seeds or continuous
paths.

\begin{table}[htbp]
\centering
\small
\setlength{\tabcolsep}{3pt}
\caption{Pure-jump predictors with matched layers, 60,000-update EMA checkpoints, training and sampling seed 0, and 1,024 length-128 sequences per setting. LOO is the fixed-source EPJ predictor; SA is ordinary self-attention. All six sampling settings are shown. Counts $D_5$ are exact and $E_5$ is rounded to one decimal.}
\label{tab:text8-ddit-pure-jump}
\begin{tabular}{rclrrrrr}
\toprule
NFE & Temp. & Model & $V_5$ (\%) $\uparrow$ & $V_6$ (\%) $\uparrow$ & $D_5$ $\uparrow$ & $E_5$ $\uparrow$ & $\operatorname{JSD}_4$ $\downarrow$ \\
\midrule
128 & 1.0 & LOO & 44.35 & 34.49 & 2135 & 1101.4 & 0.08215 \\
128 & 1.0 & SA & 54.08 & 46.45 & 2266 & 1126.0 & 0.07284 \\
128 & 0.7 & LOO & 64.67 & 52.91 & 2121 & 707.6 & 0.08250 \\
128 & 0.7 & SA & 73.41 & 65.40 & 2162 & 726.5 & 0.07988 \\
\midrule
512 & 1.0 & LOO & 47.57 & 37.03 & 2145 & 1031.1 & 0.07773 \\
512 & 1.0 & SA & 58.17 & 50.40 & 2367 & 1078.1 & 0.06881 \\
512 & 0.7 & LOO & 67.82 & 56.48 & 2155 & 706.5 & 0.08246 \\
512 & 0.7 & SA & 74.61 & 66.62 & 2140 & 704.0 & 0.08188 \\
\midrule
2048 & 1.0 & LOO & 48.72 & 39.05 & 2252 & 1137.3 & 0.07654 \\
2048 & 1.0 & SA & 58.64 & 50.74 & 2280 & 1071.3 & 0.06884 \\
2048 & 0.7 & LOO & 68.24 & 57.32 & 2183 & 743.9 & 0.08051 \\
2048 & 0.7 & SA & 75.81 & 68.53 & 2151 & 740.2 & 0.08026 \\
\bottomrule
\end{tabular}
\end{table}

\FloatBarrier

\subsection{CADD-derived corruption}
\label{app:cadd-diagnostics}

We adapt the masked categorical supervision and Gaussian interpolation
of CADD \citep[Section~4.3 and Appendix~B]{zheng2026cadd}, with detached
hints and $10\%$ whole-hint dropout from its
\href{https://github.com/apple-aiml-research/ml-cadd/blob/a10369bc60f2c96fc253d4aeb33fa3872c44390f/CADD.md}{released training guide}.
Our architecture and normalized embeddings are
adaptations; the control is not a reproduction of the original CADD
backbone. Using clean-progress time $t$ (the paper's noise time is $1-t$),
draw $t\sim\operatorname{Unif}[0,.999)$ per sequence and independent
$R_{i,t}\sim\operatorname{Bernoulli}(t)$, where $R_{i,t}=1$ means resolved.
For independent $\varepsilon_i\sim\mathcal N(0,I_{64})$, set
\[
 X_{i,t}=R_{i,t}e(Y_i)+(1-R_{i,t})\operatorname{sg}\!\left[
 H\bigl(t e(Y_i)+(1-t)\varepsilon_i\bigr)\right],
 \qquad H\sim\operatorname{Bernoulli}(0.9).
\]
Here $\operatorname{sg}$ stops gradients and one $H$ is shared by the
entire minibatch. Resolved inputs still supply embedding gradients.
Before hint dropout, the unresolved conditional covariance is
$(1-t)^2 I_{64}$, with a Gaussian initial state.

\FloatBarrier

\subsection{CADD-derived training with one shared predictor}
\label{app:cadd-shared}
\label{sec:additional-settings}

\begingroup
\color{black}
With CADD-derived training \citep{zheng2026cadd}, one predictor supplies
continuous and discrete updates. The paired seed-0 comparison changes
only the attention mask. With posterior-mean (soft) endpoints, LOO
improves validity, distinct valid-word counts, and four-gram agreement;
with argmax (hard) endpoints, it improves validity but worsens divergence
(Table~\ref{tab:cadd-shared-primary}, 128 NFE with CAT; other budgets below).
\endgroup

This comparison uses a matched pair trained for 60,000 updates
with seed 0. One transformer stream and one categorical head supply the
predictions for both continuous and discrete updates. Control and LOO
share initialization, data order, parameter count, corruption, optimizer,
and training budget; only the attention diagonal differs. Both use fixed
position-local sources for keys and values at every layer and queries
initialized independently of the inputs. LOO excludes the position's own
input and status from both predictions, including indirect routes within
one network call. Resolved positions receive no loss and remain fixed
during sampling; their embeddings remain visible to other positions.

We use the 90/5/5-million-character Text8 split, $|\cV|=27$, $L=128$,
and normalized embeddings in $\R^{64}$ with norm $8$. The network has eight blocks of
width 256, eight attention heads, SwiGLU width 704, rotary positions,
AdaLN-Zero conditioning, and feature/attention dropout $0.1$.
There are 6,878,880 trainable parameters including the embeddings.
Training uses batch size 512, AdamW with constant learning rate $10^{-3}$,
coefficients $(0.9,0.999)$, numerical constant $10^{-8}$, no warmup or
weight decay, gradient clipping at one, and EMA decay $0.9999$ for network
and embeddings. The corruption is the CADD-derived Gaussian
interpolation, linear masking, detached hints, and $10\%$ minibatch hint
dropout described in Appendix~\ref{app:cadd-diagnostics}. The loss is a
single categorical cross-entropy term at unresolved positions, divided
by $BL$ for a batch of $B$ sequences, using the one shared prediction. Our fixed-source architecture
and normalized embeddings are adaptations of the CADD recipe
\citep{zheng2026cadd}; this control does not reproduce the published model.

\paragraph{Continuous estimates and discrete rules.}
Sampling starts with independent Gaussian hints and no resolved positions.
On a uniform grid from zero to $.999$, one call at time $t$ gives logits
$z_{i,t}$ for both updates. An unresolved position commits over $[t,r]$
with probability $(r-t)/(1-t)$, except that the final interval commits
all survivors. Uncommitted positions receive a fresh hint with independent
$\varepsilon'_i\sim\mathcal N(0,I_{64})$,
\[
 X_{i,r}=r m_{i,t}+(1-r)\varepsilon'_i,\qquad
 m_{i,t}=\begin{cases}
 e(\arg\max_a z_{i,t}(a)),&\text{hard},\\
 \sum_a\softmax(z_{i,t})_a e(a),&\text{soft}.
 \end{cases}
\]
These are fresh-noise updates rather than Brownian-bridge increments.
The endpoint choice is independent of the discrete rule: CAT uses raw
probabilities, TEMP uses temperature $0.7$, TOPP uses mass $0.9$, and MAX
uses argmax only for the committed token. In particular, TEMP and TOPP
never alter the soft continuous mean. CONF preserves each sequence's
independently drawn commitment count and selects that many unresolved
positions with largest raw predicted confidence, with exchangeable random
tie-breaking; destinations remain categorical. Positions already resolved
stay at their committed embeddings. These sampling changes, and the LOO
continuous prediction without Bayesian completion, carry no guarantee of
preserving the prescribed marginals.

One NFE counts one transformer call, including the last commitment call;
there is no additional terminal call. All 60 combinations of
two predictors, two continuous estimates, five discrete rules, and
32/128/512 NFE are complete for training seed 0. Each generates 1,024
length-128 sequences with sampling seed 0, batch size 64, EMA network and
embeddings, FP32 and math attention on a V100. The dictionary, validation
reference, and metric implementation match Appendix~\ref{app:metrics}.
Different decoding cells reuse checkpoints and do not supply independent
training seeds. CAT at 128 NFE is the primary comparison; the sweep is
reported in full below, rather than selecting the best observed setting.

\paragraph{Results.}
LOO improves $V_5$ in all 30 matched comparisons, $V_6$ in 29, $D_5$ in 21,
$E_5$ in eight, and $\operatorname{JSD}_4$ in nine. At 128 NFE with CAT,
soft LOO improves all five metrics over its control. Hard LOO increases
$V_5$ and $D_5$ but reduces $E_5$ from $417.2$ to $379.5$ and worsens
$\operatorname{JSD}_4$ from $0.10837$ to $0.12347$.
For LOO, soft endpoints have larger $E_5$ and lower divergence than hard
endpoints for every tested rule and budget. TEMP and TOPP increase soft
LOO validity over CAT at 128 NFE, but reduce its diversity and worsen its
divergence. Increasing NFE from 128 to 512 raises $V_5$, reduces $E_5$,
and worsens divergence for all ten LOO settings. For soft LOO with CAT,
$V_5$ increases from $67.55\%$ to $71.27\%$, while $E_5$ falls from
$761.8$ to $542.3$ and divergence rises from $0.08610$ to $0.09606$.
These are paired observations for one training seed.

\begin{table}[htbp]
\centering
\small
\setlength{\tabcolsep}{4pt}
\caption{CADD-derived shared-predictor decoding at 128 NFE: 60,000-update EMA, training and sampling seed 0, and 1,024 sequences per cell. Hard and soft specify the continuous endpoint; the discrete rule is varied independently. Bold marks the better value in each matched control/LOO pair. No training-seed standard deviation is available.}
\label{tab:cadd-shared-128}
\label{tab:cadd-shared-primary}
\begin{tabular}{llrrrrr}
\toprule
Rule & Predictor & $V_5$ (\%) $\uparrow$ & $V_6$ (\%) $\uparrow$ & $D_5$ $\uparrow$ & $E_5$ $\uparrow$ & $\operatorname{JSD}_4$ $\downarrow$ \\
\midrule
\multicolumn{7}{l}{Hard continuous endpoint} \\
CAT & Control & 45.95 & 33.59 & 1316 & \textbf{417.2} & \textbf{0.10837} \\
CAT & LOO & \textbf{70.62} & \textbf{58.63} & \textbf{1865} & 379.5 & 0.12347 \\
TEMP & Control & 53.01 & 38.82 & 1346 & \textbf{336.8} & \textbf{0.11475} \\
TEMP & LOO & \textbf{79.55} & \textbf{66.52} & \textbf{1541} & 189.0 & 0.17239 \\
TOPP & Control & 49.58 & 37.71 & 1380 & \textbf{436.4} & \textbf{0.10492} \\
TOPP & LOO & \textbf{75.35} & \textbf{64.57} & \textbf{1802} & 334.6 & 0.13314 \\
MAX & Control & 73.68 & 52.33 & \textbf{927} & \textbf{88.6} & \textbf{0.20184} \\
MAX & LOO & \textbf{97.40} & \textbf{85.93} & 308 & 7.7 & 0.42072 \\
CONF & Control & 42.96 & 27.54 & \textbf{978} & \textbf{188.5} & \textbf{0.14950} \\
CONF & LOO & \textbf{70.64} & \textbf{35.86} & 588 & 20.9 & 0.32908 \\
\midrule
\multicolumn{7}{l}{Soft continuous endpoint} \\
CAT & Control & 43.93 & 33.07 & 1496 & 615.6 & 0.10799 \\
CAT & LOO & \textbf{67.55} & \textbf{58.79} & \textbf{2291} & \textbf{761.8} & \textbf{0.08610} \\
TEMP & Control & 51.01 & 38.98 & 1547 & \textbf{528.9} & \textbf{0.10414} \\
TEMP & LOO & \textbf{77.58} & \textbf{68.17} & \textbf{1986} & 462.6 & 0.10927 \\
TOPP & Control & 47.23 & 36.01 & 1544 & 616.4 & 0.10132 \\
TOPP & LOO & \textbf{72.78} & \textbf{64.38} & \textbf{2244} & \textbf{693.1} & \textbf{0.09075} \\
MAX & Control & 69.01 & 52.52 & \textbf{1219} & \textbf{225.6} & \textbf{0.15387} \\
MAX & LOO & \textbf{95.64} & \textbf{88.81} & 712 & 27.7 & 0.31796 \\
CONF & Control & 41.16 & 29.95 & \textbf{1189} & \textbf{377.7} & \textbf{0.13903} \\
CONF & LOO & \textbf{74.02} & \textbf{56.04} & 1024 & 86.0 & 0.22006 \\
\bottomrule
\end{tabular}
\end{table}

\FloatBarrier

\begin{table}[htbp]
\centering
\small
\setlength{\tabcolsep}{4pt}
\caption{CADD-derived shared-predictor decoding at 32 NFE: 60,000-update EMA, training and sampling seed 0, and 1,024 sequences per cell. Hard and soft specify the continuous endpoint; the discrete rule is varied independently. Bold marks the better value in each matched control/LOO pair. No training-seed standard deviation is available.}
\label{tab:cadd-shared-32}
\begin{tabular}{llrrrrr}
\toprule
Rule & Predictor & $V_5$ (\%) $\uparrow$ & $V_6$ (\%) $\uparrow$ & $D_5$ $\uparrow$ & $E_5$ $\uparrow$ & $\operatorname{JSD}_4$ $\downarrow$ \\
\midrule
\multicolumn{7}{l}{Hard continuous endpoint} \\
CAT & Control & 30.92 & 19.51 & 1089 & 366.2 & 0.11044 \\
CAT & LOO & \textbf{56.29} & \textbf{43.03} & \textbf{1849} & \textbf{467.4} & \textbf{0.10689} \\
TEMP & Control & 37.57 & 23.75 & 1105 & \textbf{294.8} & \textbf{0.11213} \\
TEMP & LOO & \textbf{67.52} & \textbf{51.60} & \textbf{1661} & 252.7 & 0.13686 \\
TOPP & Control & 33.65 & 22.15 & 1156 & 384.3 & \textbf{0.10560} \\
TOPP & LOO & \textbf{61.58} & \textbf{48.24} & \textbf{1858} & \textbf{420.4} & 0.10889 \\
MAX & Control & 55.28 & 29.88 & \textbf{835} & \textbf{93.2} & \textbf{0.17755} \\
MAX & LOO & \textbf{92.16} & \textbf{70.29} & 466 & 12.1 & 0.36684 \\
CONF & Control & 27.48 & 15.38 & 945 & \textbf{245.6} & \textbf{0.13181} \\
CONF & LOO & \textbf{53.93} & \textbf{32.06} & \textbf{1016} & 84.6 & 0.21833 \\
\midrule
\multicolumn{7}{l}{Soft continuous endpoint} \\
CAT & Control & 27.55 & 17.80 & 1166 & 506.9 & 0.12304 \\
CAT & LOO & \textbf{53.05} & \textbf{43.00} & \textbf{2224} & \textbf{943.4} & \textbf{0.08325} \\
TEMP & Control & 34.09 & 22.90 & 1245 & 473.2 & 0.11310 \\
TEMP & LOO & \textbf{65.13} & \textbf{54.62} & \textbf{2245} & \textbf{726.1} & \textbf{0.09050} \\
TOPP & Control & 31.10 & 21.06 & 1270 & 572.1 & 0.11238 \\
TOPP & LOO & \textbf{59.20} & \textbf{49.50} & \textbf{2331} & \textbf{958.3} & \textbf{0.07979} \\
MAX & Control & 46.92 & 29.35 & 1062 & \textbf{233.0} & \textbf{0.14755} \\
MAX & LOO & \textbf{88.46} & \textbf{75.38} & \textbf{1083} & 61.3 & 0.25569 \\
CONF & Control & 23.23 & 14.32 & 996 & \textbf{421.8} & 0.15064 \\
CONF & LOO & \textbf{58.38} & \textbf{45.14} & \textbf{1499} & 275.5 & \textbf{0.14467} \\
\bottomrule
\end{tabular}
\end{table}

\FloatBarrier

\begin{table}[htbp]
\centering
\small
\setlength{\tabcolsep}{4pt}
\caption{CADD-derived shared-predictor decoding at 512 NFE: 60,000-update EMA, training and sampling seed 0, and 1,024 sequences per cell. Hard and soft specify the continuous endpoint; the discrete rule is varied independently. Bold marks the better value in each matched control/LOO pair. No training-seed standard deviation is available.}
\label{tab:cadd-shared-512}
\begin{tabular}{llrrrrr}
\toprule
Rule & Predictor & $V_5$ (\%) $\uparrow$ & $V_6$ (\%) $\uparrow$ & $D_5$ $\uparrow$ & $E_5$ $\uparrow$ & $\operatorname{JSD}_4$ $\downarrow$ \\
\midrule
\multicolumn{7}{l}{Hard continuous endpoint} \\
CAT & Control & 54.24 & 42.33 & 1450 & \textbf{425.9} & \textbf{0.11701} \\
CAT & LOO & \textbf{74.56} & \textbf{62.65} & \textbf{1701} & 286.1 & 0.13670 \\
TEMP & Control & 61.98 & 47.86 & 1343 & \textbf{304.6} & \textbf{0.12622} \\
TEMP & LOO & \textbf{83.25} & \textbf{72.33} & \textbf{1400} & 158.3 & 0.18997 \\
TOPP & Control & 58.74 & 46.96 & 1478 & \textbf{430.7} & \textbf{0.11372} \\
TOPP & LOO & \textbf{80.04} & \textbf{69.98} & \textbf{1675} & 271.1 & 0.15121 \\
MAX & Control & 80.61 & 62.54 & \textbf{898} & \textbf{77.2} & \textbf{0.22200} \\
MAX & LOO & \textbf{98.05} & \textbf{90.40} & 263 & 6.9 & 0.43330 \\
CONF & Control & 55.90 & \textbf{38.45} & \textbf{903} & \textbf{140.4} & \textbf{0.18721} \\
CONF & LOO & \textbf{72.81} & 34.28 & 439 & 15.8 & 0.37798 \\
\midrule
\multicolumn{7}{l}{Soft continuous endpoint} \\
CAT & Control & 53.55 & 43.24 & 1513 & 539.4 & 0.11474 \\
CAT & LOO & \textbf{71.27} & \textbf{61.85} & \textbf{2049} & \textbf{542.3} & \textbf{0.09606} \\
TEMP & Control & 60.19 & 48.52 & 1536 & \textbf{460.0} & \textbf{0.11230} \\
TEMP & LOO & \textbf{80.62} & \textbf{71.82} & \textbf{1792} & 340.2 & 0.12702 \\
TOPP & Control & 57.41 & 47.12 & 1569 & \textbf{554.7} & 0.10915 \\
TOPP & LOO & \textbf{76.90} & \textbf{68.48} & \textbf{2019} & 507.2 & \textbf{0.10370} \\
MAX & Control & 77.99 & 64.13 & \textbf{1099} & \textbf{185.5} & \textbf{0.17410} \\
MAX & LOO & \textbf{96.86} & \textbf{91.58} & 533 & 18.2 & 0.34971 \\
CONF & Control & 55.82 & 42.74 & \textbf{1043} & \textbf{244.8} & \textbf{0.16683} \\
CONF & LOO & \textbf{76.37} & \textbf{53.38} & 749 & 39.8 & 0.27811 \\
\bottomrule
\end{tabular}
\end{table}

\FloatBarrier

\paragraph{Word concentration.}
MAX and CONF concentrate the word distribution. In every matched setting,
CONF has lower $E_5$ and higher divergence than CAT. Table~\ref{tab:cadd-concentration}
reports the share of valid word occurrences accounted for by the three
and ten most frequent words. The three most frequent are \texttt{three},
\texttt{seven}, and \texttt{eight}, in varying order. With hard LOO and
MAX, they account for $86.92\%$ of valid word occurrences of length at
least five, despite $V_5=97.40\%$. These are lexical concentration
measurements, not fractions of identical generated sequences.

\begin{table}[htbp]
\centering
\small
\setlength{\tabcolsep}{6pt}
\caption{Word concentration for shared-predictor LOO at 128 NFE. Percentages use dictionary-valid occurrences of words of at least five letters as the denominator. The checkpoints and samples are those in Table~\ref{tab:cadd-shared-128}.}
\label{tab:cadd-concentration}
\begin{tabular}{llrrr}
\toprule
Endpoint & Rule & Top-three share (\%) & Top-ten share (\%) & $E_5$ \\
\midrule
Hard & CAT & 28.24 & 32.57 & 379.5 \\
Hard & TEMP & 39.16 & 43.31 & 189.0 \\
Hard & TOPP & 29.65 & 34.41 & 334.6 \\
Hard & MAX & 86.92 & 89.23 & 7.7 \\
Hard & CONF & 72.61 & 75.27 & 20.9 \\
\midrule
Soft & CAT & 16.58 & 21.37 & 761.8 \\
Soft & TEMP & 23.84 & 29.06 & 462.6 \\
Soft & TOPP & 17.83 & 22.41 & 693.1 \\
Soft & MAX & 67.55 & 71.79 & 27.7 \\
Soft & CONF & 49.49 & 54.29 & 86.0 \\
\bottomrule
\end{tabular}
\end{table}

\FloatBarrier

\clearpage
\begingroup
\color{black}
\section{LM1B experimental details and additional results}
\label{app:lm1b}

\subsection{Data, architecture, and training}
\label{app:lm1b-setup}

Our LM1B experiments \citep{chelba2014billion} use the BERT uncased tokenizer
\citep{devlin2019bert}, a 30,522-token WordPiece vocabulary
\citep{wu2016wordpiece}, and length-128 windows, matching the tokenizer
and window size of \citet{chemseddine2026spherical}.
We compare RJD-LOO, unrestricted RJD, embedded pure jumps (EPJ),
and pure-diffusion and completion variants. RJD and RJD-LOO use
fixed-source attention and differ in access to the current position's
noisy input. Raw RJD-LOO predictions drive both drift and resets; EPJ uses
unrestricted prediction and absorbing jumps without continuous motion.
RJD, RJD-LOO, and EPJ all use $\kappa(t)=t^2$ and start from zero;
the two RJD variants use diffusion scale $\sigma_d=1$.

We use a DDiT-style backbone \citep{lou2024sedd}, adapted to continuous
token embeddings and fixed-source attention.
RJD-LOO, RJD, and EPJ have 12 layers, width and embedding dimension 768,
12 attention heads, and learned embeddings of norm 14.18.
All LM1B predictors have approximately 116M trainable parameters,
including the learned token embeddings.
As in \citet{chemseddine2026spherical}, the output head shares the
learned token embeddings with the conditional path. Given the network's
output vector $f_i$, token probabilities are obtained by applying
softmax over logits $e(a)^\top f_i+b_a$, where $b_a$ is a learned
token bias.
All LM1B models use batch size 512 and AdamW with peak learning
rate $3\times10^{-4}$, 2,500 linear warmup updates, and cosine
decay to $3\times10^{-5}$ at 1M updates. EMA decay is 0.9999.
We use one training seed per model (seed 0) and evaluate
750,000-update EMA weights.

\paragraph{Self-attention control (RJD-SA).}
\label{app:lm1b-self-attention}
RJD-SA computes queries, keys, and values from evolving hidden states,
initialized from the corrupted input, with all positions visible.
It matches the fixed-source models' depth, width, heads, time conditioning,
and tied embedding/readout.
It uses the same RJD path, all-position cross-entropy loss, and raw
unrestricted predictions for drift and resets. Training starts from
random initialization and follows the shared protocol above.
Training uses NVIDIA A100 and H100 GPUs and AMD MI300A accelerators;
evaluation uses NVIDIA A100 GPUs with the common protocol.
The single training run and mixed training hardware prevent isolating
architecture from training variability.

\subsection{Evaluation metrics}
\label{app:lm1b-metrics}
\emph{Generative perplexity (PPL)} exponentiates the mean negative
log-probability assigned to GPT-2-tokenized samples by external
GPT-2-large \citep{radford2019language}. Lower means more predictable
to this scorer.
Reported PPL is the arithmetic mean of per-generation-seed perplexities.
\emph{Token entropy} is $H=-\sum_a f(a)\log f(a)$
\citep{shannon1948communication}, where $f(a)$ is the relative frequency
of BERT token $a$ within a 128-token window, including special tokens.
We average over windows in nats. Higher values indicate less concentrated
token frequencies.
\emph{MAUVE} \citep{pillutla2021mauve} compares generated and reference
distributions in a quantized embedding space, capturing both generated
text unlike the reference and missing reference content. Scores range
from zero to one; higher is closer to the reference. We use GPT-2-large
features with a maximum input length of 256 tokens, 512 clusters, and
5,120 fixed held-out windows (reference seed 20261029). Each corpus score
averages results from clustering seeds 25, 26, and 27.
Reported SDs describe variation across generation seeds, not training runs.
Low PPL can accompany repetition
\citep{holtzman2020degeneration}.

\subsection{Uniform-time top-$p$ sweeps at 750k updates}
\label{app:lm1b-uniform}
\label{app:lm1b-earlier}

Figure~\ref{fig:lm1b-topp} uses 750,000-update EMA weights, one
training seed (0), and 128 predictor calls on a uniform
physical-time grid. RJD and PD end at
$t=0.999$ and use nearest-embedding readout; EPJ ends at $t=1$ with
categorical output. We sweep $p\in\{1,0.9,0.8,0.7\}$.
Completion changes the drift. Top-$p$ filters jump destinations for
RJD/EPJ, leaving RJD drift probabilities unfiltered; for PD it filters
drift probabilities after any completion.

All eight methods cover every $p$, with generation seeds 42, 43, and 44
and 5,120 length-128 windows per seed, giving 32 settings and 96 corpora.

\paragraph{Candidate-restricted Bayesian completion.}
\label{app:lm1b-candidate-completion}
RJD-BC-D and PD-BC approximate \eqref{eq:completion} to reduce the
cost of likelihood evaluation over the 30,522-token vocabulary.
Raw and completed variants share the same trained checkpoint.
Let $\operatorname{Top}_K(f)$ denote the tokens with the $K$ largest
values of $f$. For $0<t<1$, define
\begin{equation}
 \mathcal C^K_{i,t}(x)
 =\operatorname{Top}_K\!\bigl(q_{\theta,i,t}(\cdot\mid x_{-i})\bigr)
  \cup\operatorname{Top}_K\!\bigl(\ell_{i,t}(x_i\mid\cdot)\bigr).
 \label{eq:lm1b-bc-candidates}
\end{equation}
We use $K=64$ for each ranking and count duplicate tokens once, so
$|\mathcal C^K_{i,t}(x)|\le 2K=128$. The completed
prediction before nucleus filtering is
\begin{equation}
 \widehat p^{\,K}_{\theta,i,t}(a\mid x)
 =\frac{\mathbf 1_{\{a\in\mathcal C^K_{i,t}(x)\}}
        q_{\theta,i,t}(a\mid x_{-i})\ell_{i,t}(x_i\mid a)}
       {\sum_{b\in\mathcal C^K_{i,t}(x)}
        q_{\theta,i,t}(b\mid x_{-i})\ell_{i,t}(x_i\mid b)}.
 \label{eq:lm1b-bc-restricted}
\end{equation}
Here $\ell_{i,t}$ is the likelihood of the corresponding training path:
RJD uses \eqref{eq:local-law}, and PD uses the Gaussian bridge
likelihood. With equal-norm embeddings, the local ranking can be
computed from $x_i^\top e(a)$. The drift uses the embedding mean of
\eqref{eq:lm1b-bc-restricted}, with subsequent top-$p$ filtering for PD.
At $t=0$, the local likelihood is token-independent and we use the
full-vocabulary LOO distribution before any nucleus filtering.

\paragraph{Results across the sweep.}
Raw and completed LOO improve PPL and MAUVE over unrestricted
fixed-source prediction throughout the sweep, while completion lowers
MAUVE relative to raw LOO (Table~\ref{tab:lm1b-uniform-mauve}).
Top-$p$ filtering improves RJD-LOO; stronger filtering can further lower
PD-LOO's PPL while reducing its MAUVE. RJD-SA improves on RJD, but its
MAUVE changes across the sweep are small relative to seed variation.
The settings in Table~\ref{tab:lm1b-best-mauve} are selected from this
sweep.

\begin{table}[htbp]
\color{black}
\centering
\small
\setlength{\tabcolsep}{4pt}
\caption{MAUVE across the uniform-time top-$p$ sweep at 750k
updates and 128 NFE. Entries are mean $\pm$ sample
SD across three generation seeds, with 5,120 windows per seed.
Higher is better and bold marks the highest mean in each column.}
\label{tab:lm1b-uniform-mauve}
\begin{tabular}{lrrrr}
\toprule
Method & $p=1$ & $p=0.9$ & $p=0.8$ & $p=0.7$ \\
\midrule
RJD & $0.1878\pm0.0056$ & $0.1904\pm0.0112$ & $0.1888\pm0.0031$ & $0.2019\pm0.0032$ \\
RJD-SA & $0.4556\pm0.0089$ & $0.4663\pm0.0157$ & $0.4634\pm0.0114$ & $0.4704\pm0.0192$ \\
RJD-LOO & $0.8107\pm0.0103$ & $0.8810\pm0.0015$ & $\mathbf{0.9039}\pm0.0058$ & $\mathbf{0.9021}\pm0.0038$ \\
RJD-BC-D & $0.6450\pm0.0123$ & $0.7543\pm0.0131$ & $0.7624\pm0.0126$ & $0.7399\pm0.0081$ \\
EPJ & $\mathbf{0.8521}\pm0.0083$ & $\mathbf{0.8965}\pm0.0103$ & $0.8632\pm0.0065$ & $0.7889\pm0.0124$ \\
\midrule
PD-LOO & $0.4358\pm0.0061$ & $0.5804\pm0.0056$ & $0.5970\pm0.0089$ & $0.5459\pm0.0183$ \\
PD & $0.1584\pm0.0089$ & $0.1842\pm0.0042$ & $0.1946\pm0.0075$ & $0.2111\pm0.0138$ \\
PD-BC & $0.3576\pm0.0071$ & $0.3370\pm0.0049$ & $0.2981\pm0.0058$ & $0.2658\pm0.0036$ \\
\bottomrule
\end{tabular}
\end{table}

\FloatBarrier

\subsection{Uniform-time generation at matched NFE budgets}
\label{app:lm1b-generation}

We compare 32, 64, and 128 predictor calls using the same 750k EMA
checkpoints and uniform-time sampling and scoring
protocol of Appendix~\ref{app:lm1b-uniform}.
For each recipe, top-$p$ is selected by the highest observed mean MAUVE
at 128 NFE and held fixed across budgets: $p=0.8$ for RJD-LOO,
RJD-BC-D, and PD-LOO, and $p=0.9$ for EPJ.
RJD-BC-D uses the candidate-restricted completion defined in
Appendix~\ref{app:lm1b-candidate-completion}.
Table~\ref{tab:lm1b-generation} reports the resulting budget comparison.

\begin{table}[htbp]
\color{black}
\centering
\small
\setlength{\tabcolsep}{4pt}
\caption{LM1B generation on uniform physical-time grids at 750k updates.
Entries are mean $\pm$ sample SD over three
\emph{generation} seeds (42, 43, 44), with 5,120 length-128 windows per
seed and one training seed per model. Bold marks the lowest mean PPL and highest mean MAUVE at each NFE.}
\label{tab:lm1b-generation}
\begin{tabular}{lrrrrr}
\toprule
Method & $p$ & NFE & PPL $\downarrow$ & $H$ (nats) & MAUVE $\uparrow$ \\
\midrule
RJD-LOO & 0.8 & 32 & $83.09\pm0.55$ & $4.2999\pm0.0008$ & $0.8483\pm0.0172$ \\
RJD-LOO & 0.8 & 64 & $69.91\pm0.31$ & $4.2996\pm0.0007$ & $\mathbf{0.8855}\pm0.0090$ \\
RJD-LOO & 0.8 & 128 & $63.25\pm0.29$ & $4.2987\pm0.0002$ & $\mathbf{0.9039}\pm0.0058$ \\
\midrule
RJD-BC-D & 0.8 & 32 & $79.78\pm0.49$ & $4.2240\pm0.0014$ & $0.6618\pm0.0234$ \\
RJD-BC-D & 0.8 & 64 & $68.38\pm0.07$ & $4.2158\pm0.0011$ & $0.7162\pm0.0141$ \\
RJD-BC-D & 0.8 & 128 & $62.25\pm0.10$ & $4.2115\pm0.0007$ & $0.7624\pm0.0126$ \\
\midrule
EPJ & 0.9 & 32 & $91.70\pm0.69$ & $4.2967\pm0.0009$ & $\mathbf{0.8550}\pm0.0073$ \\
EPJ & 0.9 & 64 & $84.66\pm0.43$ & $4.2962\pm0.0019$ & $0.8853\pm0.0042$ \\
EPJ & 0.9 & 128 & $81.08\pm0.12$ & $4.2977\pm0.0014$ & $0.8965\pm0.0103$ \\
\midrule
PD-LOO & 0.8 & 32 & $\mathbf{33.90}\pm0.06$ & $4.1457\pm0.0009$ & $0.5290\pm0.0152$ \\
PD-LOO & 0.8 & 64 & $\mathbf{31.60}\pm0.05$ & $4.1628\pm0.0004$ & $0.5778\pm0.0098$ \\
PD-LOO & 0.8 & 128 & $\mathbf{30.63}\pm0.10$ & $4.1686\pm0.0015$ & $0.5970\pm0.0089$ \\
\bottomrule
\end{tabular}
\end{table}

More calls improve PPL and MAUVE for every recipe. RJD-LOO and EPJ
have similar MAUVE at lower budgets, without a clear ranking across
generation seeds. Completion lowers both PPL and MAUVE relative to raw
RJD-LOO; PD-LOO has the lowest PPL but substantially lower MAUVE.
\FloatBarrier

\subsection{Recovery after inserting a wrong local state}
\label{app:lm1b-recovery}

To probe a potential shortcut through a token's own noisy input, we test
whether models can recover from misleading local evidence using the
surrounding context.

RJD-LOO, RJD, and RJD-SA use 750k EMA checkpoints, one training seed
each, and the same 1,024 held-out text pairs, focal sites, donor tokens,
and two noise draws per pair.
Let $y_i$ be the original token at position $i$ and $d\ne y_i$ a donor
token. After drawing a noisy text at time $t$, replace its local state by
\[
  x_i' = x_i+c\bigl(e(d)-e(y_i)\bigr),\qquad
  c=\begin{cases}t^2,&\text{before any reset},\\
                 1,&\text{after a reset}.
    \end{cases}
\]
This preserves the Gaussian residual and reset times while substituting
the donor token. All 128 positions then evolve freely to $0.999$ with
128 further calls on a uniform grid over the remaining interval.
Top-$p$ filters reset destinations with
$p=0.8$ for RJD-LOO and $p=0.7$ for RJD and RJD-SA; drift probabilities
remain raw. Table~\ref{tab:lm1b-recovery} records the final focal token.

\begin{table}[htbp]
\color{black}
\centering
\small
\setlength{\tabcolsep}{4pt}
\caption{Recovery after donor replacement at 750k updates, using
128 further calls and the model-specific top-$p$ settings stated above.
Percentages use all 2,048 continuations per row. }
\label{tab:lm1b-recovery}
\begin{tabular}{lrrrrrr}
\toprule
& \multicolumn{3}{c}{Final token is original $y_i$}
& \multicolumn{3}{c}{Final token is donor $d$} \\
\cmidrule(lr){2-4}\cmidrule(lr){5-7}
Start $t$ & RJD-LOO & RJD & RJD-SA & RJD-LOO & RJD & RJD-SA \\
\midrule
0.25 & \textbf{18.99} & 5.66 & 7.86 & \textbf{4.83} & 34.47 & 30.03 \\
0.50 & \textbf{30.81} & 0.15 & 0.00 & \textbf{6.01} & 99.71 & 99.61 \\
0.80 & \textbf{31.35} & 0.00 & 0.00 & \textbf{6.25} & 100.00 & 100.00 \\
0.95 & \textbf{33.30} & 0.00 & 0.00 & \textbf{7.03} & 100.00 & 100.00 \\
\bottomrule
\end{tabular}
\end{table}

LOO more often recovers the original token after replacement;
unrestricted models retain the donor at later times.
\FloatBarrier

\subsection{LM1B subject--verb interventions}
\label{app:lm1b-context}

To test whether the models capture longer-range grammatical dependencies,
we examine whether verb predictions track the subject across intervening
words and distracting nouns.

We change only the subject's number and measure immediate verb-form
preferences of the 750k EMA RJD-LOO and RJD checkpoints.
\paragraph{Paired subject intervention.}
In ``The \emph{key} near the cabinets \emph{is} still here'', changing
only ``key'' to ``keys'' should change the preferred verb to ``are'',
despite the intervening noun. We use 800 templates embedded in length-128
windows with held-out background text: 32 controls without an intervening
noun and 768 templates with one or two intervening nouns. Each template
has two noise draws at $t\in\{0.25,0.40,0.55,0.70\}$.
The cohort uses eight subject-noun pairs in 16 subject--noun lexical
blocks and eight control blocks. Blocks requiring multi-WordPiece nouns
are excluded beforehand. Templates and corruption draws are shared
across models and subject variants.

The local verb state follows the conditional corruption law for the
form agreeing with the original subject. Changing only the subject's
number preserves its Gaussian residual and reset times and fixes every
other state, so local verb evidence conflicts with the new subject.
We compare ``is'' with ``are'', or ``was'' with ``were''. Success requires
strictly preferring the agreeing form in \emph{both} variants; ties,
unchanged preferences, and changes in the wrong direction fail.

\paragraph{Controls.}
Controls use an initially wrong local verb with the same noise and reset
history, or an artificial candidate-independent zero-mean Gaussian state.
Each stays fixed across the subject change; ``correct'' and ``wrong''
refer to the original subject and reverse after it changes.
Table~\ref{tab:lm1b-agreement} distinguishes original-subject accuracy
from paired success. LOO's invariance to local-state controls is architectural while
following the subject is learned.
\begin{table}[htbp]
\color{black}
\centering
\footnotesize
\setlength{\tabcolsep}{3pt}
\caption{Immediate agreement preferences at 750k updates, as percentages
of 1,600 cases per time. The first three rows concern the original
subject and give ties half credit. The last requires correct preferences
in both subject variants with the initially correct local verb held
fixed; ties fail. LOO denotes RJD-LOO. Bold marks the higher accuracy
within each condition and time.}
\label{tab:lm1b-agreement}
\begin{tabular}{lrrrrrrrr}
\toprule
& \multicolumn{2}{c}{$t=0.25$} & \multicolumn{2}{c}{$t=0.40$}
& \multicolumn{2}{c}{$t=0.55$} & \multicolumn{2}{c}{$t=0.70$} \\
\cmidrule(lr){2-3}\cmidrule(lr){4-5}\cmidrule(lr){6-7}\cmidrule(lr){8-9}
Condition & LOO & RJD & LOO & RJD & LOO & RJD & LOO & RJD \\
\midrule
Correct local verb & 62.00 & \textbf{97.00} & 74.94 & \textbf{100.00} & 81.56 & \textbf{100.00} & 84.06 & \textbf{100.00} \\
Wrong local verb & \textbf{62.00} & 3.25 & \textbf{74.94} & 0.00 & \textbf{81.56} & 0.00 & \textbf{84.06} & 0.00 \\
Neither candidate locally & \textbf{62.00} & 50.31 & \textbf{74.94} & 50.44 & \textbf{81.56} & 51.31 & \textbf{84.06} & 49.81 \\
Both subject variants correct & \textbf{22.00} & 0.13 & \textbf{48.69} & 0.00 & \textbf{62.44} & 0.00 & \textbf{68.19} & 0.00 \\
\bottomrule
\end{tabular}
\end{table}

\paragraph{Distance and intervening nouns.}
Figure~\ref{fig:lm1b-agreement-distance} stratifies paired success by
WordPiece position difference at early and later times, with initially
correct local verbs. Each distance has 128 templates and two draws,
except the 32 distance-one controls. Intervals use 1,000 bootstrap
resamples of lexical clusters, keeping noise draws and subject variants
together: 16 clusters per nontrivial distance and eight for controls.
They condition on one checkpoint per model and lack multiplicity correction.

LOO's success generally decreases with distance, while unrestricted
RJD rarely follows both subject variants.
Table~\ref{tab:lm1b-agreement-nouns} separates intervening-noun count
and number; matching versus opposite number has no uniform ordering.
Wording and noun configurations also vary across distances.

\begin{table}[htbp]
\color{black}
\centering
\footnotesize
\setlength{\tabcolsep}{3pt}
\caption{RJD-LOO paired subject-tracking success (\%) by intervening-noun
configuration at all four times. M/O means the intervening noun(s) match
or oppose the \emph{original} subject's number; that relation reverses
when the subject changes. Each M/O cell uses 64 templates with two draws.
\textbf{For unrestricted RJD, all corresponding rates are zero except 0.78\%
for one opposite-number noun at distance six and $t=0.25$.}}
\label{tab:lm1b-agreement-nouns}
\begin{tabular}{rrrrrrrrrr}
\toprule
& & \multicolumn{2}{c}{$t=0.25$} & \multicolumn{2}{c}{$t=0.40$}
& \multicolumn{2}{c}{$t=0.55$} & \multicolumn{2}{c}{$t=0.70$} \\
\cmidrule(lr){3-4}\cmidrule(lr){5-6}\cmidrule(lr){7-8}\cmidrule(lr){9-10}
\shortstack{Intervening\\nouns} & Distance & M & O & M & O & M & O & M & O \\
\midrule
1 & 4 & 50.00 & 49.22 & 82.03 & 87.50 & 90.63 & 93.75 & 98.44 & 95.31 \\
1 & 6 & 29.69 & 32.81 & 75.78 & 71.09 & 85.16 & 91.41 & 96.09 & 92.97 \\
1 & 9 & 6.25 & 14.84 & 41.41 & 36.72 & 57.03 & 61.72 & 70.31 & 65.63 \\
2 & 7 & 18.75 & 21.09 & 46.09 & 46.88 & 67.97 & 60.94 & 72.66 & 72.66 \\
2 & 11 & 10.94 & 4.69 & 28.13 & 31.25 & 42.19 & 51.56 & 48.44 & 50.78 \\
2 & 17 & 1.56 & 3.13 & 8.59 & 9.38 & 17.19 & 10.94 & 16.41 & 22.66 \\
\bottomrule
\end{tabular}
\end{table}

\FloatBarrier
\paragraph{Worked example for Table~\ref{tab:lm1b-agreement-nouns}.}
In ``The \emph{key} near the \emph{cabinets} \emph{is} still here'',
``cabinets'' is the single intervening noun. The sequence
``key / near / the / cabinets / is'' spans four position steps, so
this is the first row. Singular ``key'' and plural ``cabinets'' place
it in O. Replacing ``cabinets'' with ``cabinet'' gives an M example.

At $t=0.55$, the first noisy input contains local evidence for ``is''.
The second changes only ``key'' to ``keys'', fixing all other states.
Success requires preferring ``is'' first and ``are'' second. Each cell
aggregates 64 templates with two noise draws. M/O refers to the original subject and reverses
after its number changes.
\FloatBarrier

\subsection{Blinded LLM assessment and context diagnostic}
\label{app:lm1b-blind-assessment}

We assess whether generated LM1B sentences are understandable and
internally consistent, and whether earlier context helps predict later tokens.

\paragraph{Samples and blinding.}
We sampled 300 128-token texts per method (RJD-LOO, RJD, RJD-SA,
and EPJ): 100 uniformly without replacement from each generation seed
42, 43, and 44. We also include 100 held-out real texts with the same decoding.
Generated samples use the 750k, 128-NFE, $p=1$ setting of
Appendix~\ref{app:lm1b-uniform}, with one training seed per model.
LLM judges A (\texttt{gpt-6.1-sol}) and B (\texttt{gpt-6.1-terra}) used
high reasoning effort in separate fresh Codex sessions. They received
anonymous texts in different orders with the same rubric and returned
scores, reasons, and verbatim evidence. Ratings were sealed before
joining source labels.

\paragraph{Rubric and primary outcome.}
Judges scored clarity, internal consistency, grammar, and disruptive
repetition from zero to four; higher is better except for repetition.
Illustratively, ``The chef bake the bread'' conveys a clear meaning
despite a grammatical error; ``The box is empty and contains three
apples'' is grammatical but internally inconsistent.
Acceptance requires clarity and consistency of at least three: mostly
understandable content with mostly consistent relationships.
Texts with no interpretable proposition are excluded from consistency
means but remain in the acceptance denominator. Cosmetic formatting and clipped outer clauses
are ignored; texts are rated as written.

We report each judge separately. For each of 2,000 bootstrap replicates,
we resample texts with replacement within generation seeds and recompute
the mean score or acceptance rate. The 2.5th and 97.5th percentiles give
the 95\% confidence interval. Intervals condition on the
fixed judges and one training run per model; there is no human validation.

\begin{table}[htbp]
\color{black}
\centering
\small
\setlength{\tabcolsep}{3pt}
\caption{Blinded LLM assessment of 128-token LM1B texts at
750k updates and $p=1$: means and 95\% bootstrap intervals.
Consistency uses assessable texts; $U$ is the unassessable percentage.
Accept.\ is the percentage passing the joint clarity/consistency criterion,
using all 300 texts per method or 100 real texts. Bold marks the highest
clarity, grammar, and acceptance means among generators within each judge.
Zero-pass bootstrap bounds collapse to zero without establishing a zero
population rate.}
\label{tab:lm1b-blind-assessment}
\begin{tabular}{llrrrrrr}
\toprule
Corpus & Judge & Clarity & Consist. & Grammar & Repeat. & $U$ (\%) & Accept.\ (\%) \\
\midrule
RJD-LOO & A & \shortstack{$\mathbf{2.44}$\\{\scriptsize$[2.37,2.51]$}} & \shortstack{$2.50$\\{\scriptsize$[2.43,2.58]$}} & \shortstack{$\mathbf{2.14}$\\{\scriptsize$[2.07,2.21]$}} & \shortstack{$0.56$\\{\scriptsize$[0.49,0.64]$}} & 0.3 & \shortstack{$\mathbf{42.7}$\\{\scriptsize$[37.0,48.3]$}} \\
 & B & \shortstack{$\mathbf{2.53}$\\{\scriptsize$[2.47,2.60]$}} & \shortstack{$2.91$\\{\scriptsize$[2.87,2.95]$}} & \shortstack{$\mathbf{2.52}$\\{\scriptsize$[2.45,2.59]$}} & \shortstack{$0.45$\\{\scriptsize$[0.39,0.51]$}} & 0.3 & \shortstack{$\mathbf{52.0}$\\{\scriptsize$[46.3,57.7]$}} \\
\addlinespace[2pt]
RJD & A & \shortstack{$0.69$\\{\scriptsize$[0.64,0.75]$}} & \shortstack{$1.38$\\{\scriptsize$[1.30,1.47]$}} & \shortstack{$0.63$\\{\scriptsize$[0.58,0.69]$}} & \shortstack{$0.26$\\{\scriptsize$[0.21,0.31]$}} & 31.7 & \shortstack{$0.0$\\{\scriptsize$[0.0,0.0]$}} \\
 & B & \shortstack{$0.44$\\{\scriptsize$[0.38,0.50]$}} & \shortstack{$3.07$\\{\scriptsize$[3.00,3.15]$}} & \shortstack{$0.95$\\{\scriptsize$[0.92,0.98]$}} & \shortstack{$0.20$\\{\scriptsize$[0.15,0.24]$}} & 57.7 & \shortstack{$0.0$\\{\scriptsize$[0.0,0.0]$}} \\
\addlinespace[2pt]
RJD-SA & A & \shortstack{$1.12$\\{\scriptsize$[1.06,1.18]$}} & \shortstack{$1.53$\\{\scriptsize$[1.45,1.60]$}} & \shortstack{$0.99$\\{\scriptsize$[0.93,1.05]$}} & \shortstack{$0.33$\\{\scriptsize$[0.27,0.38]$}} & 10.0 & \shortstack{$0.7$\\{\scriptsize$[0.0,1.7]$}} \\
 & B & \shortstack{$0.92$\\{\scriptsize$[0.85,1.00]$}} & \shortstack{$2.89$\\{\scriptsize$[2.83,2.94]$}} & \shortstack{$1.16$\\{\scriptsize$[1.11,1.20]$}} & \shortstack{$0.20$\\{\scriptsize$[0.15,0.24]$}} & 24.3 & \shortstack{$0.3$\\{\scriptsize$[0.0,1.0]$}} \\
\addlinespace[2pt]
EPJ & A & \shortstack{$1.85$\\{\scriptsize$[1.78,1.93]$}} & \shortstack{$1.96$\\{\scriptsize$[1.89,2.04]$}} & \shortstack{$1.53$\\{\scriptsize$[1.47,1.60]$}} & \shortstack{$0.45$\\{\scriptsize$[0.39,0.51]$}} & 0.3 & \shortstack{$12.7$\\{\scriptsize$[9.0,16.3]$}} \\
 & B & \shortstack{$1.76$\\{\scriptsize$[1.69,1.83]$}} & \shortstack{$2.82$\\{\scriptsize$[2.77,2.86]$}} & \shortstack{$1.76$\\{\scriptsize$[1.69,1.83]$}} & \shortstack{$0.31$\\{\scriptsize$[0.26,0.36]$}} & 0.7 & \shortstack{$9.7$\\{\scriptsize$[6.7,13.0]$}} \\
\addlinespace[2pt]
Real LM1B & A & \shortstack{$3.98$\\{\scriptsize$[3.94,4.00]$}} & \shortstack{$3.98$\\{\scriptsize$[3.95,4.00]$}} & \shortstack{$3.79$\\{\scriptsize$[3.69,3.88]$}} & \shortstack{$0.06$\\{\scriptsize$[0.02,0.11]$}} & 0.0 & \shortstack{$99.0$\\{\scriptsize$[97.0,100.0]$}} \\
 & B & \shortstack{$3.97$\\{\scriptsize$[3.91,4.00]$}} & \shortstack{$3.99$\\{\scriptsize$[3.97,4.00]$}} & \shortstack{$3.94$\\{\scriptsize$[3.84,4.00]$}} & \shortstack{$0.09$\\{\scriptsize$[0.04,0.15]$}} & 1.0 & \shortstack{$99.0$\\{\scriptsize$[97.0,100.0]$}} \\
\bottomrule
\end{tabular}
\end{table}

\paragraph{Generation-quality results.}
Both judges favor RJD-LOO on acceptance, clarity, and grammar, while
all generators remain below real text (Table~\ref{tab:lm1b-blind-assessment}).
RJD's high consistency under judge B concerns its assessable texts,
despite poor clarity overall.

\FloatBarrier
\paragraph{Controls and reliability.}
Real-text controls shuffle words (98), splice equal-length word blocks from other
real texts (96), or repeat blocks (98), preserving sentence markers
and outer fragments. Eight ineffective controls were omitted.
Each judge also rated 159 hidden repeats, excluded from method counts;
control intervals resample parent--control pairs.
Both judges detected perturbations and largely agreed on acceptance,
but differed in individual scores (Table~\ref{tab:lm1b-judge-reliability}).
Judge B often repeated entire responses to matching items, so
within-session recognition or reuse remains possible.

\begin{table}[htbp]
\color{black}
\centering
\small
\setlength{\tabcolsep}{5pt}
\caption{Acceptance contrast, control effects, and judge agreement.
Intervals are 95\% bootstrap intervals; control changes are relative to
paired real text. Repeats use 159 pairs per judge; between-judge agreement
uses 1,592 base items, or 1,293 jointly assessable items for consistency.
All-score agreement includes assessability; full-response matches also
require identical evidence and explanation.}
\label{tab:lm1b-judge-reliability}
\begin{tabular}{lrr}
\toprule
Comparison or reliability check & Judge A & Judge B \\
\midrule
RJD-LOO minus EPJ: acceptance (pp) & \shortstack{$30.0$\\{\scriptsize$[23.3,36.7]$}} & \shortstack{$42.3$\\{\scriptsize$[35.7,49.0]$}} \\
\addlinespace[2pt]
Word shuffle: acceptance change (pp) & \shortstack{$-94.9$\\{\scriptsize$[-99.0,-89.8]$}} & \shortstack{$-95.9$\\{\scriptsize$[-99.0,-91.8]$}} \\
Block splice: acceptance change (pp) & \shortstack{$-93.8$\\{\scriptsize$[-97.9,-88.5]$}} & \shortstack{$-95.8$\\{\scriptsize$[-99.0,-91.7]$}} \\
Repetition: change in repetition score & \shortstack{$3.86$\\{\scriptsize$[3.79,3.92]$}} & \shortstack{$3.79$\\{\scriptsize$[3.67,3.88]$}} \\
Hidden repeats: all-score agreement (\%) & 82.4 & 86.2 \\
Hidden repeats: full-response matches (count) & 0/159 & 106/159 \\
Between-judge all-score agreement (\%) & \multicolumn{2}{c}{16.8} \\
Between-judge primary-outcome agreement (\%) & \multicolumn{2}{c}{93.7} \\
Between-judge consistency agreement (\%) & \multicolumn{2}{c}{40.7} \\
\bottomrule
\end{tabular}
\end{table}

\FloatBarrier
\paragraph{External-context diagnostic.}
We used GPT-2-large as an external reader to ask: \emph{Does the earlier
part of a generated sentence help predict a later token, beyond what
the most recent tokens already tell us?} We scored each selected token
under three versions of its preceding context.

Consider the illustrative sentence ``The chef mixed the dough and put
it in the \textbf{oven}.'' Treating each word as one token for this
example, we score ``oven'' with $k=4$ recent tokens, ``put it in the'':
\begin{center}
\small
\begin{tabular}{@{}ll@{}}
\toprule
Condition & Context given to the external reader \\
\midrule
Full context & The chef mixed the dough and \textbf{put it in the} \\
Last four tokens & \textbf{put it in the} \\
Older words shuffled & The dough and chef the mixed \textbf{put it in the} \\
\bottomrule
\end{tabular}
\end{center}
Shuffling preserves the initial word, recent tokens, and older word
multiset. Hypothetical target probabilities of 20\%, 5\%, and 10\%
under these contexts give gains of $\log(0.20/0.05)\approx1.39$ nats
for retaining older context and $\log(0.20/0.10)\approx0.69$ nats for
retaining its order. Positive gains favor the original context;
negative gains favor the altered context.

We apply this test to all 1,300 original texts, including real text,
using GPT-2-large in FP32 and its subword tokens, with $k\in\{4,8,16\}$.
We use spans between internal sentence markers and, with a fixed seed,
uniformly sample at most 16 targets per text and $k$ that admit a nonidentity shuffle.
Shuffling keeps word groups intact and fixes the initial group, recent
$k$ tokens, and any group crossing the old/recent boundary; truncation retains original
position indices. Target sets can differ across $k$.

$G_k^{\mathrm{short}}$ and $G_k^{\mathrm{shuffle}}$ average the
full-minus-truncated and full-minus-shuffled log-probabilities over targets.
For 95\% percentile intervals, we resample texts within generation seeds,
retaining all distances, conditions, and texts with no eligible targets: 1,000
replicates for gains and 10,000 for post-hoc pairwise differences, sampling
methods independently. Intervals condition on one training run per model and the
fixed scorer; pairwise intervals are pointwise and unadjusted for nine
comparisons.

\begin{table}[htbp]
\color{black}
\centering
\small
\setlength{\tabcolsep}{4pt}
\caption{External-context diagnostic: gains in nats per selected target
with 95\% bootstrap intervals. Coverage is the percentage of original
texts with eligible targets; Targets counts selected tokens. Bold marks
the largest mean shuffle gain among generators at each $k$.
Truncation can change scorer calibration.}
\label{tab:lm1b-context-diagnostic}
\begin{tabular}{lrrrrr}
\toprule
Corpus & $k$ & Coverage (\%) & Targets & $G_k^{\mathrm{short}}$ & $G_k^{\mathrm{shuffle}}$ \\
\midrule
RJD-LOO & 4 & 99.3 & 4,766 & \shortstack{$4.700$\\{\scriptsize$[4.574,4.832]$}} & \shortstack{$\mathbf{0.154}$\\{\scriptsize$[0.118,0.189]$}} \\
 & 8 & 99.3 & 4,744 & \shortstack{$4.279$\\{\scriptsize$[4.161,4.406]$}} & \shortstack{$\mathbf{0.025}$\\{\scriptsize$[-0.003,0.052]$}} \\
 & 16 & 97.0 & 4,233 & \shortstack{$3.722$\\{\scriptsize$[3.597,3.835]$}} & \shortstack{$\mathbf{-0.017}$\\{\scriptsize$[-0.036,0.002]$}} \\
\addlinespace[2pt]
RJD & 4 & 99.7 & 4,764 & \shortstack{$3.346$\\{\scriptsize$[3.247,3.457]$}} & \shortstack{$-0.090$\\{\scriptsize$[-0.117,-0.060]$}} \\
 & 8 & 99.7 & 4,722 & \shortstack{$3.145$\\{\scriptsize$[3.036,3.241]$}} & \shortstack{$-0.092$\\{\scriptsize$[-0.113,-0.071]$}} \\
 & 16 & 97.0 & 4,317 & \shortstack{$2.928$\\{\scriptsize$[2.826,3.028]$}} & \shortstack{$-0.051$\\{\scriptsize$[-0.068,-0.035]$}} \\
\addlinespace[2pt]
RJD-SA & 4 & 99.3 & 4,760 & \shortstack{$3.780$\\{\scriptsize$[3.657,3.896]$}} & \shortstack{$-0.001$\\{\scriptsize$[-0.034,0.032]$}} \\
 & 8 & 99.3 & 4,741 & \shortstack{$3.549$\\{\scriptsize$[3.455,3.647]$}} & \shortstack{$-0.039$\\{\scriptsize$[-0.063,-0.015]$}} \\
 & 16 & 98.7 & 4,506 & \shortstack{$3.184$\\{\scriptsize$[3.084,3.286]$}} & \shortstack{$-0.029$\\{\scriptsize$[-0.048,-0.011]$}} \\
\addlinespace[2pt]
EPJ & 4 & 99.0 & 4,743 & \shortstack{$4.270$\\{\scriptsize$[4.146,4.388]$}} & \shortstack{$0.070$\\{\scriptsize$[0.036,0.103]$}} \\
 & 8 & 99.0 & 4,725 & \shortstack{$3.887$\\{\scriptsize$[3.764,4.009]$}} & \shortstack{$-0.015$\\{\scriptsize$[-0.043,0.012]$}} \\
 & 16 & 98.3 & 4,437 & \shortstack{$3.467$\\{\scriptsize$[3.328,3.592]$}} & \shortstack{$-0.033$\\{\scriptsize$[-0.051,-0.014]$}} \\
\addlinespace[2pt]
Real LM1B & 4 & 98.0 & 1,565 & \shortstack{$5.770$\\{\scriptsize$[5.531,5.992]$}} & \shortstack{$0.749$\\{\scriptsize$[0.676,0.825]$}} \\
 & 8 & 98.0 & 1,549 & \shortstack{$5.511$\\{\scriptsize$[5.261,5.773]$}} & \shortstack{$0.430$\\{\scriptsize$[0.362,0.497]$}} \\
 & 16 & 97.0 & 1,431 & \shortstack{$4.826$\\{\scriptsize$[4.621,5.033]$}} & \shortstack{$0.173$\\{\scriptsize$[0.131,0.218]$}} \\
\bottomrule
\end{tabular}
\end{table}

\begin{table}[htbp]
\color{black}
\centering
\small
\setlength{\tabcolsep}{8pt}
\caption{Difference in context-shuffling gain: RJD-LOO minus the
specified competitor, in nats per selected target. Positive values
favor RJD-LOO. Pointwise 95\% bootstrap intervals appear underneath;
these post-hoc intervals are not adjusted for multiple comparisons.}
\label{tab:lm1b-context-pairwise}
\begin{tabular}{lrrr}
\toprule
Competitor & $k=4$ & $k=8$ & $k=16$ \\
\midrule
RJD & \shortstack{$0.244$\\{\scriptsize$[0.197,0.291]$}} & \shortstack{$0.117$\\{\scriptsize$[0.083,0.151]$}} & \shortstack{$0.034$\\{\scriptsize$[0.008,0.059]$}} \\
\addlinespace[2pt]
RJD-SA & \shortstack{$0.155$\\{\scriptsize$[0.106,0.203]$}} & \shortstack{$0.064$\\{\scriptsize$[0.029,0.101]$}} & \shortstack{$0.012$\\{\scriptsize$[-0.015,0.039]$}} \\
\addlinespace[2pt]
EPJ & \shortstack{$0.083$\\{\scriptsize$[0.034,0.133]$}} & \shortstack{$0.040$\\{\scriptsize$[0.003,0.078]$}} & \shortstack{$0.016$\\{\scriptsize$[-0.012,0.044]$}} \\
\bottomrule
\end{tabular}
\end{table}

\paragraph{Context results and interpretation.}
RJD-LOO has the largest shuffle gain among generators, with a clearly
positive absolute gain only for $k=4$. Real text benefits at all three
values of $k$ (Table~\ref{tab:lm1b-context-diagnostic}).
Pairwise intervals favor RJD-LOO over all competitors for $k=4,8$,
and over RJD alone for $k=16$
(Table~\ref{tab:lm1b-context-pairwise}).
Truncation gains can reflect boundary and position-calibration effects;
length-preserving shuffling controls these changes, while repetition
and topic can still affect predictability.

\FloatBarrier
\endgroup

\clearpage
\section{ZINC250k experimental details}
\label{app:zinc-details}
\label{sec:zinc}

\subsection{Data, representation, and predictor}

We use the GDSS benchmark CSV and test indices at commit
\href{https://github.com/harryjo97/GDSS/tree/24cc490e0c5b39cbd265fda33ab2bc8b528d4454}{\texttt{24cc490e0c5b}}
\citep{jo2022gdss}. From its 224,568 training examples, we reserve 5,000
for development using split seed 314159 and grouping identical
non-isomeric canonical SMILES, leaving 219,568 fitting examples. Together
with the 24,887 official test examples, these comprise 249,455 molecules.
The input molecules are connected.
Fitting and development identities do not
overlap. The official test split is unchanged; 295 test entries share
fitting identities after non-isomeric canonicalization. This inherited
overlap is retained and recorded. Test examples supply neither training
batches nor the atom-count distribution. Holding out development examples
reduces the fitting set relative to methods using the entire published
training pool.

Graphs have at most 38 heavy atoms and are represented by
$L=38+\binom{38}{2}=741$ coordinates. Atom counts are sampled from the
fitting-set distribution.
The nine atom types are
$\{\mathrm C,\mathrm N,\mathrm O,\mathrm F,\mathrm P,\mathrm S,
\mathrm{Cl},\mathrm{Br},\mathrm I\}$; bond labels are no bond, single,
double, and triple. Kekulization encodes aromatic bonds through single and double bonds.
Hydrogens are implicit; the model predicts neither stereochemistry nor
three-dimensional coordinates nor separate formal-charge labels.
Decoding follows the GDSS positive-charge convention where applicable.
Given the sampled atom count $n$, the first $n$ atom slots and all
$\binom n2$ corresponding pair slots are active, including pairs labeled
``no bond.'' Padding is an external mask, not an atom category; inactive
slots are excluded from the loss and attention sources.

The atom and bond embeddings are learned vectors in $\R^{64}$, each
normalized to norm $8$.

The predictor has eight blocks, width 256, eight attention heads,
32 time features, 32 context channels, dropout $0.1$, AdaLN conditioning,
and SwiGLU layers. The atom and bond classifiers share this backbone;
they are not separate drift and jump heads. Its 6,923,552 parameters
and the $(9+4)\times64=832$ embedding parameters total 6,924,384 for each
of the five trained models. Parameters have matched shapes and
initializations within each seed.

An atom query aggregates features of other atoms and their incident
bonds; a bond query uses symmetric endpoint features and pairs of bonds
through a third atom. Each such context excludes the target coordinate's
state and absorption indicator. All features use noisy observations,
without access to the clean adjacency. Attention keys and values are
fixed functions of individual input coordinates, as in
\eqref{eq:architecture}; updated queries never serve as sources.
LOO excludes the target source in every layer. Allowing this source is
the only architectural change for predictors of $p_{i,t}(\cdot\mid X_t)$.
Shared features and the absence of positional embeddings give equivariance
under joint permutations of atom slots and their induced pair slots.

\subsection{Training and sampling}

With $u=t^2$, a continuous coordinate before absorption has conditional
law $\mathcal N(u e(y_i),u(1-u)I_{64})$ and starts at zero. Jump coordinates
have first-jump CDF $t^2$ and hazard $2t/(1-t^2)$; after jumping they remain
at the selected anchor. An observed indicator records absorption. In EPJ,
the coordinate remains zero before its jump. For unabsorbed continuous
coordinates, Bayesian completion uses the Gaussian likelihood above.
PD-BC completes the LOO prediction for the drift, and AJD-BC-D does so
only for the drift, retaining LOO jump probabilities. EPJ uses
$p_{\theta,i,t}$. Table~\ref{tab:zinc-methods} maps the seven sampling
variants to the five trained models. PD-LOO and PD-BC share one checkpoint,
as do AJD-LOO and AJD-BC-D; completion is a sampling-time operation.
The absorbing process differs from RJD: continuous motion stops after a
jump, and Proposition~\ref{prop:contextual-generator} is not a
marginal-preservation result for this augmented absorbing process.

\begin{table}[htbp]
\centering
\small
\setlength{\tabcolsep}{5pt}
\caption{ZINC250k methods. All predictors have the same architecture and
parameter count. Unrestricted and LOO prediction differ only in allowing
or excluding the target input as an attention source. BC denotes analytic
likelihood completion; all jump destinations use CAT in the test comparison.}
\label{tab:zinc-methods}
\begin{tabular}{lllll}
\toprule
Method & Conditional path & Predictor & Drift & Jump probabilities \\
\midrule
EPJ & Pure jump & Unrestricted & -- & Unrestricted \\
PD & Diffusion & Unrestricted & Unrestricted & -- \\
PD-LOO & Diffusion & LOO & Raw LOO & -- \\
PD-BC & Diffusion & LOO & BC & -- \\
AJD & Absorbing jump diffusion & Unrestricted & Unrestricted & Unrestricted \\
AJD-LOO & Absorbing jump diffusion & LOO & Raw LOO & Raw LOO \\
AJD-BC-D & Absorbing jump diffusion & LOO & BC & Raw LOO \\
\bottomrule
\end{tabular}
\end{table}

Training samples $t$ uniformly from $[0,0.999)$ and minimizes
$\tfrac12\mathcal L_{\rm atom}+\tfrac12\mathcal L_{\rm bond}$,
where each cross-entropy is normalized over its eligible coordinates.
These are all active coordinates for PD and only unabsorbed active
coordinates for jump paths. No additional bond-class weights are used.
AdamW uses batch size 512, accumulated in microbatches of 128,
learning rate $10^{-3}$ after 1,000 warmup
updates, cosine decay to $10^{-4}$ at 40,000 updates, zero weight decay,
and gradient-norm clipping at one. EMA includes predictor and anchors,
with decay $\min(0.999,(1+s)/(10+s))$ at update $s$.
The test comparison uses the EMA weights at exactly 40,000 updates,
rather than selecting checkpoints by validation loss. Computation uses
FP32 with TF32 disabled.

For $M$ NFE, the sampling grid is
$t_j=1-\cos(\pi j/(2M))$, $j=0,\ldots,M$, with $u_j=t_j^2$.
There is one predictor call at each left endpoint. Given its predicted
anchor mean $m$, the continuous update is
\[
 x' = m+\alpha(x-m)
       +\sqrt{(u_{j+1}-u_j)\alpha}\,\varepsilon,
 \qquad \alpha=\frac{1-u_{j+1}}{1-u_j},
 \qquad \varepsilon\sim\mathcal N(0,I_{64}).
\]
An eligible coordinate jumps with probability
$(u_{j+1}-u_j)/(1-u_j)$. Thus the final interval reaches $u=1$ without
a network call at $t=1$, and $M$ NFE counts exactly $M$ backbone calls.
Pure diffusion is decoded by nearest anchor, equivalently maximum anchor
inner product. Test jump destinations use CAT at temperature one,
without correction steps. Except for the evaluation specified below,
atom counts and generation seeds are paired across methods, although
different samplers consume random numbers differently. Evaluation uses
batches of 32 graphs and sampling seed base 1729. For AJD-LOO at 512 NFE
with training seed 1, we use an independent H100 evaluation with batch
size 256 and sampling seed base 271828, keeping the same 40,000-update
EMA checkpoint, strict LOO predictor, and CAT sampler without BC.
This evaluation has the same sample count and test reference but does
not share the paired atom-count and sampling-seed realization.
EPJ, PD, PD-BC, AJD, and AJD-BC-D use H100 for both seeds.
PD-LOO and AJD-LOO evaluations use V100 for seed 0 and H100 for seed 1 at
64 and 128 NFE; 512-NFE evaluations also use A100 GPUs.
Table~\ref{tab:zinc-secondary} reports generation time and hardware
separately for each seed, excluding scoring and queue waits. These
times are not a comparison of sampler speed on a common GPU and batch size.

\subsection{Molecular evaluation metrics}
\label{app:zinc-metrics}

Validity (no correction) is the number of outputs passing full RDKit
sanitization before repair divided by the number generated. Valid \& connected
additionally requires exactly one connected component, with the same
denominator; it is not connectivity conditional on validity.
For comparison with GDSS \citep{jo2022gdss}, we separately report the
fraction passing its valence check before bond correction, labeled
GDSS valence (no correction). This uses the GDSS \texttt{check\_valency}
routine rather than full sanitization; the two checks need not agree,
and neither requires a connected graph.
Corrected outputs use GDSS bond downgrading followed by largest-component
selection; we compute their metrics separately from raw outputs.

FCD compares the empirical means and covariances of pretrained ChemNet
features \citep{preuer2018fcd}, using \texttt{fcd-torch} 1.0.7. For NSPDK,
we use normalized GDSS/EDeN graph features at complexity four
\citep{costa2010nspdk,jo2022gdss}. If $\phi(G)$ denotes these features,
the reported statistic is
\[
 \widehat{\mathrm{MMD}}^{\,2}
 =\left\|\frac1{N}\sum_{r=1}^{N}\phi(G_r)
       -\frac1{N_{\rm ref}}\sum_{r=1}^{N_{\rm ref}}\phi(G_r^{\rm ref})
  \right\|_2^2,
\]
the biased empirical squared MMD for the corresponding linear kernel.
Here $N$ counts valid raw outputs or valid corrected outputs, respectively;
invalid outputs do not enter either distributional metric. Their frequency
is therefore reported separately by validity. The FCD reference retains
the benchmark SMILES, including stereochemical information where present.

Uniqueness is the number of distinct valid non-isomeric canonical SMILES
divided by the number of valid outputs. Novelty is the fraction of distinct
valid identities absent from the training reference, reported against both
the fitting set and the full published training pool. The fitting-copy
fraction instead counts repetitions and divides the number of valid
samples matching fitting identities by the valid sample count. Raw
identities retain all components; corrected identities retain only the
selected component. Component pruning is the fraction of generated
graphs for which decoding discards components after bond repair.

The test evaluation uses 10,000 attempted graphs per method, NFE budget,
and training seed, with all 24,887 test molecules as reference. Invalid
outputs are not replaced to reach a target number of valid molecules.
The tables report training seeds 0 and 1. Each metric is computed
separately per seed and summarized by its mean and sample standard
deviation; generated collections are not pooled for FCD. The development
sweep instead uses seed 0, 1,000 graphs,
and all 5,000 development references. Absolute FCD values across these
protocols are not a matched measure of training progress.
Evaluation uses RDKit 2025.03.6, NumPy 2.2.6, NetworkX 3.4.2, and a fixed
Python hash seed of zero for the NSPDK feature map.

\subsection{Test results at 40,000 updates}
\label{app:zinc-results}

Table~\ref{tab:zinc-complete-paths} extends the main comparison to 64 and
512 NFE and includes corrected NSPDK. At 128 NFE, raw LOO drift gives
AJD the lowest mean FCD, while completion gives it the lowest mean NSPDK; these
metrics therefore rank the drift choices differently. Increasing NFE
improves EPJ and AJD-BC-D raw validity and connected validity, whereas PD
and PD-BC change less. At 512 NFE, AJD-LOO has the highest mean connected
validity and lowest mean raw FCD, while AJD-BC-D has the lowest mean
corrected FCD and NSPDK. These rankings describe means over two seeds.

\begin{table}[htbp]
\centering
\footnotesize
\setlength{\tabcolsep}{3pt}
\caption{ZINC250k test results: 40,000-update EMA weights, training seeds
0 and 1, 10,000 attempted graphs per seed, and all 24,887 test references.
Each cell shows the mean above $\pm$ sample standard deviation.
All jumps use CAT and $\kappa(t)=t^2$.
The validity percentages use all generated graphs as denominator.
NSPDK is squared MMD. Corrected metrics use bond repair and
largest-component selection. The seed-1 AJD-LOO evaluation at 512 NFE
uses the independent sampling realization described above.}
\label{tab:zinc-complete-paths}
\label{tab:zinc-paths}
\begin{tabular}{lrrrrrrr}
\toprule
& & \multicolumn{4}{c}{Raw} & \multicolumn{2}{c}{Corrected} \\
\cmidrule(lr){3-6}\cmidrule(lr){7-8}
Method & NFE & \shortstack{Valid\\(\%) $\uparrow$} & \shortstack{Valid \& conn.\\(\%) $\uparrow$} & FCD $\downarrow$ & NSPDK $\downarrow$ & FCD $\downarrow$ & NSPDK $\downarrow$ \\
\midrule
EPJ & 64 & \shortstack{$59.88$\\$\pm 0.45$} & \shortstack{$34.25$\\$\pm 0.62$} & \shortstack{$21.375$\\$\pm 0.159$} & \shortstack{$0.04826$\\$\pm 0.00035$} & \shortstack{$21.534$\\$\pm 0.028$} & \shortstack{$0.05192$\\$\pm 0.00037$} \\
PD & 64 & \shortstack{$88.98$\\$\pm 0.52$} & \shortstack{$45.36$\\$\pm 0.01$} & \shortstack{$22.178$\\$\pm 0.022$} & \shortstack{$0.05229$\\$\pm 0.00021$} & \shortstack{$21.933$\\$\pm 0.015$} & \shortstack{$0.05197$\\$\pm 0.00014$} \\
PD-LOO & 64 & \shortstack{$52.56$\\$\pm 5.35$} & \shortstack{$12.03$\\$\pm 3.87$} & \shortstack{$28.285$\\$\pm 2.367$} & \shortstack{$0.27730$\\$\pm 0.07328$} & \shortstack{$25.727$\\$\pm 2.735$} & \shortstack{$0.18862$\\$\pm 0.04773$} \\
PD-BC & 64 & \shortstack{$91.33$\\$\pm 0.05$} & \shortstack{$50.61$\\$\pm 2.76$} & \shortstack{$21.694$\\$\pm 0.025$} & \shortstack{$0.05189$\\$\pm 0.00098$} & \shortstack{$21.544$\\$\pm 0.069$} & \shortstack{$0.05159$\\$\pm 0.00091$} \\
AJD & 64 & \shortstack{$86.54$\\$\pm 1.03$} & \shortstack{$43.80$\\$\pm 0.97$} & \shortstack{$22.309$\\$\pm 0.029$} & \shortstack{$0.05135$\\$\pm 0.00066$} & \shortstack{$22.051$\\$\pm 0.047$} & \shortstack{$0.05129$\\$\pm 0.00053$} \\
AJD-LOO & 64 & \shortstack{$72.11$\\$\pm 0.95$} & \shortstack{$61.27$\\$\pm 1.32$} & \shortstack{$20.827$\\$\pm 0.078$} & \shortstack{$0.05063$\\$\pm 0.00081$} & \shortstack{$21.506$\\$\pm 0.113$} & \shortstack{$0.05254$\\$\pm 0.00039$} \\
AJD-BC-D & 64 & \shortstack{$65.82$\\$\pm 1.27$} & \shortstack{$48.27$\\$\pm 1.38$} & \shortstack{$20.786$\\$\pm 0.110$} & \shortstack{$0.04420$\\$\pm 0.00079$} & \shortstack{$21.304$\\$\pm 0.086$} & \shortstack{$0.04657$\\$\pm 0.00038$} \\
\midrule
EPJ & 128 & \shortstack{$69.02$\\$\pm 0.22$} & \shortstack{$41.81$\\$\pm 0.11$} & \shortstack{$21.190$\\$\pm 0.039$} & \shortstack{$0.04882$\\$\pm 0.00088$} & \shortstack{$21.305$\\$\pm 0.002$} & \shortstack{$0.05142$\\$\pm 0.00048$} \\
PD & 128 & \shortstack{$89.98$\\$\pm 0.33$} & \shortstack{$49.97$\\$\pm 0.37$} & \shortstack{$21.919$\\$\pm 0.031$} & \shortstack{$0.05154$\\$\pm 0.00015$} & \shortstack{$21.740$\\$\pm 0.003$} & \shortstack{$0.05127$\\$\pm 0.00009$} \\
PD-LOO & 128 & \shortstack{$55.38$\\$\pm 5.42$} & \shortstack{$20.07$\\$\pm 3.91$} & \shortstack{$24.966$\\$\pm 1.164$} & \shortstack{$0.15202$\\$\pm 0.02161$} & \shortstack{$21.670$\\$\pm 1.250$} & \shortstack{$0.10993$\\$\pm 0.00862$} \\
PD-BC & 128 & \shortstack{$91.84$\\$\pm 0.40$} & \shortstack{$54.19$\\$\pm 2.21$} & \shortstack{$21.642$\\$\pm 0.066$} & \shortstack{$0.05149$\\$\pm 0.00088$} & \shortstack{$21.535$\\$\pm 0.090$} & \shortstack{$0.05134$\\$\pm 0.00057$} \\
AJD & 128 & \shortstack{$87.41$\\$\pm 0.95$} & \shortstack{$47.82$\\$\pm 2.49$} & \shortstack{$22.065$\\$\pm 0.088$} & \shortstack{$0.05054$\\$\pm 0.00043$} & \shortstack{$21.943$\\$\pm 0.082$} & \shortstack{$0.05062$\\$\pm 0.00011$} \\
AJD-LOO & 128 & \shortstack{$83.96$\\$\pm 0.52$} & \shortstack{$71.07$\\$\pm 0.61$} & \shortstack{$20.460$\\$\pm 0.019$} & \shortstack{$0.05141$\\$\pm 0.00011$} & \shortstack{$20.968$\\$\pm 0.015$} & \shortstack{$0.05242$\\$\pm 0.00019$} \\
AJD-BC-D & 128 & \shortstack{$78.53$\\$\pm 0.16$} & \shortstack{$59.11$\\$\pm 0.38$} & \shortstack{$20.677$\\$\pm 0.090$} & \shortstack{$0.04629$\\$\pm 0.00003$} & \shortstack{$20.990$\\$\pm 0.053$} & \shortstack{$0.04715$\\$\pm 0.00024$} \\
\midrule
EPJ & 512 & \shortstack{$77.12$\\$\pm 0.15$} & \shortstack{$49.34$\\$\pm 0.25$} & \shortstack{$20.921$\\$\pm 0.041$} & \shortstack{$0.04978$\\$\pm 0.00014$} & \shortstack{$21.119$\\$\pm 0.056$} & \shortstack{$0.05157$\\$\pm 0.00013$} \\
PD & 512 & \shortstack{$89.94$\\$\pm 0.35$} & \shortstack{$51.38$\\$\pm 0.73$} & \shortstack{$21.693$\\$\pm 0.097$} & \shortstack{$0.05081$\\$\pm 0.00059$} & \shortstack{$21.636$\\$\pm 0.093$} & \shortstack{$0.05079$\\$\pm 0.00052$} \\
PD-LOO & 512 & \shortstack{$59.42$\\$\pm 5.88$} & \shortstack{$24.91$\\$\pm 4.55$} & \shortstack{$24.185$\\$\pm 1.237$} & \shortstack{$0.13591$\\$\pm 0.01791$} & \shortstack{$21.281$\\$\pm 1.161$} & \shortstack{$0.10386$\\$\pm 0.00698$} \\
PD-BC & 512 & \shortstack{$92.55$\\$\pm 0.08$} & \shortstack{$55.77$\\$\pm 2.77$} & \shortstack{$21.264$\\$\pm 0.117$} & \shortstack{$0.05061$\\$\pm 0.00046$} & \shortstack{$21.235$\\$\pm 0.176$} & \shortstack{$0.05073$\\$\pm 0.00028$} \\
AJD & 512 & \shortstack{$87.95$\\$\pm 0.95$} & \shortstack{$49.94$\\$\pm 2.74$} & \shortstack{$21.929$\\$\pm 0.001$} & \shortstack{$0.04994$\\$\pm 0.00054$} & \shortstack{$21.858$\\$\pm 0.154$} & \shortstack{$0.05013$\\$\pm 0.00015$} \\
AJD-LOO & 512 & \shortstack{$91.08$\\$\pm 0.06$} & \shortstack{$76.21$\\$\pm 0.25$} & \shortstack{$20.575$\\$\pm 0.150$} & \shortstack{$0.05357$\\$\pm 0.00006$} & \shortstack{$20.908$\\$\pm 0.133$} & \shortstack{$0.05375$\\$\pm 0.00017$} \\
AJD-BC-D & 512 & \shortstack{$86.84$\\$\pm 0.28$} & \shortstack{$65.91$\\$\pm 0.80$} & \shortstack{$20.615$\\$\pm 0.045$} & \shortstack{$0.04663$\\$\pm 0.00030$} & \shortstack{$20.843$\\$\pm 0.056$} & \shortstack{$0.04697$\\$\pm 0.00035$} \\
\bottomrule
\end{tabular}
\end{table}

\paragraph{Secondary metrics and generation time.}
All 42 test evaluations have 100\% corrected validity and no decoding
errors. Novelty against both training references is 100\%, with one
exception: PD-LOO at 128 NFE for seed 1 contains one fitting-set copy,
giving raw and corrected novelty of 99.9825\% and 99.9867\%, respectively.
Its raw and corrected fitting-copy fractions are $1/5921$ and $1/10000$;
all other evaluations have zero observed fitting copies.
Raw uniqueness is 100\% except for PD-LOO; corrected uniqueness is also
100\% except for PD-LOO and AJD at 128 NFE for seed 0 and 512 NFE for
seed 1 (99.99\% for each of these two AJD evaluations).
Table~\ref{tab:zinc-secondary} gives the nonconstant secondary metrics.
The low uniqueness of PD-LOO shows an additional limitation of its raw
LOO drift. High novelty or corrected validity alone does not establish
good raw validity, connectivity, or distributional agreement.

\begin{table}[htbp]
\centering
\footnotesize
\setlength{\tabcolsep}{3pt}
\caption{Secondary ZINC250k metrics and generation time for the same
two-seed test samples as Table~\ref{tab:zinc-complete-paths}. GDSS valence
is checked before correction. Uniqueness uses valid outputs as denominator;
pruning uses all attempted graphs. All four metric columns are percentages,
with the mean above $\pm$ sample standard deviation. Each time column
shows generation minutes for 10,000 graphs above the GPU used, separately
for each seed; scoring and queue waits are excluded. A slash indicates
generation used both listed GPU types. Batch size is 32 except for
seed-1 AJD-LOO at 512 NFE (256).}
\label{tab:zinc-secondary}
\begin{tabular}{lrrrrrrr}
\toprule
Method & NFE & \shortstack{GDSS\\valence} & \shortstack{Raw\\unique} & \shortstack{Corrected\\unique} & \shortstack{Component\\pruned} & \shortstack{Seed 0\\min. / GPU} & \shortstack{Seed 1\\min. / GPU} \\
\midrule
EPJ & 64 & \shortstack{$59.86$\\$\pm 0.45$} & \shortstack{$100.00$\\$\pm 0.00$} & \shortstack{$100.00$\\$\pm 0.00$} & \shortstack{$41.56$\\$\pm 0.59$} & \shortstack{$15.5$\\H100} & \shortstack{$15.6$\\H100} \\
PD & 64 & \shortstack{$88.95$\\$\pm 0.53$} & \shortstack{$100.00$\\$\pm 0.00$} & \shortstack{$100.00$\\$\pm 0.00$} & \shortstack{$48.44$\\$\pm 0.16$} & \shortstack{$15.4$\\H100} & \shortstack{$15.6$\\H100} \\
PD-LOO & 64 & \shortstack{$52.56$\\$\pm 5.35$} & \shortstack{$67.85$\\$\pm 10.75$} & \shortstack{$49.96$\\$\pm 9.07$} & \shortstack{$79.54$\\$\pm 5.61$} & \shortstack{$54.1$\\V100} & \shortstack{$15.4$\\H100} \\
PD-BC & 64 & \shortstack{$91.26$\\$\pm 0.10$} & \shortstack{$100.00$\\$\pm 0.00$} & \shortstack{$100.00$\\$\pm 0.00$} & \shortstack{$44.30$\\$\pm 3.05$} & \shortstack{$15.6$\\H100} & \shortstack{$15.5$\\H100} \\
AJD & 64 & \shortstack{$86.49$\\$\pm 1.02$} & \shortstack{$100.00$\\$\pm 0.00$} & \shortstack{$100.00$\\$\pm 0.00$} & \shortstack{$48.73$\\$\pm 1.73$} & \shortstack{$15.5$\\H100} & \shortstack{$15.5$\\H100} \\
AJD-LOO & 64 & \shortstack{$72.11$\\$\pm 0.95$} & \shortstack{$100.00$\\$\pm 0.00$} & \shortstack{$100.00$\\$\pm 0.00$} & \shortstack{$14.91$\\$\pm 0.74$} & \shortstack{$54.1$\\V100} & \shortstack{$15.5$\\H100} \\
AJD-BC-D & 64 & \shortstack{$65.81$\\$\pm 1.29$} & \shortstack{$100.00$\\$\pm 0.00$} & \shortstack{$100.00$\\$\pm 0.00$} & \shortstack{$26.46$\\$\pm 0.70$} & \shortstack{$15.6$\\H100} & \shortstack{$15.7$\\H100} \\
\midrule
EPJ & 128 & \shortstack{$69.00$\\$\pm 0.19$} & \shortstack{$100.00$\\$\pm 0.00$} & \shortstack{$100.00$\\$\pm 0.00$} & \shortstack{$38.32$\\$\pm 0.07$} & \shortstack{$30.9$\\H100} & \shortstack{$30.8$\\H100} \\
PD & 128 & \shortstack{$89.94$\\$\pm 0.34$} & \shortstack{$100.00$\\$\pm 0.00$} & \shortstack{$100.00$\\$\pm 0.00$} & \shortstack{$44.15$\\$\pm 0.06$} & \shortstack{$30.9$\\H100} & \shortstack{$31.0$\\H100} \\
PD-LOO & 128 & \shortstack{$55.37$\\$\pm 5.43$} & \shortstack{$92.78$\\$\pm 5.10$} & \shortstack{$71.76$\\$\pm 5.01$} & \shortstack{$67.68$\\$\pm 4.50$} & \shortstack{$108.2$\\V100} & \shortstack{$31.0$\\H100} \\
PD-BC & 128 & \shortstack{$91.81$\\$\pm 0.39$} & \shortstack{$100.00$\\$\pm 0.00$} & \shortstack{$100.00$\\$\pm 0.00$} & \shortstack{$40.61$\\$\pm 2.84$} & \shortstack{$31.1$\\H100} & \shortstack{$31.1$\\H100} \\
AJD & 128 & \shortstack{$87.36$\\$\pm 0.96$} & \shortstack{$100.00$\\$\pm 0.00$} & \shortstack{$100.00$\\$\pm 0.01$} & \shortstack{$44.72$\\$\pm 3.34$} & \shortstack{$31.0$\\H100} & \shortstack{$30.9$\\H100} \\
AJD-LOO & 128 & \shortstack{$83.95$\\$\pm 0.52$} & \shortstack{$100.00$\\$\pm 0.00$} & \shortstack{$100.00$\\$\pm 0.00$} & \shortstack{$15.32$\\$\pm 0.50$} & \shortstack{$108.4$\\V100} & \shortstack{$32.8$\\H100} \\
AJD-BC-D & 128 & \shortstack{$78.50$\\$\pm 0.15$} & \shortstack{$100.00$\\$\pm 0.00$} & \shortstack{$100.00$\\$\pm 0.00$} & \shortstack{$24.75$\\$\pm 0.62$} & \shortstack{$31.1$\\H100} & \shortstack{$31.1$\\H100} \\
\midrule
EPJ & 512 & \shortstack{$77.11$\\$\pm 0.15$} & \shortstack{$100.00$\\$\pm 0.00$} & \shortstack{$100.00$\\$\pm 0.00$} & \shortstack{$35.61$\\$\pm 0.16$} & \shortstack{$123.7$\\H100} & \shortstack{$125.7$\\H100} \\
PD & 512 & \shortstack{$89.88$\\$\pm 0.33$} & \shortstack{$100.00$\\$\pm 0.00$} & \shortstack{$100.00$\\$\pm 0.00$} & \shortstack{$42.37$\\$\pm 0.70$} & \shortstack{$123.7$\\H100} & \shortstack{$123.6$\\H100} \\
PD-LOO & 512 & \shortstack{$59.41$\\$\pm 5.87$} & \shortstack{$95.54$\\$\pm 4.17$} & \shortstack{$76.41$\\$\pm 4.17$} & \shortstack{$63.73$\\$\pm 4.50$} & \shortstack{$258.8$\\A100/V100} & \shortstack{$138.3$\\A100/H100} \\
PD-BC & 512 & \shortstack{$92.53$\\$\pm 0.11$} & \shortstack{$100.00$\\$\pm 0.00$} & \shortstack{$100.00$\\$\pm 0.00$} & \shortstack{$39.38$\\$\pm 2.97$} & \shortstack{$124.4$\\H100} & \shortstack{$124.3$\\H100} \\
AJD & 512 & \shortstack{$87.91$\\$\pm 0.93$} & \shortstack{$100.00$\\$\pm 0.00$} & \shortstack{$100.00$\\$\pm 0.01$} & \shortstack{$42.72$\\$\pm 3.46$} & \shortstack{$123.8$\\H100} & \shortstack{$123.5$\\H100} \\
AJD-LOO & 512 & \shortstack{$91.08$\\$\pm 0.06$} & \shortstack{$100.00$\\$\pm 0.00$} & \shortstack{$100.00$\\$\pm 0.00$} & \shortstack{$16.11$\\$\pm 0.09$} & \shortstack{$258.8$\\A100/V100} & \shortstack{$114.4$\\H100} \\
AJD-BC-D & 512 & \shortstack{$86.83$\\$\pm 0.28$} & \shortstack{$100.00$\\$\pm 0.00$} & \shortstack{$100.00$\\$\pm 0.00$} & \shortstack{$24.16$\\$\pm 0.37$} & \shortstack{$124.4$\\H100} & \shortstack{$124.2$\\H100} \\
\bottomrule
\end{tabular}
\end{table}

\FloatBarrier

\subsection{Development sampling ablations at 20,000 updates}
\label{app:zinc-sampling}

We separately compare the five sampling rules from
\textcolor{black}{Appendix~\ref{sec:text8-sampling}} on fixed 20,000-update EMA checkpoints:
CAT, MAX, TEMP with temperature $0.7$, TOPP with mass $0.9$, and CONF.
For CONF, event counts are preserved separately for atom and bond
coordinates, then assigned to the most confident eligible coordinates
within each group. The drift remains at temperature one and retains
the prediction choice in Table~\ref{tab:zinc-methods}; only jump sampling
is varied. These rules change the sampling dynamics and are not covered
by the repeated-reset marginal-preservation guarantee.

Tables~\ref{tab:zinc-sampling-64} and~\ref{tab:zinc-sampling-128} report
the complete seed-0 sweep with 1,000 generated graphs per setting and
5,000 development references. MAX markedly increases EPJ validity but
worsens FCD and NSPDK. For AJD-LOO at 128 NFE, TEMP improves validity,
connected validity, and FCD over CAT, while worsening NSPDK; TOPP
improves all six reported metrics over CAT. Thus a sampling rule that
improves validity need not improve distributional agreement, and these
development results do not substitute for the 40,000-update test comparison.

\begin{table}[htbp]
\centering
\footnotesize
\setlength{\tabcolsep}{3pt}
\caption{ZINC250k development sampling sweep at 64 NFE: 20,000-update
EMA weights, seed 0, 1,000 attempted graphs, and 5,000 development
references. Validity percentages use all generated graphs. NSPDK is
squared MMD. These scores use a different checkpoint, sample count,
and reference from the test comparison.}
\label{tab:zinc-sampling-64}
\begin{tabular}{llrrrrrr}
\toprule
& & \multicolumn{4}{c}{Raw} & \multicolumn{2}{c}{Corrected} \\
\cmidrule(lr){3-6}\cmidrule(lr){7-8}
Method & Rule & \shortstack{Valid\\(\%) $\uparrow$} & \shortstack{Valid \& conn.\\(\%) $\uparrow$} & FCD $\downarrow$ & NSPDK $\downarrow$ & FCD $\downarrow$ & NSPDK $\downarrow$ \\
\midrule
EPJ & CAT & $57.8$ & $31.7$ & $23.125$ & $0.05012$ & $22.489$ & $0.05342$ \\
EPJ & MAX & $97.7$ & $21.9$ & $34.198$ & $0.11466$ & $36.581$ & $0.08680$ \\
EPJ & TEMP & $70.7$ & $18.2$ & $26.424$ & $0.05897$ & $23.525$ & $0.05374$ \\
EPJ & TOPP & $83.6$ & $32.6$ & $24.914$ & $0.04133$ & $24.032$ & $0.03467$ \\
EPJ & CONF & $14.6$ & $9.9$ & $27.652$ & $0.05537$ & $25.745$ & $0.07739$ \\
\midrule
AJD & CAT & $76.6$ & $34.2$ & $23.596$ & $0.05172$ & $22.812$ & $0.05234$ \\
AJD & MAX & $76.5$ & $32.8$ & $23.763$ & $0.05328$ & $23.073$ & $0.05375$ \\
AJD & TEMP & $77.4$ & $32.7$ & $23.867$ & $0.05283$ & $23.060$ & $0.05310$ \\
AJD & TOPP & $77.2$ & $33.0$ & $23.619$ & $0.05191$ & $22.834$ & $0.05259$ \\
AJD & CONF & $77.7$ & $36.1$ & $23.400$ & $0.05360$ & $22.905$ & $0.05366$ \\
\midrule
AJD-LOO & CAT & $70.9$ & $61.6$ & $22.173$ & $0.05323$ & $22.594$ & $0.05507$ \\
AJD-LOO & MAX & $80.9$ & $63.3$ & $23.565$ & $0.09184$ & $23.504$ & $0.08439$ \\
AJD-LOO & TEMP & $76.1$ & $63.3$ & $21.908$ & $0.05365$ & $22.354$ & $0.05533$ \\
AJD-LOO & TOPP & $73.8$ & $62.1$ & $21.715$ & $0.04921$ & $21.981$ & $0.05044$ \\
AJD-LOO & CONF & $56.0$ & $32.6$ & $22.605$ & $0.05338$ & $21.350$ & $0.04609$ \\
\midrule
AJD-BC-D & CAT & $63.8$ & $49.7$ & $21.471$ & $0.04470$ & $22.066$ & $0.04739$ \\
AJD-BC-D & MAX & $71.1$ & $55.6$ & $23.430$ & $0.06766$ & $23.425$ & $0.06458$ \\
AJD-BC-D & TEMP & $67.8$ & $53.6$ & $22.092$ & $0.05049$ & $22.111$ & $0.05108$ \\
AJD-BC-D & TOPP & $68.1$ & $53.0$ & $21.242$ & $0.04635$ & $21.664$ & $0.04833$ \\
AJD-BC-D & CONF & $34.2$ & $16.7$ & $23.743$ & $0.04131$ & $21.374$ & $0.05049$ \\
\bottomrule
\end{tabular}
\end{table}

\begin{table}[htbp]
\centering
\footnotesize
\setlength{\tabcolsep}{3pt}
\caption{ZINC250k development sampling sweep at 128 NFE, with the same
20,000-update checkpoints, seed 0, 1,000 attempted graphs, and 5,000
development references as Table~\ref{tab:zinc-sampling-64}.
Metric definitions and sampling rules are unchanged.}
\label{tab:zinc-sampling-128}
\begin{tabular}{llrrrrrr}
\toprule
& & \multicolumn{4}{c}{Raw} & \multicolumn{2}{c}{Corrected} \\
\cmidrule(lr){3-6}\cmidrule(lr){7-8}
Method & Rule & \shortstack{Valid\\(\%) $\uparrow$} & \shortstack{Valid \& conn.\\(\%) $\uparrow$} & FCD $\downarrow$ & NSPDK $\downarrow$ & FCD $\downarrow$ & NSPDK $\downarrow$ \\
\midrule
EPJ & CAT & $66.8$ & $39.3$ & $21.843$ & $0.05129$ & $21.554$ & $0.05298$ \\
EPJ & MAX & $97.4$ & $21.9$ & $33.926$ & $0.13650$ & $36.180$ & $0.10706$ \\
EPJ & TEMP & $77.2$ & $21.7$ & $25.524$ & $0.05885$ & $23.017$ & $0.05377$ \\
EPJ & TOPP & $87.9$ & $37.4$ & $24.598$ & $0.04136$ & $23.504$ & $0.03666$ \\
EPJ & CONF & $21.7$ & $15.9$ & $26.011$ & $0.05571$ & $25.490$ & $0.07004$ \\
\midrule
AJD & CAT & $81.5$ & $39.0$ & $23.132$ & $0.05230$ & $22.468$ & $0.05219$ \\
AJD & MAX & $80.7$ & $39.3$ & $23.215$ & $0.05250$ & $22.594$ & $0.05203$ \\
AJD & TEMP & $81.5$ & $39.3$ & $23.055$ & $0.05184$ & $22.450$ & $0.05188$ \\
AJD & TOPP & $81.1$ & $39.2$ & $23.018$ & $0.05257$ & $22.567$ & $0.05252$ \\
AJD & CONF & $79.7$ & $43.1$ & $22.964$ & $0.05158$ & $22.342$ & $0.05160$ \\
\midrule
AJD-LOO & CAT & $81.9$ & $71.5$ & $22.087$ & $0.05565$ & $22.387$ & $0.05665$ \\
AJD-LOO & MAX & $95.2$ & $79.2$ & $23.468$ & $0.10235$ & $23.382$ & $0.09912$ \\
AJD-LOO & TEMP & $87.6$ & $73.8$ & $21.345$ & $0.06001$ & $21.521$ & $0.05928$ \\
AJD-LOO & TOPP & $89.3$ & $76.2$ & $21.836$ & $0.05439$ & $22.114$ & $0.05415$ \\
AJD-LOO & CONF & $79.9$ & $50.3$ & $22.068$ & $0.05828$ & $21.329$ & $0.05310$ \\
\midrule
AJD-BC-D & CAT & $80.0$ & $63.6$ & $21.680$ & $0.04538$ & $21.854$ & $0.04673$ \\
AJD-BC-D & MAX & $85.0$ & $66.5$ & $22.769$ & $0.07209$ & $22.817$ & $0.06924$ \\
AJD-BC-D & TEMP & $82.2$ & $63.5$ & $21.656$ & $0.05076$ & $21.784$ & $0.05112$ \\
AJD-BC-D & TOPP & $81.3$ & $63.4$ & $21.346$ & $0.04509$ & $21.634$ & $0.04658$ \\
AJD-BC-D & CONF & $47.3$ & $26.2$ & $22.769$ & $0.03769$ & $20.746$ & $0.04308$ \\
\bottomrule
\end{tabular}
\end{table}

\FloatBarrier

\end{document}